%% file: neurips_2024.tex
\documentclass{article}

    \PassOptionsToPackage{numbers, compress}{natbib}

    \usepackage[final]{neurips_2024}

\usepackage[utf8]{inputenc} 
\usepackage[T1]{fontenc}    
\usepackage{url}            
\usepackage{booktabs}       
\usepackage{amsfonts}       
\usepackage{nicefrac}       
\usepackage{microtype}      
\usepackage{xcolor}         
\definecolor{nipsblue}{rgb}{0.21,0.49,0.74}
\definecolor{darkerblue}{rgb}{0,0.08,0.45}

\usepackage[pagebackref,breaklinks,colorlinks,citecolor=darkerblue]{hyperref}

\usepackage{graphicx} 
\usepackage{caption}
\usepackage{adjustbox}
\usepackage{array}
\usepackage{calc}
\usepackage{amsmath}
\usepackage{multirow}
\usepackage{subfigure} 

\usepackage{wrapfig}
\usepackage{marvosym}

\usepackage{verbatim}
\usepackage{bbm}

\newcommand{\modelname}{DreamClear}
\newcommand{\gendataname}{GenIR}
\newcommand{\eg}{\textit{e.g.}}
\newcommand{\ie}{\textit{i.e.}}

\NewDocumentCommand{\qz}{ mO{} }{\textcolor{blue}
{\textsuperscript{\textit{QZ}}\textsf{\textbf{\small[#1]}}}}
\NewDocumentCommand{\xt}{ mO{} }{\textcolor{red}{\textsuperscript{\textit{XT}}\textsf{\textbf{\small[#1]}}}}
\NewDocumentCommand{\zy}{ mO{} }{\textcolor{green}{\textsuperscript{\textit{ZY}}\textsf{\textbf{\small[#1]}}}}
\NewDocumentCommand{\zhouxq}{ mO{} }{\textcolor{blue}{\textsuperscript{\textit{ZhouXQ}}\textsf{\textbf{\small[#1]}}}}
\NewDocumentCommand{\HHB}{ mO{} }{\textcolor{cyan}{\textsuperscript{\textit{HHB}}\textsf{\textbf{\small[#1]}}}}

\title{DreamClear: High-Capacity Real-World Image Restoration with Privacy-Safe Dataset Curation}

\author{Yuang Ai${}^{\clubsuit,\heartsuit}$\quad Xiaoqiang Zhou${}^{\clubsuit,\diamondsuit}$ \quad Huaibo Huang${}^{\clubsuit,\heartsuit,}$\textsuperscript{\Letter}\\ \textbf{Xiaotian Han}${}^{\spadesuit}$ \quad \textbf{Zhengyu Chen}${}^{\spadesuit}$ \quad 
\textbf{Quanzeng You}${}^{\spadesuit}$ \quad \textbf{Hongxia Yang}${}^{\spadesuit}$\\
${}^{\clubsuit}$MAIS \& NLPR, Institute of Automation, Chinese Academy of Sciences \\
${}^{\heartsuit}$School of Artificial Intelligence, University of Chinese Academy of Sciences \\
${}^{\spadesuit}$ByteDance, Inc~ ${}^{\diamondsuit}$University of Science and Technology of China\\
\texttt{\small shallowdream555@gmail.com, huaibo.huang@cripac.ia.ac.cn}\\
\texttt{\small Code and models:~\url{https://github.com/shallowdream204/DreamClear}
}
}

\begin{document}

\maketitle

\begin{figure}[ht]
\vspace{-24pt}
\begin{center}
	\includegraphics[width=0.97\linewidth]{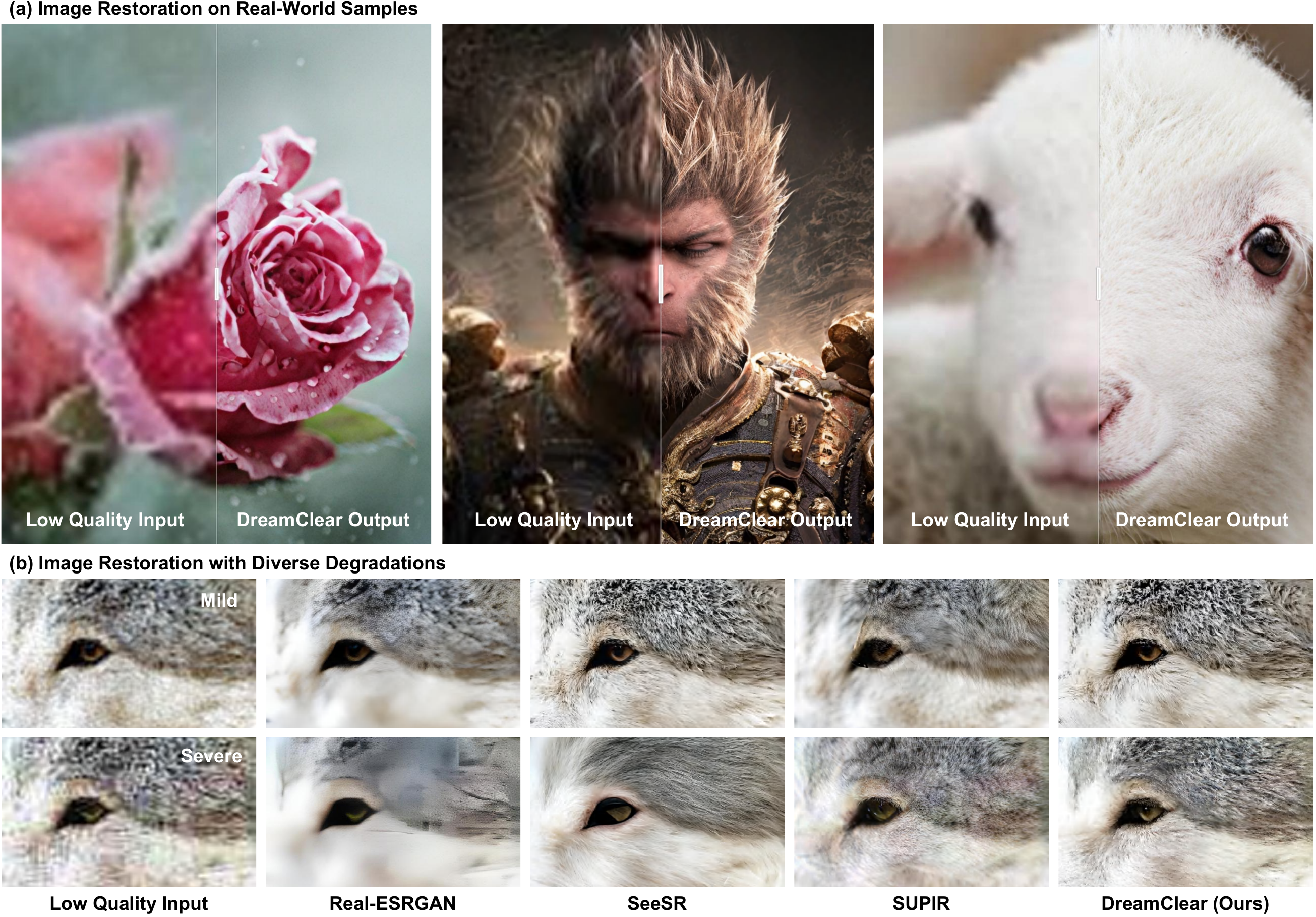}
\end{center}
\vspace{-8pt}
\caption{\small
We present \textbf{DreamClear}, a high-capacity image restoration model that delivers photorealistic restoration of real-world LQ images, outperforming SOTA diffusion-based models in handling diverse degradations.
}
\vspace{-2pt}
\label{fig:fig1}
\end{figure}

\begin{abstract}

Image restoration (IR) in real-world scenarios presents significant challenges due to the lack of high-capacity models and comprehensive datasets.
To tackle these issues, we present a dual strategy: GenIR, an innovative data curation pipeline, and DreamClear, a cutting-edge Diffusion Transformer (DiT)-based image restoration model.
\textbf{GenIR}, our pioneering contribution, is a dual-prompt learning pipeline that overcomes the limitations of existing datasets, which typically comprise only a few thousand images and thus offer limited generalizability for larger models. 
GenIR streamlines the process into three stages: image-text pair construction, dual-prompt based fine-tuning, and data generation \& filtering. This approach circumvents the laborious data crawling process, ensuring copyright compliance and providing a cost-effective, privacy-safe solution for IR dataset construction. The result is a large-scale dataset of one million high-quality images.
Our second contribution, \textbf{DreamClear}, is a DiT-based image restoration model. It utilizes the generative priors of text-to-image (T2I) diffusion models and the robust perceptual capabilities of multi-modal large language models (MLLMs) to achieve photorealistic restoration. To boost the model's adaptability to diverse real-world degradations, we introduce the Mixture of Adaptive Modulator (MoAM). It employs token-wise degradation priors to dynamically integrate various restoration experts, thereby expanding the range of degradations the model can address.
Our exhaustive experiments confirm DreamClear's superior performance, underlining the efficacy of our dual strategy for real-world image restoration.

\end{abstract}

\section{Introduction} \label{xq_introduction_section_label}

Image restoration (IR), a vital field in computer vision, targets transforming degraded low-quality (LQ) images into high-quality (HQ) counterparts. While IR has achieved significant advancements under predefined conditions, such as super-resolution~\cite{zhang2023ntire,cao2023ntire} and denoising~\cite{chen2023masked,li2023ntire} tasks, real-world IR remains a formidable challenge due to the diversity and complexity of degradation types. The disconnect between training data and real-world scenarios is substantial, as existing datasets inadequately encapsulate the intricacies of real-world degradations.
Efforts to bridge this gap include domain adaptation~\cite{ai2024uncertainty,DADA,DRN,CinCGAN}, dataset collection~\cite{drealsr,realsr,chen2019camera,zhang2019zoom}, and degradation simulation~\cite{realesrgan,bsrgan,PDMSR,wu2022animesr}. However, in contrast to the leaps in Natural Language Processing (NLP)~\cite{achiam2023gpt} and AI-Generated Content (AIGC)~\cite{ldm} enabled by large-scale models and extensive data, IR's progress is not as pronounced. Real-world challenges persist, and the potential of large-scale data and high-capacity models remains largely untapped.
This leads us to two critical questions: \textit{how can we obtain a large-scale dataset that accurately represents real-world IR, and based on this, how can we construct powerful models tailored for real-world IR scenarios?}

Addressing the first question, considerable efforts have been made to curate IR datasets. Given the challenge of collecting real-world paired IR data, these datasets are typically constructed by acquiring HQ images and then simulating degradations to generate corresponding LQ images. While many works~\cite{realesrgan,bsrgan,PDMSR,wu2022animesr} have refined the degradation simulation process, this paper focuses on the acquisition of HQ images and the associated challenges of copyright and privacy protection.
The predominant method for obtaining HQ images is web scraping. Current open-source IR datasets, such as DIV2K~\cite{div2k} and Flickr2K~\cite{flickr2k}, contain only a few thousand images, insufficient for covering a broad spectrum of real-world scenarios. Larger collections like SUPIR~\cite{supir}, with 20 million images, highlight the labor-intensive nature of large-scale dataset curation. Moreover, images sourced from the internet often involve copyright issues and privacy concerns, particularly with identifiable human faces.
To advance the IR field effectively, there is an urgent need for a dataset curation method that is privacy-safe and cost-effective.

In response,  we present an under-explored approach in the image restoration (IR) field: creating high-quality, non-existent images to enhance dataset curation efficiency, while evading copyright and privacy issues. We unveil \textbf{GenIR}, a privacy-conscious, automated data curation pipeline that repurposes the generative prior in pretrained text-to-image (T2I) models for IR tasks, and uses multimodal large language models (MLLMs) to generate text prompts, thereby improving data synthesis quality.
GenIR operates in three stages: (1) image-text pairs construction, (2) dual-prompt fine-tuning, and (3) data generation \& filtering.
Initially, GenIR utilizes existing IR datasets and the advanced MLLM, Gemini-1.5-Pro~\cite{team2023gemini}, to create image-text pairs, while generating negative samples via an image-to-image pipeline~\cite{meng2021sdedit}.
Subsequently, we apply a dual-prompt learning strategy to adapt pretrained T2I models to the IR task, generating suitable prompts for data synthesis.
In the final stage, MLLMs create various scene descriptions and synthesize images using the adapted image prior, with a focus on ensuring no identifiable individuals are included. MLLMs also assess and filter the synthesized data based on quality, producing high-quality images that are privacy-safe and copyright-free.
Through GenIR, we generate a dataset of one million high-quality images, proving its efficacy in training a robust real-world IR model.

Armed with a large-scale, high-quality image dataset, our focus shifts to the construction of a high-capacity IR model that can robustly generalize to real-world scenarios.  Recent state-of-the-art approaches~\cite{pasd,seesr,supir} employ the generative priors in pretrained Stable Diffusion~\cite{ldm} (SD) for realistic image restoration, underlining the power of rich generative prior in SD. As Fig.\ref{fig:fig1} illustrates, SD-based methods outperform GAN-based ones. However, these strategies often neglect the degradation priors in input low-quality images, a critical element in blind IR~\cite{kdsr}. This insight leads us to investigate the integration of degradation prior into diffusion-based IR models, and how to optimize its synergy with large models.

In this paper, we introduce \textbf{\modelname}, a high-capacity real-world image restoration model, grounded on a large dataset. \modelname{} is based on Diffusion Transformer (DiT)~\cite{dit}, the cornerstone of modern diffusion-based systems (\eg, Sora~\cite{sora}, Stable Diffusion 3~\cite{sd3}). Our model employs a dual-branch framework with textual guidance from multi-modal large language models (MLLMs) for photorealistic restoration. \modelname{} first processes the low-quality image through a lightweight network to produce a reference image. We propose ControlFormer to enhance the control over DiT-based T2I models, thereby better utilizing the low-quality and reference images to guide the content of the generated image. To further improve the model's generalization across diverse and complex degradations, we incorporate implicit prior degradation information to refine the solution space. Specifically, we suggest a Mixture of Adaptive Modulator (MoAM), which extracts token-wise degradation representations and dynamically integrates various restoration experts for each token based on the Mixture-of-Experts (MoE)~\cite{shazeer2017outrageously} structure, thereby enhancing the model's adaptability to different degradation severities (See Fig.~\ref{fig:fig1}).

The main contributions of this work can be summarized as follows:
\begin{itemize}
    \item 
    We propose \gendataname, a pioneering automated data curation pipeline for image restoration. It addresses the urgent need for privacy-safe and cost-effective methods in image restoration, yielding a dataset of one million high-quality images.

    \item We present DreamClear, a robust, high-capacity IR model that incorporates degradation priors into diffusion-based frameworks. This model improves control over content generation, adapts to various degradations, and generalizes well across diverse real-world scenarios.
    \item Extensive experiments across both low-level (synthetic \& real-world) and high-level (detection \& segmentation) benchmarks have demonstrated \modelname's state-of-the-art performance in handling intricate real-world scenarios.   
\end{itemize}

\section{Related Work}
\paragraph{Image Restoration.} Image Restoration aims at restoring a high-quality image from the low-quality input image. Over the past decade, different approaches have been proposed for image super-resolution~\cite{zhou2023msra,huang2017wavelet, sinsr,zhou2024ristra}, denoising~\cite{zhang2018ffdnet, zhang2023xformer, wang2023lg}, deblurring~\cite{realblur, whang2022deblurring, nrknet}, deraining~\cite{drsformer, huang2021memory,huang2022memory,huang2021selective}, inpainting~\cite{zhou2021image,ai2024lora}, etc. Recently, researchers have increasingly focused on enhancing the generalization ability to diverse degradations in real-world applications~\cite{DADA, drealsr, stablesr}. The degradation simulation is improved from simple degradations to complex degradation processes, such as BSRGAN~\cite{bsrgan}, Real-ESRGAN~\cite{realesrgan} and AnimeSR~\cite{wu2022animesr}. With the improved degradation simulation process, many recent methods can deal with diverse degradation types and achieve promising performance in real-world scenarios~\cite{dasr, realsr}. With the paired data, training a randomly initialized restoration model from scratch is one way to improve generalization ability~\cite{liang2021swinir}. The other way is to exploit the generative prior in the pre-trained generative model, such as GAN or diffusion models~\cite{pan2021exploiting,stablesr,ai2024multimodal,ddrm,ddnm,deqir}. In this work, we propose a data synthesis pipeline and introduce a real-world image restoration model with high generalizability.

\paragraph{Generative Prior.} Generative models learn the image synthesis process and embed the image prior in the model weights. The image prior in a high-quality image generator, such as StyleGAN~\cite{karras2019style} and Stable Diffusion~\cite{ldm}, can be adapted to other visual restoration tasks~\cite{dasr, realsr, lee2024ugpnet, fei2023generative}. To use the image prior in GANs, an additional encoder is often applied to convert the input image to the latent space~\cite{yang2021gan, liu2023survey}. For the diffusion models, the forward process adds noise to the image gradually and finally converts the image to the latent noise space~\cite{choi2021ilvr, meng2021sdedit}. By manipulating in the latent feature space, the input image is integrated into the generation process as a conditional input, and the synthesis process exploits the image prior in the pre-trained models. The generative prior in the pre-trained models can also serve as a good initialization for downstream synthesis tasks~\cite{stablesr, ye2023ip, karras2023dreampose}. We exploit the generative prior in the pre-trained diffusion models to synthesize datasets for image restoration tasks and train a restoration model for real-world applications.

\paragraph{Synthetic Dataset.} Data size and data quality are widely recognized as essential for many vision tasks. A large-scale high-quality dataset can facilitate the large model training and improve the model ability greatly~\cite{zhai2022scaling, gu2019div8k, fu2021dvg, guo2016ms,han2024infimm}. Existing large-scale datasets are often manually collected with laborious human efforts~\cite{deng2009imagenet, li2023lsdir}. More importantly, the data crawled from the internet may leak privacy information~\cite{ristani2016performance, guo2016ms}, raising concerns related to AI security. The synthesized datasets can not only reduce the laborious human efforts, but also avoid the privacy information leakage. High-quality synthesized datasets are verified to be effective in many vision tasks~\cite{he2022synthetic,hammoud2024synthclip,azizi2023synthetic}. Our work is the first to explore the dataset synthesis in the image restoration field.

\begin{figure}[t] 
\centering
    \includegraphics[width=1.0\textwidth]{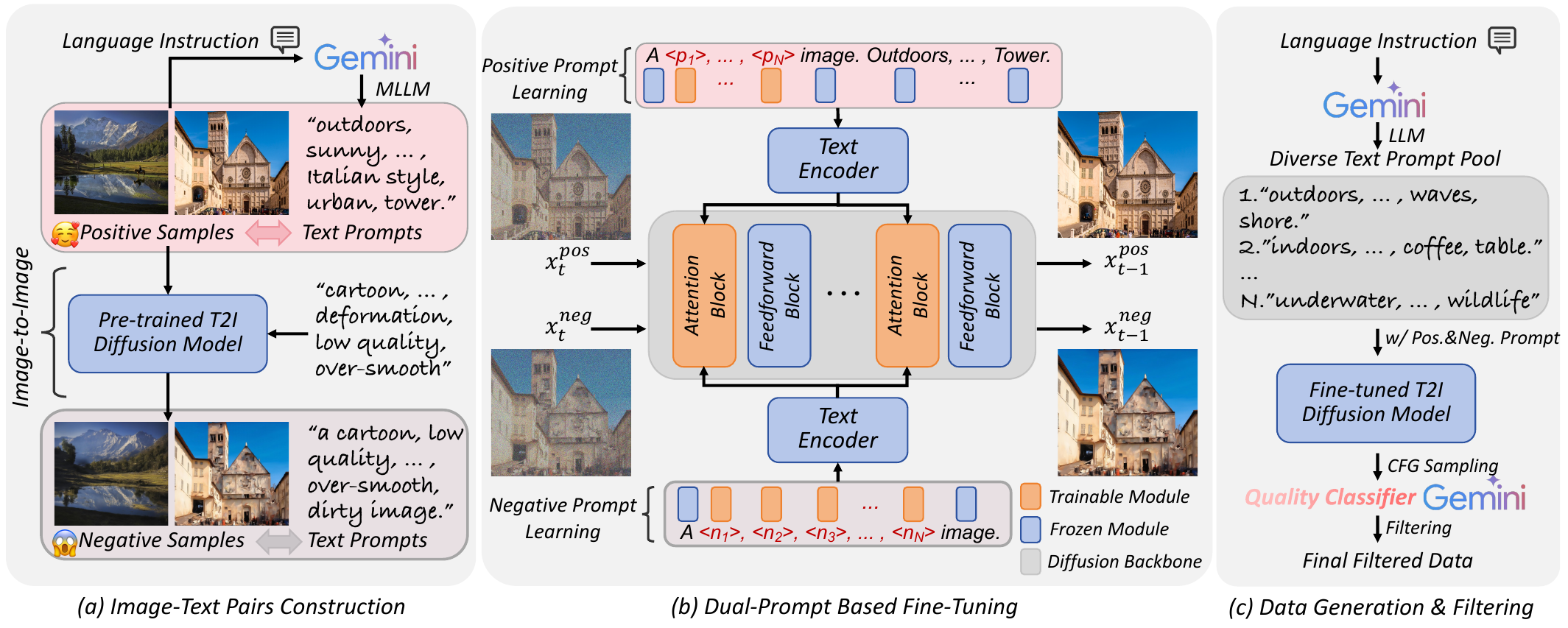}
    \caption{An overview of the three-stage \textbf{\gendataname{}} pipeline, which includes (a) Image-Text Pairs Construction, (b) Dual-Prompt Based Fine-Tuning, and (c) Data Generation \& Filtering.} 
    \label{fig:data}
    \vspace{-0.1cm}
\end{figure}

\section{Privacy-Safe  Dataset Curation Pipeline}
\label{sec:method_data}

Traditionally, IR datasets are created by scraping web images and simulating degradations to generate low-quality (LQ) counterparts. This process is labor-intensive and rife with copyright and privacy issues, especially with identifiable human faces. 
To advance the IR field, a privacy-safe and cost-effective dataset curation method is needed. 
Drawing inspiration from the success of text-to-image (T2I) models in synthesizing high-quality images, we introduce the GenIR pipeline. This novel approach leverages the generative priors of pre-trained T2I models to construct extensive, privacy-safe datasets. However, the efficacy of T2I models is contingent upon carefully crafted prompts for generating high-quality images fitting for IR tasks.

To tackle this, GenIR, as illustrated in Fig.~\ref{fig:data}, employs a streamlined three-stage process. Initially, we construct positive and negative samples, each paired with corresponding text prompts. Subsequently, a dual-prompt based finetuning strategy concurrently learns both positive and negative prompts. Finally, we utilize LLMs to generate a diverse array of text prompts, leading to the creation and filtering of data. Throughout this process, we maintain stringent privacy standards, ensuring no specific personal information is embedded in the text prompts or the generated images.

\paragraph{Image-Text Pairs Construction.}
We use high-resolution, texture-rich images in existing IR datasets~\cite{div2k,flickr2k,gu2019div8k,li2023lsdir} as positive samples. Given the unavailability of corresponding text prompts, we employ the sophisticated MLLM, Gemini-1.5-Pro~\cite{team2023gemini}, to generate necessary prompts via language instructions.
Moreover, to identify and eliminate unwanted content during the T2I model's fine-tuning and enhance image quality, we generate negative samples representing undesirable outcomes using the T2I model. As depicted in Fig.~\ref{fig:data} (a), we adopt the image-to-image pipeline proposed in~\cite{meng2021sdedit}, using the T2I model and manually designed prompts such as ``\textit{cartoon, painting, ... , over-smooth, dirty}'', to directly generate negative samples. 

\paragraph{Dual-Prompt based Fine-Tuning.}

Rather than relying on complex, labor-intensive prompts with limited applicability, we propose an innovative dual-prompt based fine-tuning approach to refine the T2I model for our data needs. As illustrated in Fig.~\ref{fig:data} (b), we employ positive and negative samples to learn their corresponding prompts.
Specifically, we use $M$ positive tokens $\{\left \langle p_1 \right \rangle ,\cdots,\left \langle p_M \right \rangle\}$ and $N$ negative tokens $\{\left \langle n_1 \right \rangle ,\cdots,\left \langle n_N \right \rangle\}$ to represent desired and undesired attributes, respectively, and subsequently learn their embeddings. We initialize these new positive and negative tokens using frequently used positive (e.g., ``\textit{4k, highly detailed, professional ...}'') and negative text prompts (e.g., ``\textit{deformation, low quality, over-smooth ...}''). As the text condition is integrated into the diffusion model via cross-attention, we also refine the attention block to better comprehend these new tokens. After fine-tuning the T2I model with our curated image-text pairs, we can efficiently employ the learned prompts and refined diffusion model to readily generate the needed data.

\paragraph{Data Generation \& Filtering.} 

In addition to the quality of images, the diversity of scenes within the IR dataset is of paramount importance. To address this, we leverage Gemini to generate one million text prompts, describing varied scenes under carefully curated language instructions. These instructions explicitly proscribe the inclusion of personal or sensitive information, thereby ensuring privacy. As depicted in Fig.~\ref{fig:data} (c), we employ the fine-tuned T2I model in conjunction with the learned positive and negative prompts to generate HQ images.

Classifier-free guidance (CFG)~\cite{ho2022classifier} provides a mechanism to effectively utilize negative prompts, thereby mitigating the generation of undesired content. During the sampling phase, the denoising model $\epsilon_\theta$ anticipates two outcomes: one associated with the positive prompt $pos$ and the other with the negative prompt $neg$. The final CFG prediction is formulated as
\begin{align}
        \epsilon_\theta(z_t,t,pos,neg)=\omega\times\epsilon_\theta(z_t,t,pos)+(1-\omega )\times\epsilon_\theta(z_t,t,neg),
\end{align}
where $\omega$ denotes the CFG guidance scale. Post-sampling, the generated images are evaluated by a quality classifier, which decides whether to retain the images based on the predicted probabilities. This binary classifier is trained on positive and negative samples. Gemini is subsequently used to ascertain whether the images exhibit blatant semantic errors or inappropriate content.

Contrasted with direct web crawling, our  \gendataname{} provides a more cost-effective and privacy-preserving approach to data acquisition. It circumvents the potential infringement of personal privacy information prevalent on the web, thereby ensuring our research is both ethical and secure - a crucial aspect in the current artificial intelligence landscape characterized by extensive data usage. Ultimately, we gather one million high-resolution ($2040\times1356$) images, each of superior quality.

\section{High-Capacity Image Restoration Model} \label{xq_method_DreamClear_label}
The complex and varied degradation of real-world images presents a major challenge to the practical applicability of restoration models. We introduce \modelname{}, a high-capacity image restoration model that dynamically integrates various restoration experts, guided by prior degradation information. \modelname{} is built upon on PixArt-$\alpha$~\cite{chen2023pixart}, a pre-trained T2I diffusion model based on the Diffusion Transformer (DiT)~\cite{dit} architecture, which has proven its powerful generative capabilities~\cite{opensora}.

\begin{figure}[tb] 
\centering
    \includegraphics[width=1.0\textwidth]{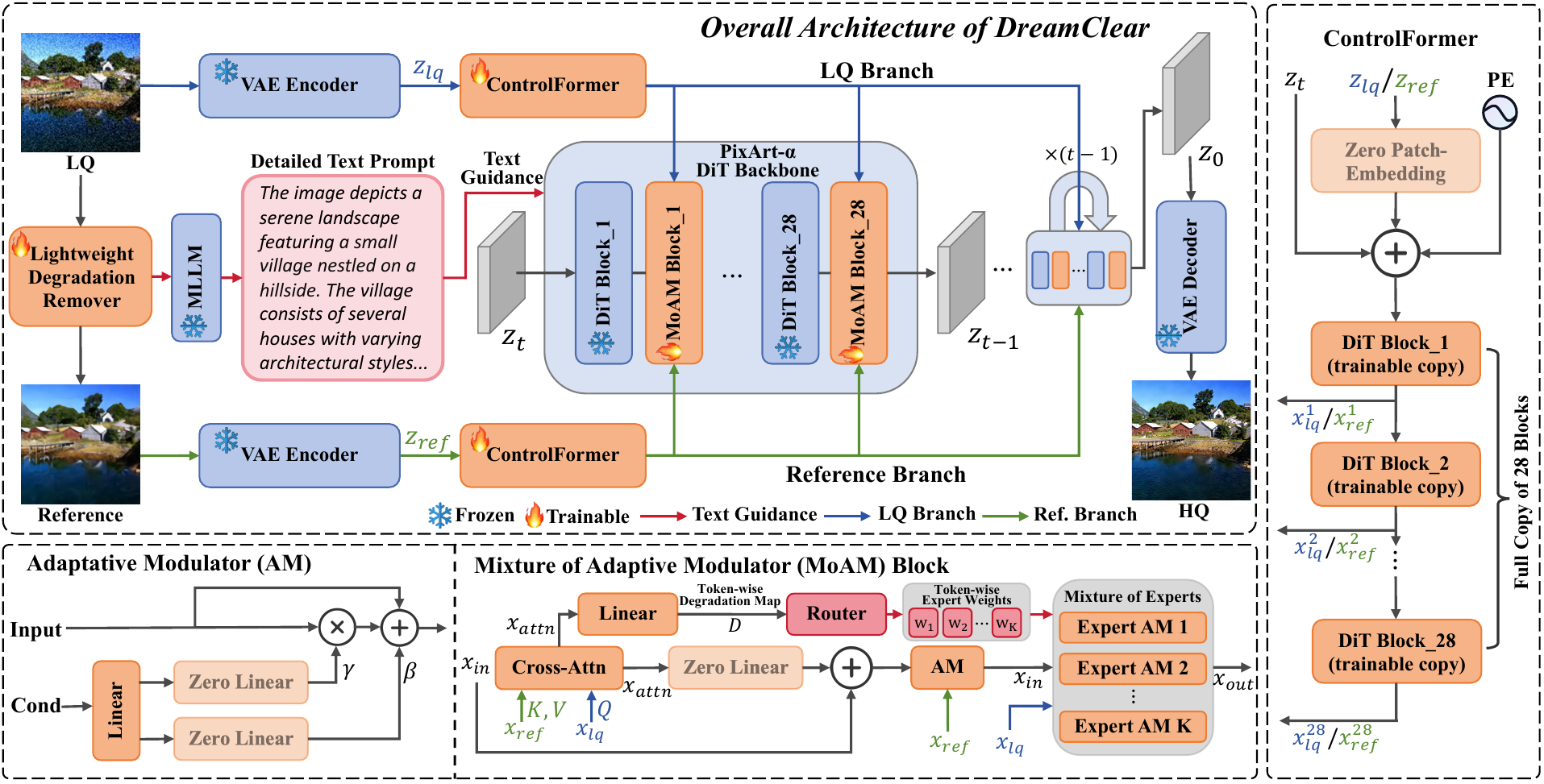}
    \caption{Architecture of the proposed \textbf{\modelname}. \modelname{} adopts a dual-branch structure, using Mixture of Adaptive Modulator to merge LQ features and Reference features. We utilize MLLM to generate detailed text prompt as the guidance for T2I model.} 
    \label{fig:arch}
\end{figure}

\paragraph{Architecture Overview.}
Fig.~\ref{fig:arch} illustrates that \modelname{} features a dual-branch architecture, encompassing an LQ Branch and a Reference Branch. LQ images $I_{lq}$ are processed by SwinIR~\cite{liang2021swinir}, a lightweight degradation remover, resulting in smoother, albeit less detailed, reference images $I_{ref}$. Considering potential detail loss in $I_{ref}$, we employ both $I_{lq}$ and $I_{ref}$ to direct the diffusion model. Moreover, we utilize the open-source MLLM, LLaVA~\cite{llava}, to generate detailed captions for training images using the prompt ``\textit{Describe this image and its style in a very detailed manner}'', supporting the T2I diffusion model in attaining more realistic restoration.

\paragraph{ControlFormer.}

ControlNet~\cite{controlnet}, a prevalent structure for managing diffusion models, is tailored for the U-Net~\cite{unet} structure in SD. It is unsuitable for DiT, stemming from the architecture difference. To address this, we present ControlFormer, which inherits ControlNet's core features (trainable copy blocks and zero-initialized layers) but is adapted for the DiT-based T2I model, as shown in Fig.~\ref{fig:arch}. ControlFormer, duplicating all DiT Blocks from PixArt-$\alpha$, employs the MoAM block to combine LQ features $x_{lq}$ and reference features $x_{ref}$. This DiT-optimized ControlFormer maintains ControlNet's essential components, providing effective spatial control within DiT.

\paragraph{Mixture-of-Adaptive-Modulator.}

To enhance our model's robustness to real-world degradations, we propose a degradation-aware Mixture-of-Adaptive-Modulator (MoAM) for effective LQ and reference feature fusion. As shown in Fig.~\ref{fig:arch}, MoAM consists of adaptive modulators (AM), a cross-attention layer, and a router block. AM employs AdaLN~\cite{adaln} to learn dimension-wise scale $\gamma$ and shift $\beta$ parameters, embedding conditional information into input features.

MoAM operates in three steps: 1) For DiT features $x_{in}$, we calculate the cross-attention output $x_{attn}\in\mathbb{R}^{N\times C}$ between $x_{lq}\in\mathbb{R}^{N\times C}$ and $x_{ref}\in\mathbb{R}^{N\times C}$, where $N$ and $C$ denote the number of visual tokens and hidden size. $x_{in}$ is then modulated using $x_{attn}$ followed by a zero linear layer. A token-wise degradation map $D\in\mathbb{R}^{N\times C}$ is generated through the linear mapping of $x_{attn}$. 2) Features are further modulated using AM, with $x_{ref}$ as the AM condition to extract clean features. 3) We adopt a mixture of degradation-aware experts to adapt to diverse degradations, detailed below.

Given varying degradations in real-world images, our method dynamically processes tokens using degradation priors. Each MoAM block consists of $K$ restoration experts (\ie, AM) $\{E_1,\cdots,E_K\}$, each specialized for specific degradation scenarios. A routing network $R(\cdot)$ dynamically merges expert guidance for tokens, based on $D$. The routing network, a two-layer MLP followed by softmax, yields token-wise expert weights $w=R(D)\in\mathbb{R}^{N\times K}$. The dynamic mixture of restoration experts is formulated as
\begin{align}
    \gamma (i)=\sum _{k=1}^{K}w(i,k)\times Net_k^{\gamma }[x_{lq}(i)],\quad\beta (i)=\sum _{k=1}^{K}w(i,k)\times Net_k^{\beta }[x_{lq}(i)],\label{eq:moe}
\end{align}
\begin{align}
    x_{out}=(1+\gamma )\otimes x_{in}+\beta,
\end{align}
where $i$ and $k$ index tokens and experts respectively, $Net^\gamma$ and $Net^\beta$ map within an expert to $\gamma$ and $\beta$, and $\otimes$ denotes element-wise multiplication. MoAM dynamically fuses expert knowledge, leveraging degradation priors to tackle complex degradations.

\section{Experiments} \label{xq_experiments_label}
\input{tables/compare_sota}
\input{tables/compare_coco_ade}

\subsection{Experimental Setup} \label{xq_exp_setup_label}
\paragraph{Datasets.} We adopt a combination of DIV2K~\cite{div2k}, Flickr2K~\cite{flickr2k}, LSDIR~\cite{li2023lsdir}, DIV8K~\cite{gu2019div8k}, and our generated dataset to train \modelname{}. We employ the Real-ESRGAN~\cite{realesrgan} degradation pipeline to generate LQ images, using the same degradation settings as SeeSR~\cite{seesr} to ensure a fair comparison. All experiments are conducted on scaling factor $\times4$.

For testing datasets, following previous works~\cite{stablesr,supir,seesr}, we evaluate our method on both synthetic and real-world benchmarks. For synthetic benchmarks, we randomly crop 3,000 patches from the validation sets of DIV2K and LSDIR, and degrade them using the same settings as training. We name these two benchmarks as \textit{DIV2K-Val} and \textit{LSDIR-Val} respectively. For real-world benchmarks, we conduct experiments on commonly used \textit{RealSR}~\cite{realsr} and \textit{DRealSR}~\cite{drealsr} datasets. Besides, we establish another real-world benchmark, called \textit{RealLQ250}, which includes a total of 250 LQ images of size $256\times256$ used in previous works~\cite{seesr,liang2022details,realesrgan,zhang2017beyond,supir} or sourced from the Internet, without corresponding GT images. For all testing datasets with GT images, the resolution of the HQ-LQ image pairs is $1024\times1024$ and $256\times256$, respectively.

\paragraph{Metrics.}
Following SeeSR~\cite{seesr}, we adopt PSNR and SSIM (calculated on the Y channel of transformed YCbCr space) as reference-based distortion metrics, LPIPS~\cite{lpips} and DISTS~\cite{dists} as reference-based perceptual metrics, NIQE~\cite{niqe}, MANIQA~\cite{yang2022maniqa}, MUSIQ~\cite{yang2022maniqa} and CLIPIQA~\cite{yang2022maniqa} as no-reference metrics. FID~\cite{fid} is used to evaluate the image quality. These metrics can achieve a comprehensive evaluation of the restoration effects.

\paragraph{Implementation Details}  \label{implement_details}
For training GenIR and DreamClear, we both use the original latent diffusion loss~\cite{ldm}.
The proposed GenIR framework is built on SDXL~\cite{podell2023sdxl} and trained over 5 days using 16 NVIDIA A100 GPUs. The training is conducted on $1024\times1024$ resolution images with a batch-size of 256. For data generation, we use 256 NVIDIA V100 GPUs and spend 5 days to generate the large-scale dataset. Our \modelname{} is built upon PixArt-$\alpha$~\cite{chen2023pixart} and LLaVA~\cite{llava}. The SwinIR model in DiffBIR~\cite{diffbir} is used as the lightweight degradation remover. We use the AdamW optimizer with a learning rate of $5e^{-5}$ to train our model. The training is conducted on $1024\times1024$ resolution images, running for 7 days on 32 NVIDIA A100 GPUs with a batch-size of 128. The number of experts $K$ in Eq.~(\ref{eq:moe}) is set to $3$. For inference of DreamClear, we adopt iDDPM~\cite{iddpm} with 50 sampling steps, CFG guidance scale $\omega=4.5$.

\input{figs/tex/sota_compare}
\input{figs/tex/userstudy}

\subsection{Comparison with State-of-the-Art Methods}\label{xq_compare_sota_label}
We compare our method with state-of-the-art GAN-based methods (BSRGAN~\cite{bsrgan}, Real-ESRGAN~\cite{realesrgan}, SwinIR-GAN~\cite{liang2021swinir}, and DASR~\cite{dasr}) and recent diffusion-based methods (StableSR~\cite{stablesr}, DiffBIR~\cite{diffbir}, ResShift~\cite{resshift}, SinSR~\cite{sinsr}, SeeSR~\cite{seesr}, and SUPIR~\cite{supir}).

\paragraph{Quantitative Comparisons.} 

Tab.~\ref{tab:methods} presents quantitative results on various benchmarks. Our method consistently excels in perceptual metrics (LPIPS, DISTS, FID) on synthetic datasets, signifying high perceptual quality. On real-world benchmarks, our method performs strongly across most no-reference metrics (NIQE, MANIQA, MUSIQ, CLIPIQA), attesting to the high quality of our restorations. Our diffusion-based method prioritizes photorealistic restoration. Despite lower PSNR/SSIM scores, recent works~\cite{depictqa_v1,supir} argue these metrics inadequately represent visual quality, and it is necessary to reconsider the reference values of existing metrics and propose more effective methods to evaluate modern image restoration methods. We believe that as the field of image quality assessment (IQA) evolves, more suitable metrics will emerge to adequately measure the performance of advanced IR models.

\paragraph{Qualitative Comparisons.} We provide qualitative comparisons in Fig.~\ref{fig:sota_compare}. When handling severe degradations (the first row), only our \modelname{} can not only reason the correct structure  but also generate clear details, while other methods may generate deformed structure and blurry results. When it comes to real-world images, our method can achieve results that are rich in detail and more natural (the third row). More real-world visual comparisons are in Apendix~\ref{app:visual-results}.

\paragraph{User Study.} 
We conducted a user study to evaluate our model's restoration quality using 100 low-quality images, restored by our method and five others. Users were asked to rank the restorations considering visual quality, naturalness, and accuracy, among others. The study, involving 256 evaluators, was designed for fairness and wide participation.
Two metrics, vote percentage and Top-K ratio, were used to analyze the results. As shown in Fig.~\ref{fig:user-study}, our model led on both metrics, receiving over 45\% of total votes and being the top choice for 80\% of the images, demonstrating its consistent superiority in producing high-quality images. Refer to Appendix~\ref{app:user-study} for more details.

\subsection{Evaluation on Downstream Benchmarks}

We assess the benefits of image restoration for downstream high-level vision tasks by conducting detection and segmentation experiments on the COCO 2017~\cite{coco} and ADE20K~\cite{ade20k} datasets using various restoration models. Low-quality images are generated and restored under the same conditions as in training. We use the robust visual backbone RMT~\cite{fan2023rmt}  (with Mask R-CNN~\cite{he2017mask} 1$\times$ for COCO, with UperNet~\cite{xiao2018unified} for ADE20K) for these tasks. Tab.~\ref{tab:coco_ade} shows that our model obtains the best performance, implying its superiority in benefiting downstream tasks. Despite its superior performance in semantic restoration and fine-grained image recognition tasks, there's still substantial room for improvement.

\subsection{Ablation Study}

\input{figs/tex/abla_data}
\input{tables/abla_model}

\paragraph{Analysis of Generated Dataset for Real-World IR.} 

Due to the extensive time required to train diffusion-based models, we use SwinIR-GAN to investigate the impact of generated datasets on real-world image restoration (IR). SwinIR is trained on varying quantities of generated data for comparison with the DF2K-trained model. Fig.~\ref{fig:ablation:datascaling} shows that the model trained on an equivalent number of generated images exhibits marginally lower perceptual but higher no-reference metrics than the DF2K model. As the dataset size increases, all metrics improve, reinforcing our belief that larger datasets enhance model generalizability and restoration performance. Notably, the model trained with 100,000 generated images outperforms the DF2K model, underscoring the advantages of utilizing large-scale synthetic datasets for real-world IR. More ablations of GenIR are provided in Appendix~\ref{app:ablation}.

\paragraph{Ablations of \modelname{}.} 
We conduct ablation studies to scrutinize the contribution of each component within \modelname{}. Evaluating perceptual fidelity via LPIPS, DISTS, and FID metrics, and assessing the image quality of restoration results with MANIQ, MUSIQ, and CLIPIQA, we find that \modelname{} outperforms its ablated versions across most metrics, substantiating the importance of these components (Tab.~\ref{tab:ablation_model}). Notably, null prompts slightly outperform detailed prompts on perceptual metrics. However, superior results on no-reference and high-level vision metrics suggest that the MLLM-provided detailed text prompts better preserve semantic information. Visual comparisons detailed in Appendix~\ref{app:ablation} further reinforce the benefits of semantic guidance in image restoration provided by text prompts.

\section{Limitations and Broader Impact} \label{xq_limitation_section_label}
Our model leverages the generative prior of pre-trained diffusion models for image restoration, with diverse synthesized datasets used during training to enhance model performance. In situations of severe image degradation, while our method could predict reasonable and realistic results, the synthesized texture details may not exist in the ground-truth image. A high-quality reference image or explicit human instruction may compensate such a limitation in some degree.

Another limitation lies in the deployment in practical applications. Our model is a diffusion-based model and it needs multiple inference steps to restore the input low-quality image. While our model can predict more plausible results than existing methods, it can not meet the requirement of real-time inference speed in many practical applications. Model distillation and model quantization may compensate the limitation of inference speed.

This paper is a purely academic study of real-world image restoration (IR). However, considering image restoration’s vital role in many practical applications, this work may not only bring some positive societal influence, \eg, improving the quality of images captured by smartphones, but also lead to some potential risks like privacy information leakage from photos on social media. However, the positive societal effects of image restoration far exceed the potential negative impacts, and people may make use of some other techniques, such as inpainting and watermarking, to remove the private information in images.

\section{Conclusion}
To address the challenges in real-world image restoration (IR), we develop \textbf{GenIR}, a privacy-safe automated pipeline that generates a large-scale dataset of one million high-quality images, serving as a robust training resource for IR models. Additionally, we introduce \textbf{\modelname}, a potent IR model that seamlessly integrates degradation priors into diffusion-based IR models. It introduces the novel Mixture of Adaptive Modulator (MoAM) to adapt to diverse real-world degradations.  Our comprehensive experiments underline its outstanding performance in managing complex real-world situations, marking a substantial progression in IR.

\section*{Acknowledgements} 
This research is partially funded by Beijing Nova Program (20230484276), and Youth Innovation Promotion Association CAS (Grant No. 2022132).

\bibliography{ref}
\bibliographystyle{cite}

\clearpage
\appendix

\section{Appendix}
\subsection{More Implementation Details}
For generating negative samples, the strength in SDEdit~\cite{meng2021sdedit} is set to $0.6$.  To minimize the risk of generating images that contain private information in GenIR, we employ Gemini for both text prompt filtering and generated image filtering. The prompts for Gemini are set as ``\textit{You are an AI language assistant, and you are analyzing a series of text prompts. Your task is to identify whether these text prompts contain any inappropriate content such as personal privacy violations or NSFW material. Delete any inappropriate text prompts and return the remaining ones in their original format.}'' and ``\textit{You are an AI visual assistant, and you are analyzing a single image. Your task is to check the image for any anomalies, irregularities, or content that does not align with common sense or normal expectations. Additionally, identify any inappropriate content such as personal privacy violations or NSFW material. If the image does not contain any of the aforementioned issues, it has passed the inspection. Please determine whether this image has passed the inspection (answer yes/no) and provide your reasoning.}'', respectively.
\subsection{More Details of User Study} \label{app:user-study}

To evaluate our approach, we conducted a user study emphasizing restoration quality. We randomly selected 100 low-quality images from the test sets (Tab.~\ref{tab:methods}), applying our model and five other leading methods to produce restored images, generating 100 groups of seven images each.
Users, guided by the low-quality image, were asked to select the best-restored image from each group, considering factors such as visual quality, naturalness, detail accuracy, and absence of distortions or artifacts. Fairness was ensured by presenting each user with 10 randomly selected groups, randomizing image sequence within each group, and masking the methods.
We widely disseminated the online questionnaire without restrictions, amassing feedback from 256 evaluators.

We employed two metrics to quantify our study's results: vote percentage and Top-K ratio. The former represents the proportion of total votes each method received, while the latter measures the frequency a method was among the top-k selections, indicating the proportion of most preferred images and a model's consistency in producing high-quality images. The Top-K ratio is defined as: $ R^{k}_{i} = \frac{1}{N} \sum_{j=1}^{N} \mathbbm{1} (i\in F_{topk}(s_j, k))$, where \(s_j=\{s_{ij}|i=0,...,5\}\) are the selection scores of the \(j\)-th group from \(N\) groups (with \(N=100\)), and \(F_{topk}\) is the top-k operation.

As shown in Fig.~\ref{fig:user-study}, our model outperformed others on both metrics. Our model received over 45\% of the total votes, indicating a strong user preference. Additionally, our method was the top choice for 80\% of the images, and it was ranked first or second for 98\% of the images, highlighting our method's reliable ability to generate superior quality images compared to other methods.


\subsection{More Ablations}\label{app:ablation}

\paragraph{More analysis of \modelname{} Ablation.} 

We give a more detailed analysis in the following.

(a) Mixture of Adaptive Modulator (MoAM). MoAM acts as an interaction module between the LQ branch and the reference branch, aiming to enhance the model's robustness to intricate real-world degradations by explicitly guiding it through the introduction of token-wise degradation priors. It obtains the degradation map through cross-attention between the LQ features and reference features, guiding the dynamic fusion of expert knowledge to modulate LQ features.

Tab.~\ref{tab:ablation_model} presents the ablation studies of MoAM. Notably, when we substitute the Mixture of AM design with AM, all metrics undergo a substantial decrease, underscoring the importance of the degradation prior guidance in MoAM for steering restoration experts. Moreover, we conducted experiments replacing AM with cross-attention and a zero-linear layer. The use of cross-attention, in comparison to AM, leads to a general reduction in model performance. While the zero-linear layer provides minor improvements in MANIQA and MUSIQ scores, it results in a significant drop in perceptual fidelity.

(b) Dual-branch framework. The incorporation of a reference branch allows the model to focus less on degradation removal and more on enhancing image details through generative priors, ultimately producing more photorealistic images. The results in Tab.~\ref{tab:ablation_model} indicate that our dual-branch structure significantly outperforms using only the LQ branch across all metrics.

\begin{figure}[t]  
\centering
  \includegraphics[width=1.0\linewidth]{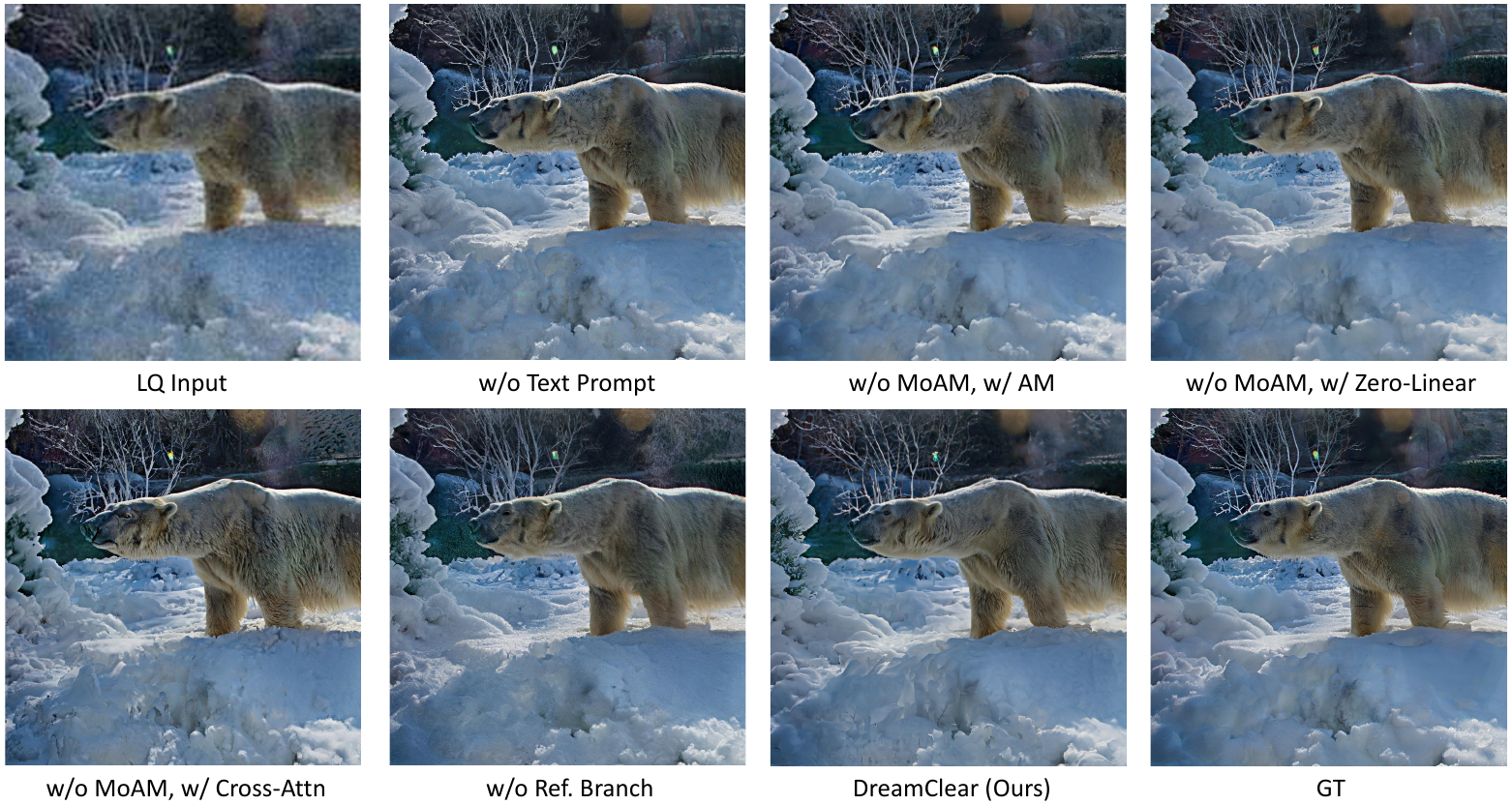}
    \caption{Visual comparisons for ablation study on \modelname{}.}
    \label{fig:abla_visual}
\end{figure}

(c) Text prompt guidance. 
We evaluate the impact of using detailed text prompts generated by MLLMs versus null prompts. Despite null prompts slightly outperforming detailed prompts on perceptual metrics, the superior performance on no-reference and high-level vision metrics indicates that text prompts more effectively retain semantic information. Visual comparisons in Fig.~\ref{fig:abla_visual} further underscore the advantages of text prompts in providing semantic guidance for image restoration.

\input{tables_appendix/abla_genir}

\begin{figure}[t]  
\centering
  \includegraphics[width=1.0\linewidth]{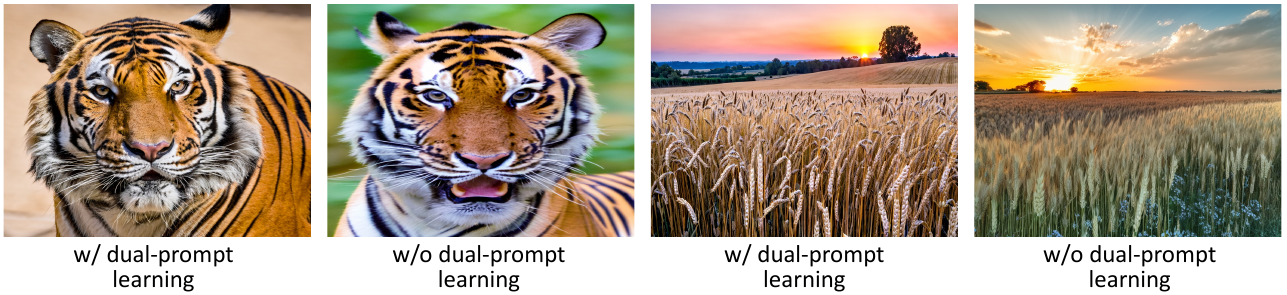}
    \caption{Visual comparisons for ablation study on GenIR.}
    \label{fig:abla_dual_prompt}
\end{figure}

\begin{figure}[t]  
\centering
  \includegraphics[width=1.0\linewidth]{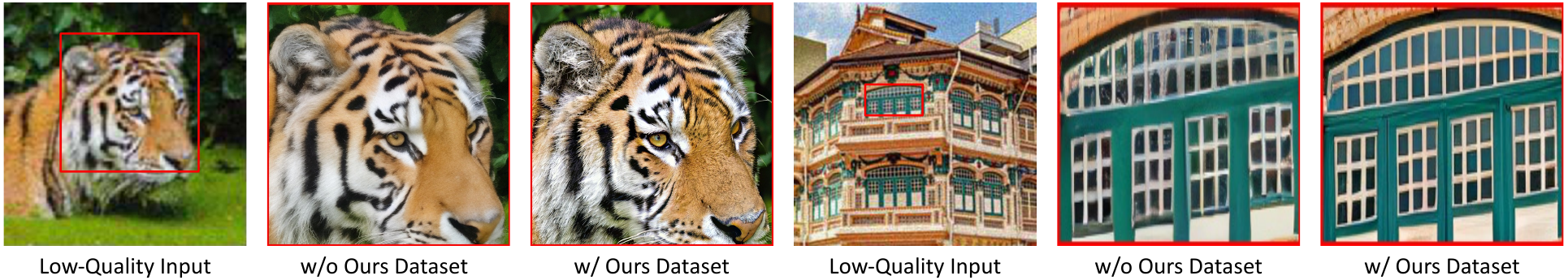}
    \caption{Visual comparisons for ablation study on training datasets.}
    \label{fig:abla_data}
\end{figure}

\begin{figure}[!htbp]
\centering
  \includegraphics[width=1.0\linewidth]{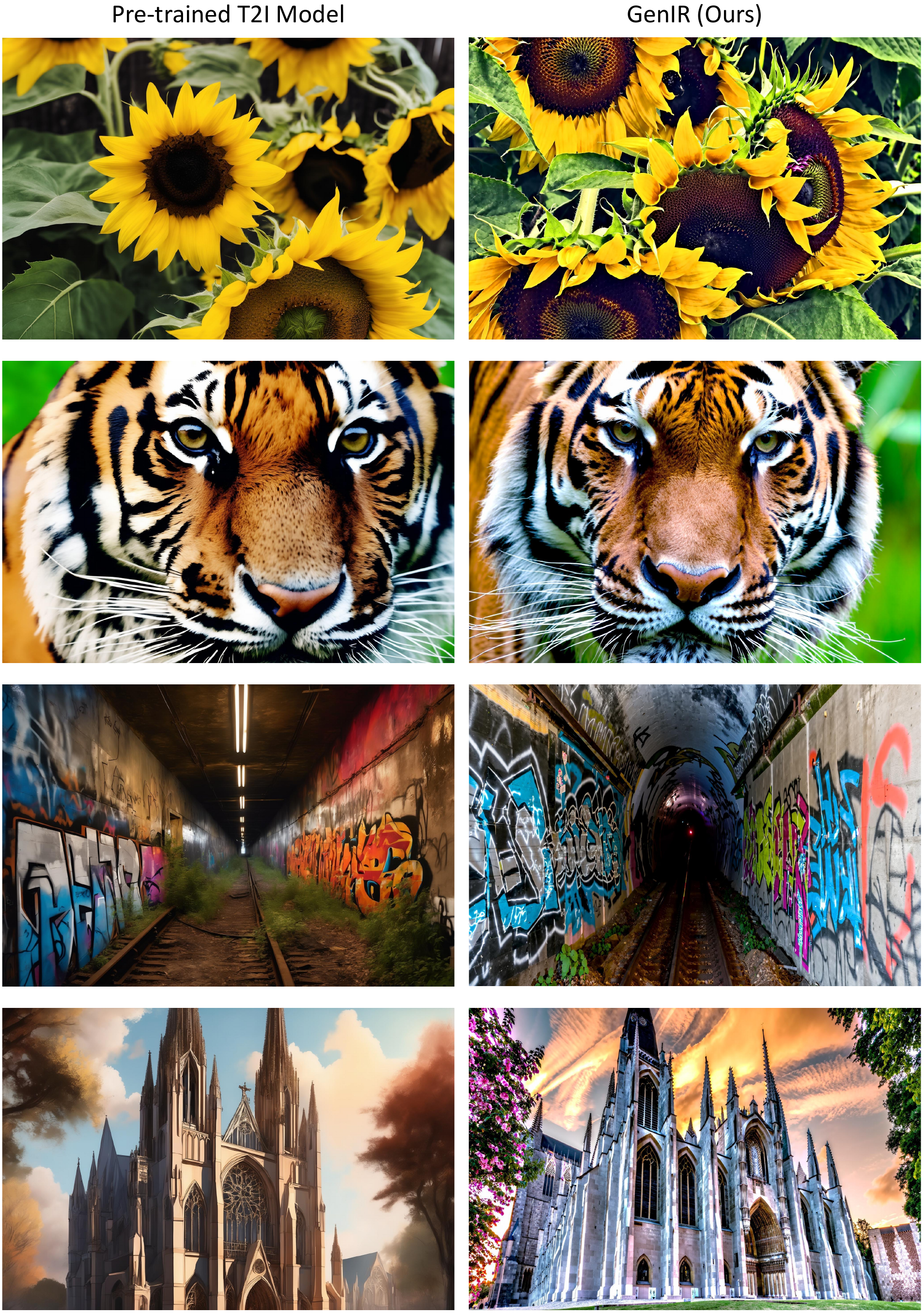}
    \caption{Visual comparison of images generated using the pre-trained T2I model and GenIR. Our proposed GenIR produces images with enhanced texture and more realistic details, exhibiting less blurring and distortion. This makes it better suited for training real-world IR models.} 
    \label{fig:abla_genir_visual}
\end{figure}

\paragraph{Visual results for \modelname{} Ablation.}

Following the setting in Tab.~\ref{tab:ablation_model}, we provide more visual comparison results in Fig.~\ref{fig:abla_visual}. We find that when using a null prompt instead of a text prompt generated by MLLM, there are significant semantic errors in the eyes of the bear in the restoration results. This demonstrates that the semantic information provided by MLLM-generated detailed text prompts helps the model achieve more ideal restoration results. When using AM, zero-linear, and cross-attention instead of MoAM, the model tends to produce results that are either too smooth or contain semantic errors, proving the effectiveness of MoAM. Removing the reference branch results in a significant deterioration of the restoration outcomes. Overall, our full model, \modelname{}, achieves the best results in terms of fidelity and perception.

\paragraph{Ablations for GenIR.} 
We use the exact same prompts and sampling parameters to compare images generated by GenIR with those generated by the originally pre-trained T2I model. As shown in Fig.~\ref{fig:abla_genir_visual}, the images generated by our proposed GenIR are more realistic and contain more texture details, while those generated by the pre-trained T2I model tend to have issues like being overly smooth and blurry. Intuitively, images generated by GenIR are likely to be more helpful for real-world IR. To verify this, following the setting of Fig.~\ref{fig:ablation:datascaling}, we use manually designed prompts with the pre-trained T2I model to generate 3450 images for training SwinIR-GAN. As shown in Tab.~\ref{tab:ablation_genir}, the model trained using images generated by our GenIR shows significant improvements across all metrics, quantitatively demonstrating the effectiveness of our approach. 

In addition, we provide visual comparisons in Fig.~\ref{fig:abla_dual_prompt} to verify the effectiveness of dual-prompt learning in GenIR. It shows that the dual-prompt learning strategy can effectively enhance image texture details, making the generated images more suitable for image restoration training. Fig.~\ref{fig:abla_data} also demonstrates the effectiveness of our generated data in enhancing the visual effect of IR models.

\subsection{More Real-World Visual Comparisons}\label{app:visual-results}
We provide more real-world visual comparisons with state-of-the-art diffusion-based real-world image restoration methods in Fig.~\ref{fig:visual_1}, Fig.~\ref{fig:visual_2} and Fig.~\ref{fig:visual_3}. 

\input{figs_appendix/tex/visual_1}
\input{figs_appendix/tex/visual_2}
\input{figs_appendix/tex/visual_3}

\end{document}

%% file: tables/compare_sota.tex
\begin{table}[t]
\captionsetup{font=small}
\small
\centering
\renewcommand\arraystretch{1.1}
\setlength\tabcolsep{4pt}
\caption{Quantitative comparison with state-of-the-art real-world IR methods on both synthetic and real-world benchmarks. Best and second best performance are highlighted in
\textcolor{red}{\textbf{red}} and \textcolor{blue}{\textbf{blue}}, respectively.}
\vspace{2mm}
\resizebox{\textwidth}{!}{
\begin{tabular}{@{}c|c|cccc|ccccccc@{}}
\toprule
Datasets                                                               & Metrics & BSRGAN \cite{bsrgan} & \begin{tabular}[c]{@{}c@{}}Real- \cite{realesrgan}\\ ESRGAN\end{tabular} & \begin{tabular}[c]{@{}c@{}}SwinIR- \\ GAN~\cite{liang2021swinir}\end{tabular}  & DASR \cite{dasr}  & StableSR \cite{stablesr} & DiffBIR \cite{diffbir} & ResShift \cite{resshift}  & SinSR \cite{sinsr} & SeeSR~\cite{seesr} & SUPIR~\cite{supir} & \textbf{\modelname{}}  \\ \midrule
\multirow{9}{*}{\begin{tabular}[c]{@{}c@{}}\textit{DIV2K-Val}\end{tabular}} & PSNR $\uparrow$   & 19.88&	\textcolor{blue}{\textbf{19.92}}&	19.66&	19.73&	19.73&	\textcolor{red}{\textbf{19.98}}&	19.80 &	19.37&	19.59&	18.68&	18.69  \\
                                                                            & SSIM $\uparrow$   & 0.5137&	\textcolor{red}{\textbf{0.5334}}&	\textcolor{blue}{\textbf{0.5253}}&	0.5122&	0.5039&	0.4987&	0.4985&	0.4613&	0.5045&	0.4664&	0.4766 \\
                                                                            & LPIPS $\downarrow$  & 0.4303&	0.3981&	0.3992&	0.4350&	0.4145&	0.3866&	0.4450&	0.4383&	\textcolor{blue}{\textbf{0.3662}}&	0.3976&	\textcolor{red}{\textbf{0.3657}} \\
                                                                            & DISTS $\downarrow$  & 0.2484&	0.2304&	0.2253&	0.2606&	0.2162&	0.2396&	0.2383&	0.2175&	0.1886&	\textcolor{blue}{\textbf{0.1818}}&	\textcolor{red}{\textbf{0.1637}} \\
                                                                            & FID $\downarrow$    & 54.42&	48.44&	49.17&	59.62&	29.64&	37.00&	46.12&	37.84&	\textcolor{blue}{\textbf{24.98}}&	28.11&	\textcolor{red}{\textbf{20.61}}  \\
                                                                            & NIQE $\downarrow$   & 3.9322&	3.8762&	3.7468&	3.9725&	4.4255&	4.5659&	5.9852&	5.7320&	4.1320&	\textcolor{blue}{\textbf{3.4014}}&	\textcolor{red}{\textbf{3.2126}} \\
                                                                            & MANIQA $\uparrow$ & 0.3514&	0.3854&	0.3654&	0.3110&	0.2942&	0.4268&	0.3782&	0.4206&	\textcolor{red}{\textbf{0.5251}}&	0.4291&	\textcolor{blue}{\textbf{0.4320}} \\
                                                                            & MUSIQ $\uparrow$  & 63.93&	64.50&	64.54&	59.66&	58.60&	64.77&	62.67&	65.27&	\textcolor{red}{\textbf{72.04}}&	\textcolor{blue}{\textbf{69.34}}&	68.44  \\
                                                                            & CLIPIQA $\uparrow$& 0.5589&	0.5804&	0.5682&	0.5565&	0.5190&	0.6527&	0.6498&	0.6961&	\textcolor{red}{\textbf{0.7181}}&	0.6035&	\textcolor{blue}{\textbf{0.6963}} \\ \midrule
\multirow{9}{*}{\begin{tabular}[c]{@{}c@{}}\textit{LSDIR-Val}\end{tabular}} & PSNR $\uparrow$ &   \textcolor{blue}{\textbf{18.27}}&	18.13&	17.98&	18.15&	18.11&	\textcolor{red}{\textbf{18.42}}&	18.24&	17.94&	18.03&	16.95&	17.01  \\
                                                                            & SSIM $\uparrow$ &  0.4673&	\textcolor{red}{\textbf{0.4867}}&	\textcolor{blue}{\textbf{0.4783}}&	0.4679&	0.4508&	0.4618&	0.4579&	0.4302&	0.4564&	0.4080&0.4236 \\
                                                                            & LPIPS $\downarrow$ & 0.4378&	0.3986&	0.4020&	0.4503&	0.4152&	0.4049&	0.4524&	0.4523&	\textcolor{red}{\textbf{0.3759}}&	0.4119&	\textcolor{blue}{\textbf{0.3836}} \\
                                                                            & DISTS $\downarrow$  & 0.2539&	0.2278&	0.2253&	0.2615&	0.2159&	0.2439&	0.2436&	0.2265&	0.1966&	\textcolor{blue}{\textbf{0.1838}}&	\textcolor{red}{\textbf{0.1656}} \\
                                                                            & FID $\downarrow$&    53.25&	46.46&	45.31&	60.60&	31.26&	35.91&	43.25&	36.01&	\textcolor{blue}{\textbf{25.91}}&	30.03&	\textcolor{red}{\textbf{22.06}}  \\
                                                                            & NIQE $\downarrow$   & 3.6885&	3.4078&	3.3715&	3.6432&	4.0218&	4.3750&	5.5635&	5.4240&	4.0590&	\textcolor{red}{\textbf{2.9820}}&	\textcolor{blue}{\textbf{3.0707}} \\
                                                                            & MANIQA $\uparrow$ & 0.3829&	0.4381&	0.3991&	0.3315&	0.3098&	0.4551&	0.3995&	0.4309&	\textcolor{red}{\textbf{0.5700}}&	0.4683&	\textcolor{blue}{\textbf{0.4811}} \\
                                                                            & MUSIQ $\uparrow$  & 65.98&	68.25&	67.10&	60.96&	59.37&	65.94&	63.25&	65.35&	\textcolor{red}{\textbf{73.00}}&	\textcolor{blue}{\textbf{70.98}}&	70.40 \\
                                                                            & CLIPIQA $\uparrow$& 0.5648&	0.6218&	0.5983&	0.5681&	0.5190&	0.6592&	0.6501&	0.6900&	\textcolor{red}{\textbf{0.7261}}&	0.6174&	\textcolor{blue}{\textbf{0.6914}} \\ \midrule
\multirow{9}{*}{\begin{tabular}[c]{@{}c@{}}\textit{RealSR}\end{tabular}}     & PSNR $\uparrow$   & \textcolor{blue}{\textbf{25.01}} & 24.22 & 24.89 & \textcolor{red}{\textbf{25.51}} & 24.60 & 24.77 & 24.94 & 24.47 & 24.66 & 22.67 & 22.56  \\
                                                                            & SSIM $\uparrow$   & 0.7422 & 0.7401 & \textcolor{red}{\textbf{0.7543}} & \textcolor{blue}{\textbf{0.7526}} & 0.7387 & 0.6902 & 0.7178 & 0.6710 & 0.7209 & 0.6567 & 0.6548 \\
                                                                            & LPIPS $\downarrow$  & 0.2853&	0.2901&	\textcolor{red}{\textbf{0.2680}}&	0.3201&	\textcolor{blue}{\textbf{0.2736}}&	0.3436&	0.3864&	0.4208&	0.2997&	0.3545&	0.3684 \\
                                                                            & DISTS $\downarrow$  & 0.1967&	0.1892&	\textcolor{red}{\textbf{0.1734}}	&0.2056&	\textcolor{blue}{\textbf{0.1761}}&	0.2195&	0.2467&	0.2432&	0.2029&	0.2185&	0.2122 \\
                                                                            & FID $\downarrow$    & 84.49&	90.10&	80.07&	91.16&	88.89&	\textcolor{blue}{\textbf{69.94}}&	88.91&	70.83&	71.92&	71.63&	\textcolor{red}{\textbf{65.37}} \\
                                                                            & NIQE $\downarrow$   & 4.9261&	5.0069&	4.9475&	5.9659&	5.6124&	6.1294&	6.6044&	6.4662&	4.9102&	\textcolor{blue}{\textbf{4.5368}} &	\textcolor{red}{\textbf{4.4381}} \\
                                                                            & MANIQA $\uparrow$ & 0.3660 &	0.3656&	0.3432&	0.2819&	0.3465&	0.4182&	0.3781&	0.4009&	\textcolor{red}{\textbf{0.5189}}&	0.4296&	\textcolor{blue}{\textbf{0.4337}} \\
                                                                            & MUSIQ $\uparrow$  & 64.67	&62.06&	60.97&	50.94&	61.07&	61.74&	60.28&	60.36&	\textcolor{red}{\textbf{69.38}}&	\textcolor{blue}{\textbf{66.09}}&	65.33  \\
                                                                            & CLIPIQA $\uparrow$& 0.5329& 	0.4872&	0.4548&	0.3819&	0.5139&	0.6202&	0.5778&	0.6587&	\textcolor{blue}{\textbf{0.6839}}&	0.5371&	\textcolor{red}{\textbf{0.6895}} \\ \midrule
\multirow{9}{*}{\begin{tabular}[c]{@{}c@{}}\textit{DrealSR}\end{tabular}}     & PSNR $\uparrow$   & 27.09  & 26.95                                                  & 27.00  & \textcolor{red}{\textbf{28.19}}  & \textcolor{blue}{\textbf{27.39}}  & 27.31  & 27.16    & 26.15    & 27.10  & 24.41   & 24.48  \\
                                                                            & SSIM $\uparrow$   & 0.7759 & 0.7812                                                 & 0.7815 & \textcolor{red}{\textbf{0.8051}} & \textcolor{blue}{\textbf{0.7830}} & 0.7140 & 0.7388   & 0.6564   & 0.7596 & 0.6696  & 0.6508 \\
                                                                            & LPIPS $\downarrow$  & 0.2950 & 0.2876                                                 & \textcolor{blue}{\textbf{0.2789}} & 0.3165 & \textcolor{red}{\textbf{0.2710}} & 0.3920 & 0.4101   & 0.4690  & 0.3117 & 0.3844  & 0.3972 \\
                                                                            & DISTS $\downarrow$  & 0.1956 & 0.1857                                                & \textcolor{blue}{\textbf{0.1787}} & 0.2072 & \textcolor{red}{\textbf{0.1737}} & 0.2443 & 0.2553   & 0.2540   & 0.2103 & 0.2264  & 0.2145 \\
                                                                            & FID  $\downarrow$   & 84.26 & 83.79                                                & 84.22 & 94.96 & 80.23 & 76.89 & 91.82   & 85.26   & \textcolor{blue}{\textbf{75.07}} & 90.78  & \textcolor{red}{\textbf{74.78}} \\
                                                                            & NIQE $\downarrow$   & 5.5866 & 5.7422                                                 & 5.5749 & 6.9663 & 6.1699 & 6.3433 & 7.5616   & 6.8770   & 5.7696 & \textcolor{blue}{\textbf{5.1115}}  & \textcolor{red}{\textbf{4.6295}} \\
                                                                            & MANIQA $\uparrow$ & 0.3420 & 0.3423                                                 & 0.3269 & 0.2754 & 0.3171 & 0.3801 & 0.3350   & 0.3890   & \textcolor{red}{\textbf{0.4974}} & \textcolor{blue}{\textbf{0.4174}}  & 0.3676 \\
                                                                            & MUSIQ $\uparrow$  & 61.22  & 58.37                                                  & 57.33  & 46.49  & 56.43  & 55.14  & 55.27    & 58.50    & \textcolor{red}{\textbf{67.42}}  & \textcolor{blue}{\textbf{64.53}}   & 59.83 \\
                                                                            & CLIPIQA $\uparrow$& 0.5385 & 0.4847                                                 & 0.4819 & 0.3828 & 0.5344 & 0.6005 & 0.5788   & 0.6734   & \textcolor{red}{\textbf{0.7022}} & 0.5800  & \textcolor{blue}{\textbf{0.6620}} \\ \midrule
\multirow{4}{*}{\begin{tabular}[c]{@{}c@{}}\textit{RealLQ250}\end{tabular}}  & NIQE $\downarrow$   & 4.5229 & 4.1091                                                 & 4.0912 & 4.7486 & 4.6349 & 4.8160      & 5.9727   & 5.7768   & 4.4126 & \textcolor{blue}{\textbf{3.6336}}  & \textcolor{red}{\textbf{3.5556}} \\
                                                                            & MANIQA $\uparrow$ & 0.3523 & 0.3592                                                 & 0.3632 & 0.2782 & 0.2939 & 0.4017      & 0.3816   & 0.4229   & \textcolor{red}{\textbf{0.4992}} & 0.3926  & \textcolor{blue}{\textbf{0.4351}} \\
                                                                            & MUSIQ $\uparrow$  & 63.66  & 62.74                                                  & 63.63  & 53.39  & 57.11  & 62.18      & 61.55    & 64.09   & \textcolor{red}{\textbf{70.57}}  & 66.03   & \textcolor{blue}{\textbf{66.76}}  \\
                                                                            & CLIPIQA $\uparrow$& 0.5695 & 0.5465                                                 & 0.5583 & 0.4671 & 0.5208 & 0.6420      & 0.6298   & 0.7044   & \textcolor{blue}{\textbf{0.7104}} & 0.5800  & \textcolor{red}{\textbf{0.7116}} \\ \bottomrule
\end{tabular}

}
\label{tab:methods}
\end{table}

%% file: tables/compare_coco_ade.tex
\begin{table*}[t]\small
\captionsetup{font=small}
\centering
\caption{Evaluation on COCO val2017 (object detection \& instance segmentation)  and ADE20K (semantic segmentation).  }
\renewcommand\arraystretch{1.1}
\setlength\tabcolsep{4pt}
\resizebox{\textwidth}{!}{

\begin{tabular}{@{}c|cc|cccc|ccccccc@{}}
\toprule
Metrics  & GT & Zoomed LQ & BSRGAN & \begin{tabular}[c]{@{}c@{}}Real-\\ ESRGAN\end{tabular} & \begin{tabular}[c]{@{}c@{}}SwinIR-\\ GAN\end{tabular} & DASR & StableSR & DiffBIR & ResShift & SinSR  & SeeSR & SUPIR & \textbf{\modelname{}} \\ \midrule
Object Detection ($AP^b$) & 49.0   &  7.4  &  11.0      &  12.8     &   11.8  &  10.5    &     16.9   &   \underline{18.7}  &      15.6    &   13.8    &   18.2  &    16.6     &   \textbf{19.3}    \\
Object Detection ($AP^b_{50}$) &  70.6  &  12.0  &   17.6     &   20.7 &   18.9  &  17.0    &   26.7     &   \underline{29.9}  &     25.0     &   22.3     &  29.1   &    27.2     &   \textbf{30.8}    \\
Object Detection ($AP^b_{75}$) &  53.8  &  7.5  &   11.4    &   13.1 &   12.1  &  10.7    &   17.6     &   \underline{19.4}  &     15.9     &   14.2     &  18.9    &    17.0     &   \textbf{19.8}    \\
Instance Segmentation ($AP^m$) &  43.9  & 6.4   &   9.6     &    11.3  &   10.2  &   9.3   &    14.6    &  \underline{16.2}   &   13.6       &    12.0    &   15.9   &    14.1     &    \textbf{16.7}   \\
Instance Segmentation ($AP^m_{50}$) &  67.7  & 11.2   &   16.4     &    19.3  &   17.5  &   15.9   &    24.6   &  \underline{27.5}   &   23.3       &    20.6    &   26.6   &    24.5     &    \textbf{28.2}   \\
Instance Segmentation ($AP^m_{75}$) &  47.3  & 6.3   &   9.7     &    11.5  &   10.2  &   9.4   &   14.9   & \underline{16.6}   &   13.7      &    12.1    &   16.1   &    14.0     &    \textbf{16.8}   \\
Semantic Segmentation (mIoU) &  50.4  &  11.5  & 18.6  &  17.3  &   14.3  &  \underline{30.4}    &   19.6     &  23.6   &    29.7      &    19.6      &   26.9   &   27.7      &   \textbf{31.9}    \\ \bottomrule
\end{tabular}
}
\label{tab:coco_ade}
\end{table*}

%% file: figs/tex/sota_compare.tex
\begin{figure}[t]
	\captionsetup{font=small}
	\scriptsize
	\centering
	\newcommand{\h}{0.105}
	\newcommand{\wa}{0.12}
	\newcommand{\wb}{0.16}
	\newcommand{\g}{-0.7mm}
 \definecolor{softgreen}{rgb}{0.3, 0.7, 0.3}

 	\setlength\tabcolsep{0.8pt}
	\renewcommand{\arraystretch}{1}
	\resizebox{1.00\linewidth}{!} {
			\renewcommand{\h}{0.176}
			\newcommand{\w}{0.176}
				\begin{adjustbox}{valign=t}
					\begin{tabular}{ccccccc}
						\includegraphics[height=\h \textwidth, width=\w \textwidth]{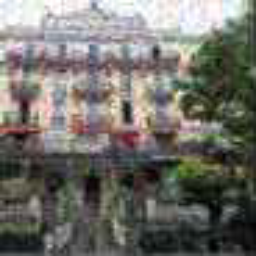} \hspace{\g} &
						\includegraphics[height=\h \textwidth, width=\w \textwidth]{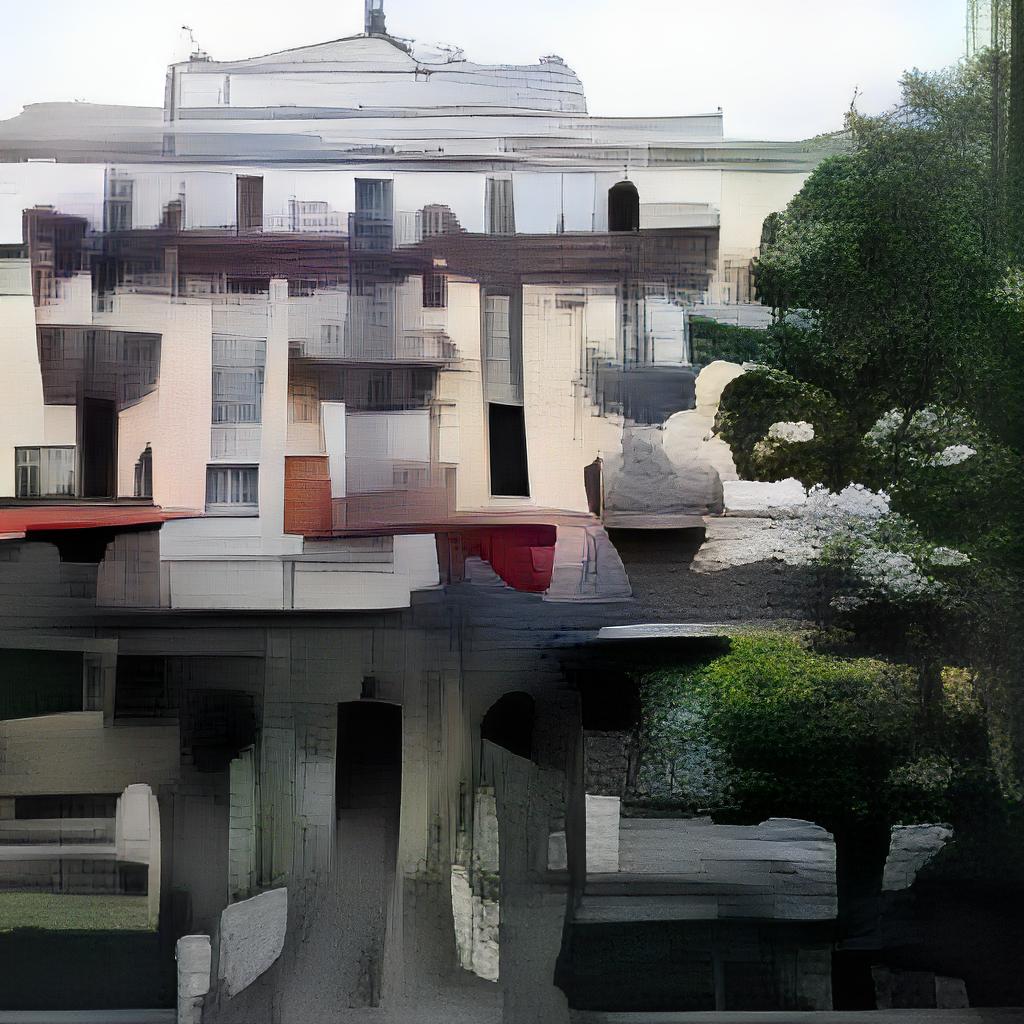} \hspace{\g} &
						\includegraphics[height=\h \textwidth, width=\w \textwidth]{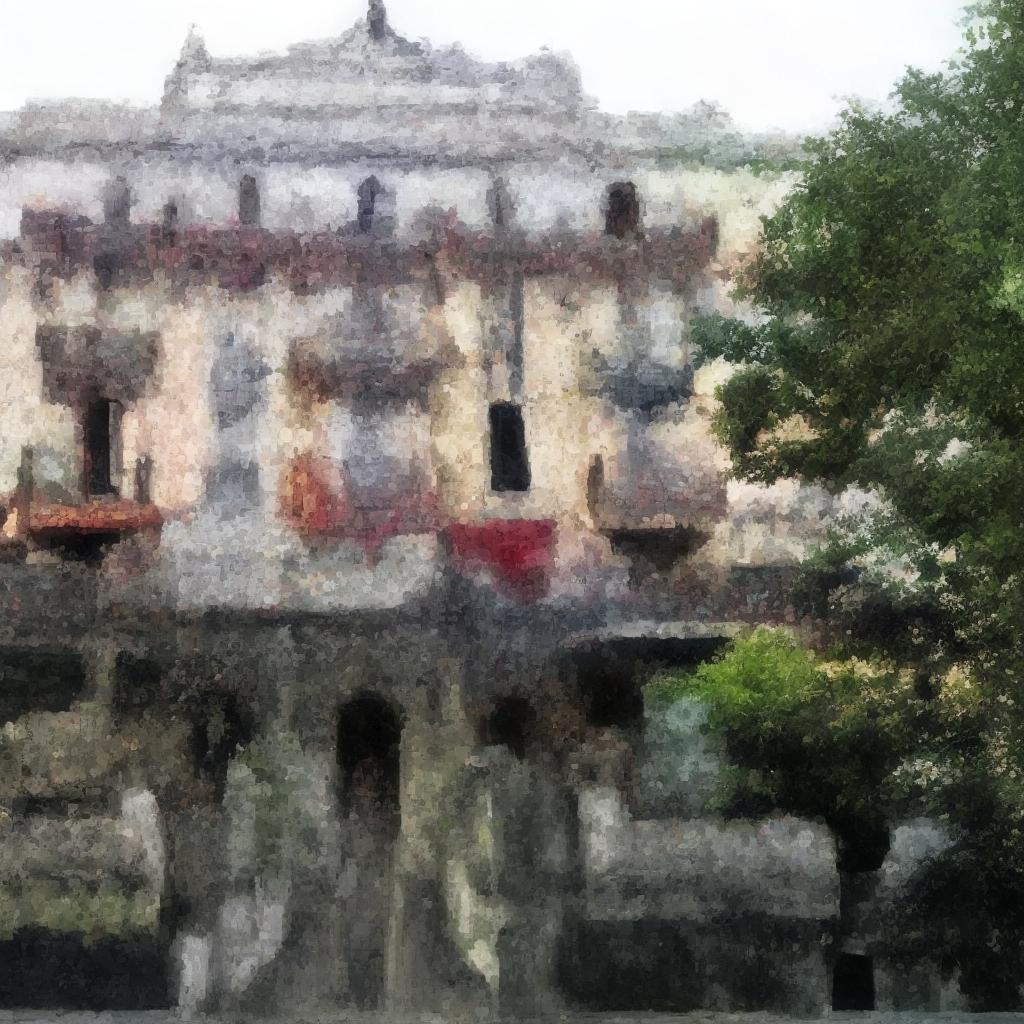} \hspace{\g} &
						\includegraphics[height=\h \textwidth, width=\w \textwidth]{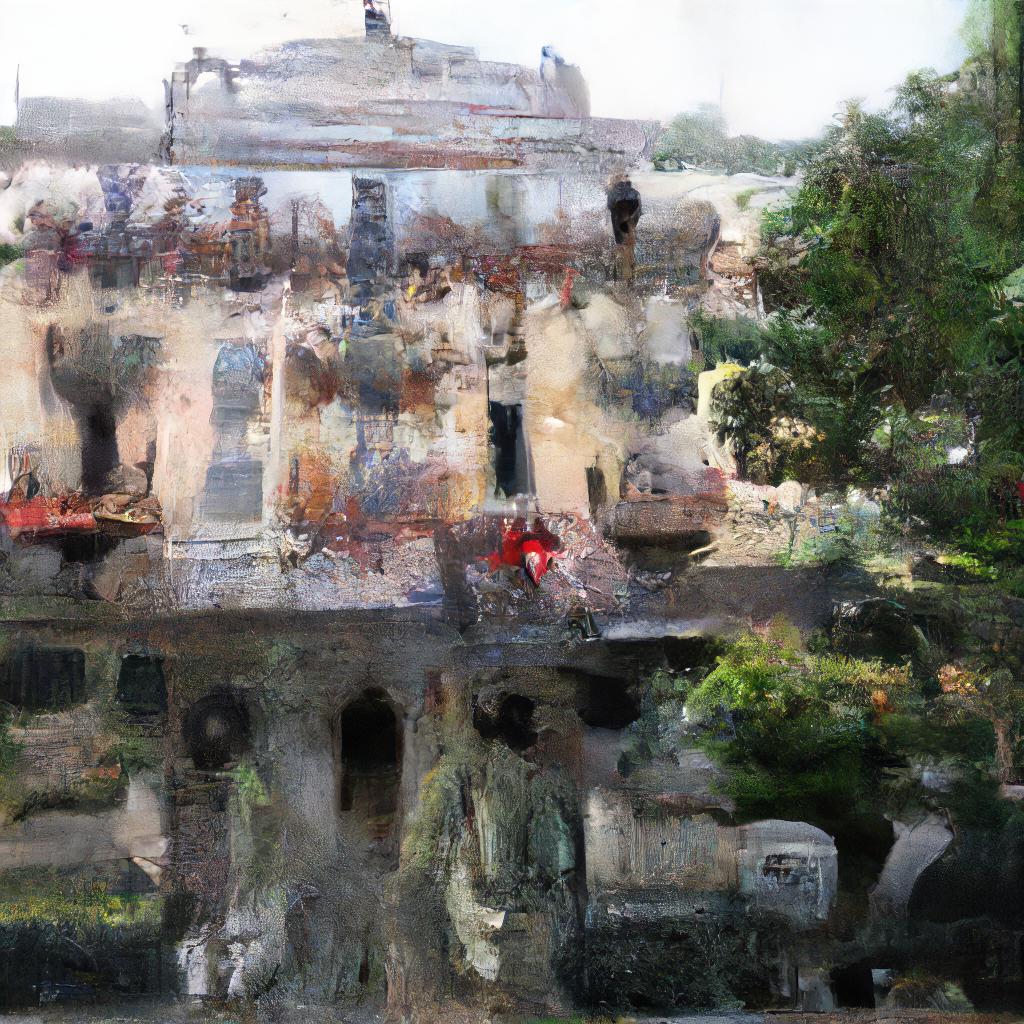} \hspace{\g} &
                            \includegraphics[height=\h \textwidth, width=\w \textwidth]{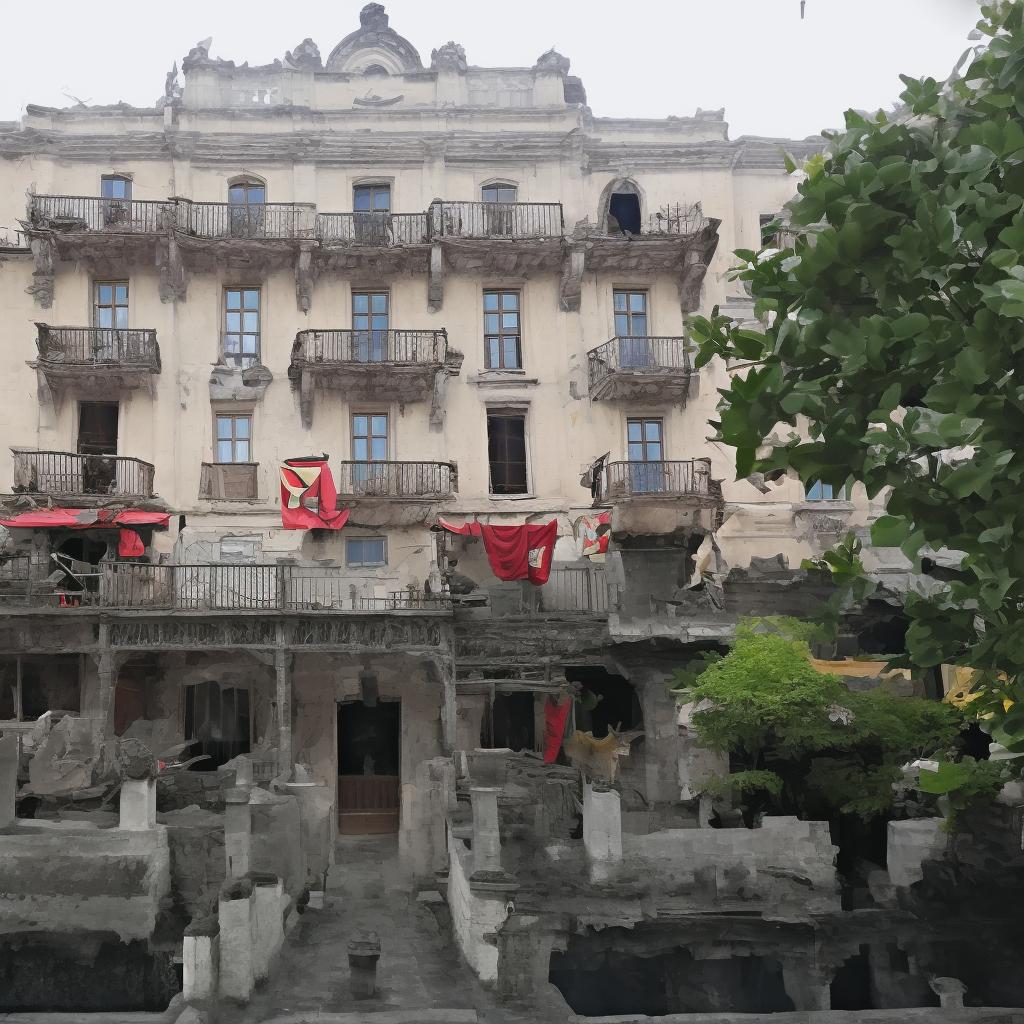}
                            \hspace{\g} &
                            \includegraphics[height=\h \textwidth, width=\w \textwidth]{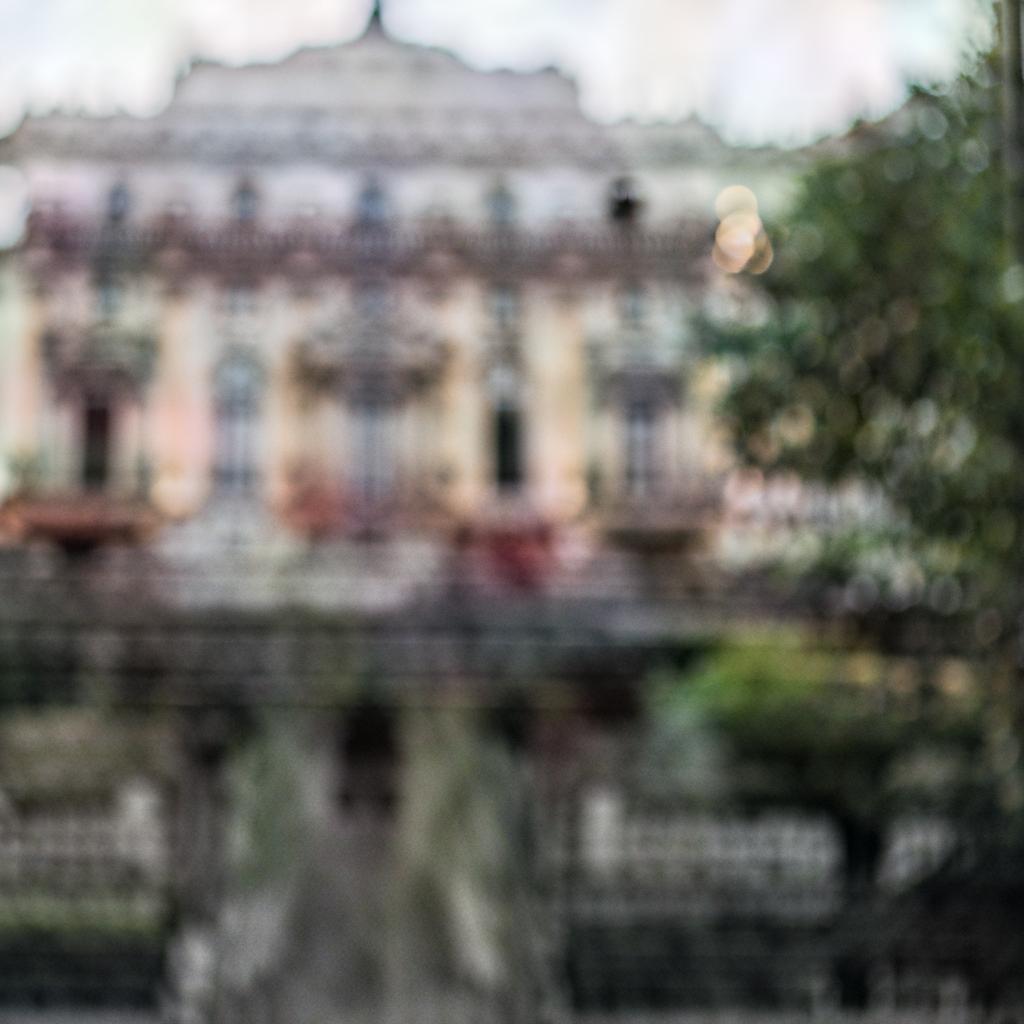}
                            \hspace{\g} &
                            \includegraphics[height=\h \textwidth, width=\w \textwidth]{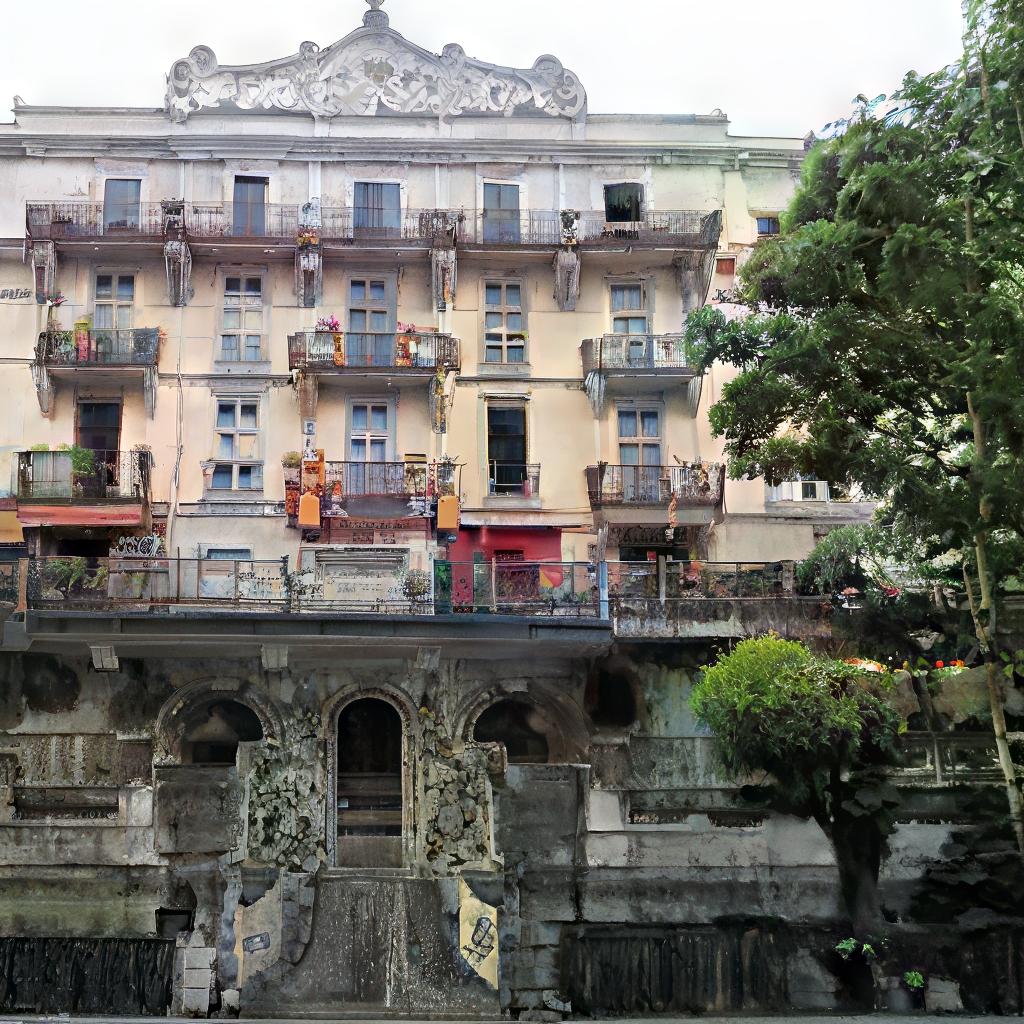}
                                    \vspace{-0.3mm}
						\\
                 						\includegraphics[height=\h \textwidth, width=\w \textwidth]{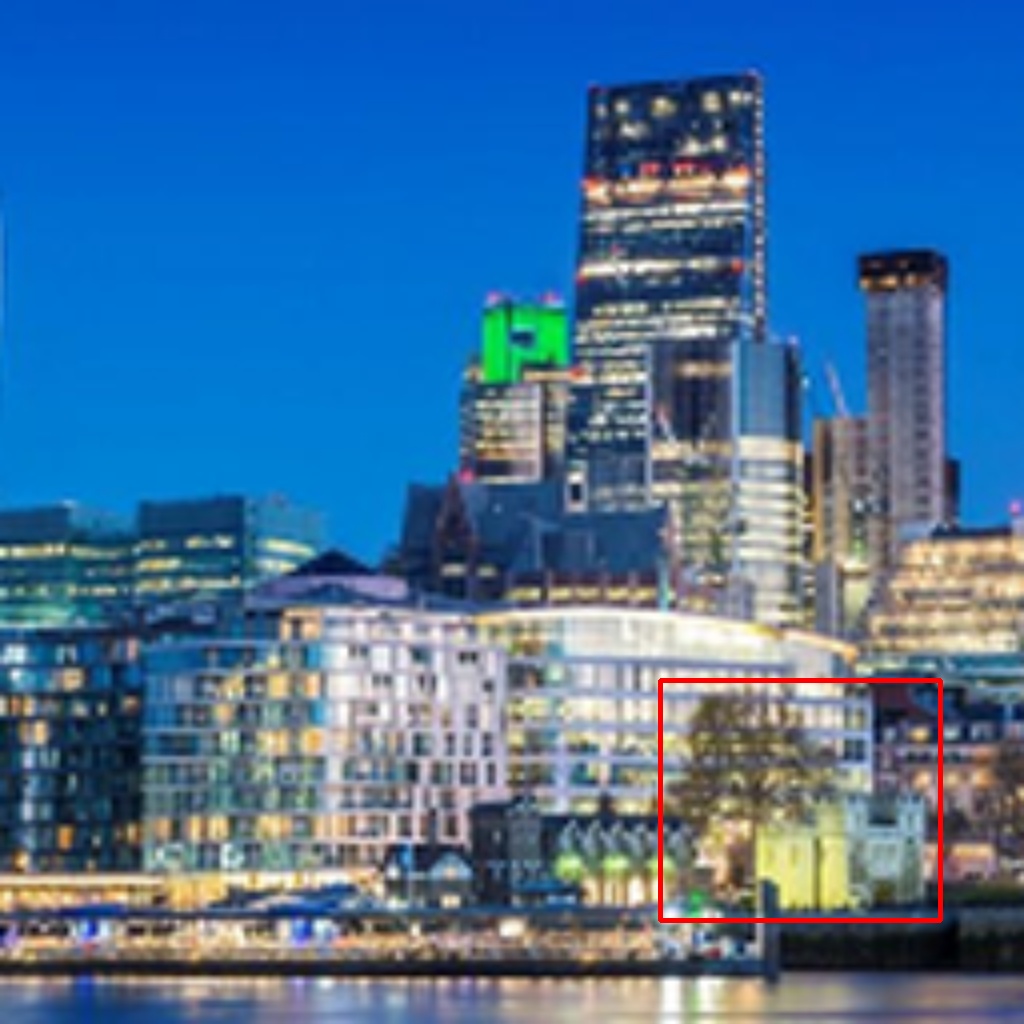} \hspace{\g} &
						\includegraphics[height=\h \textwidth, width=\w \textwidth]{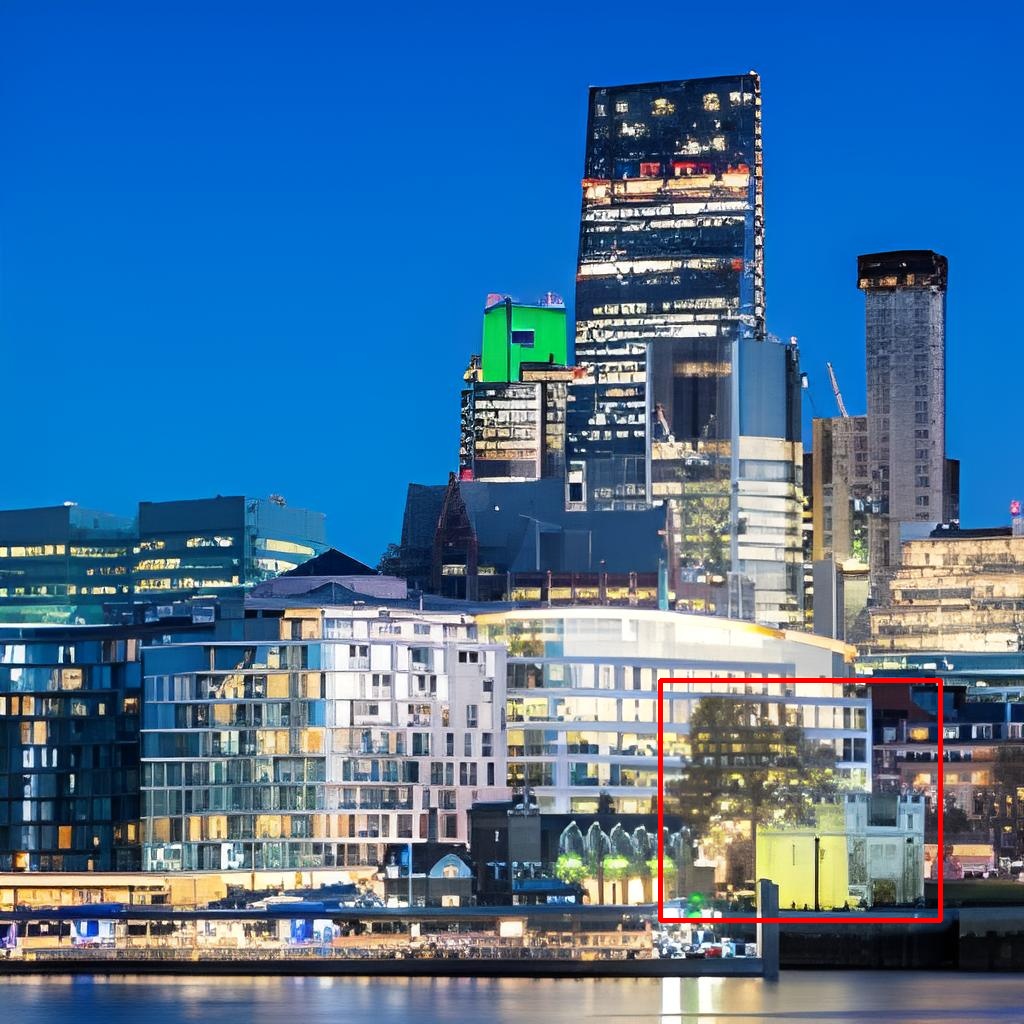} \hspace{\g} &
						\includegraphics[height=\h \textwidth, width=\w \textwidth]{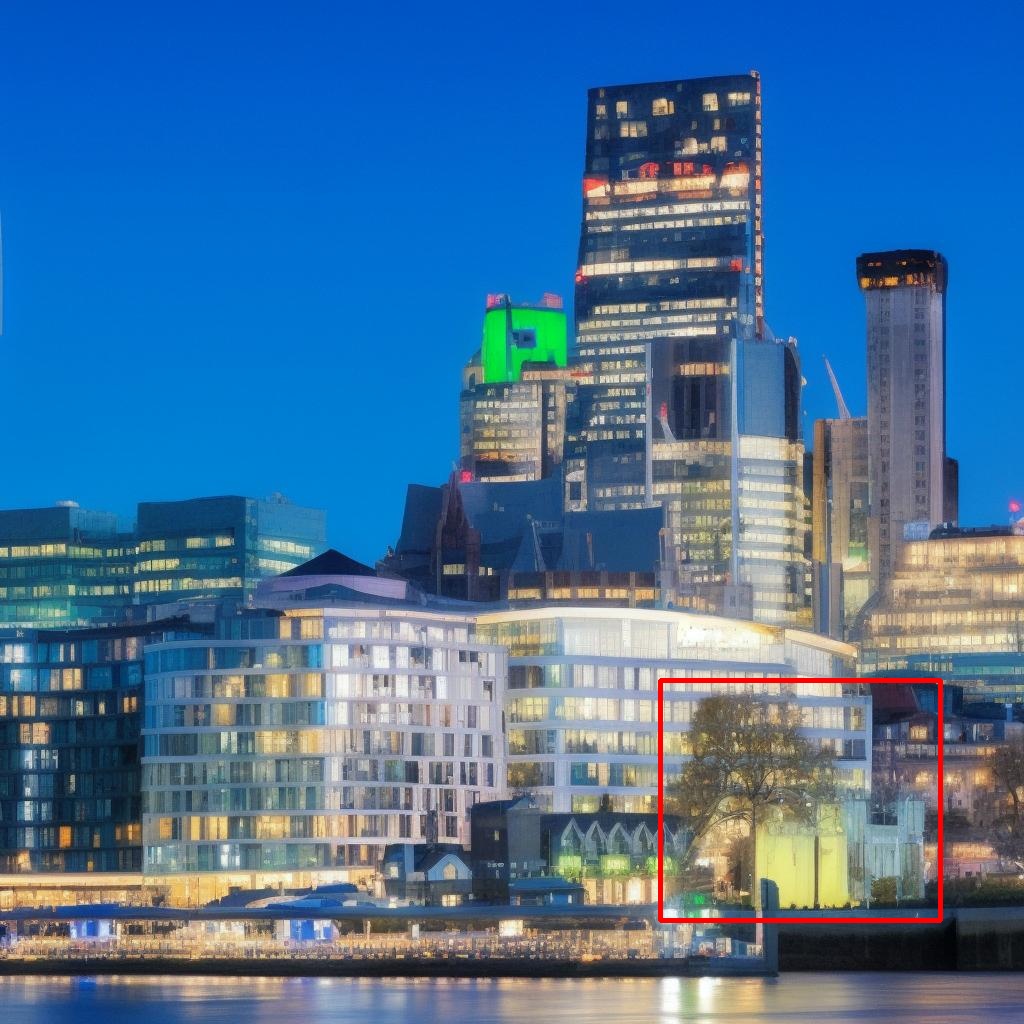} \hspace{\g} &
						\includegraphics[height=\h \textwidth, width=\w \textwidth]{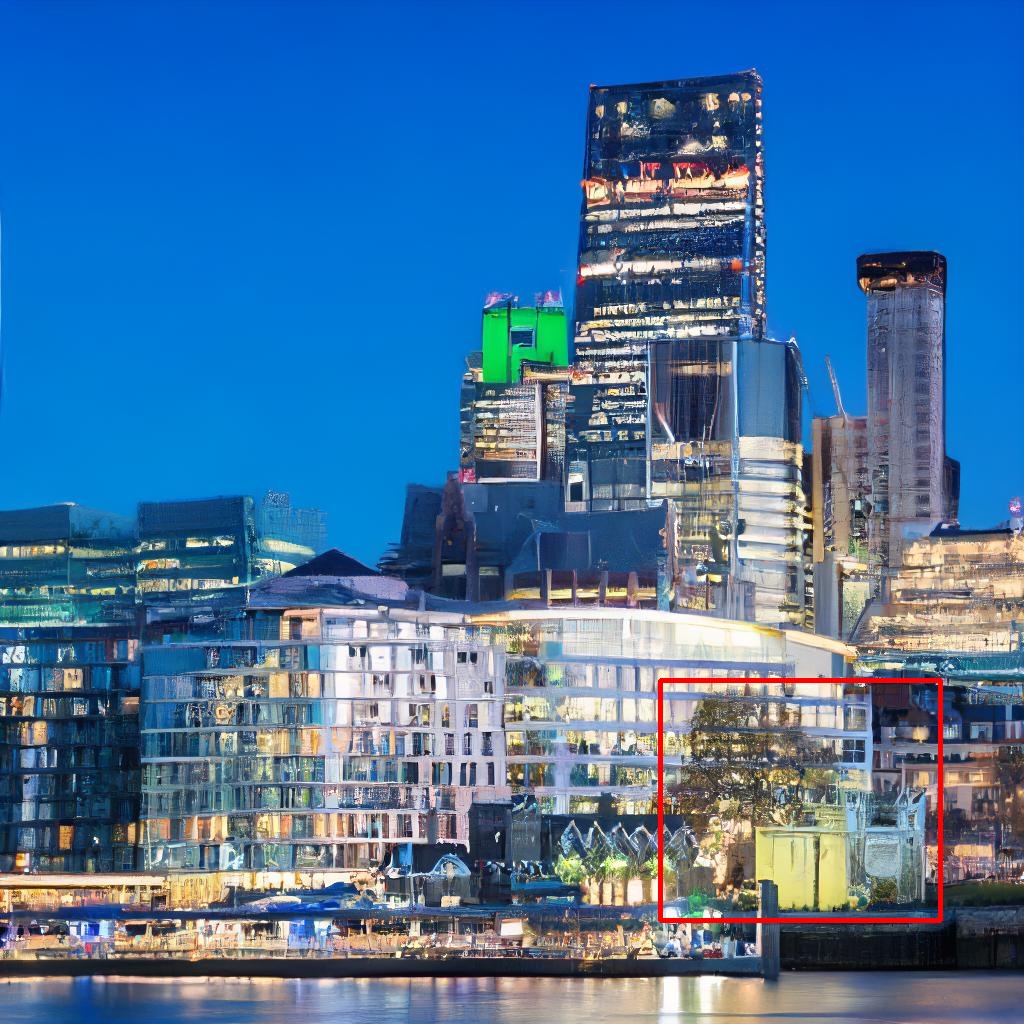} \hspace{\g} &
                            \includegraphics[height=\h \textwidth, width=\w \textwidth]{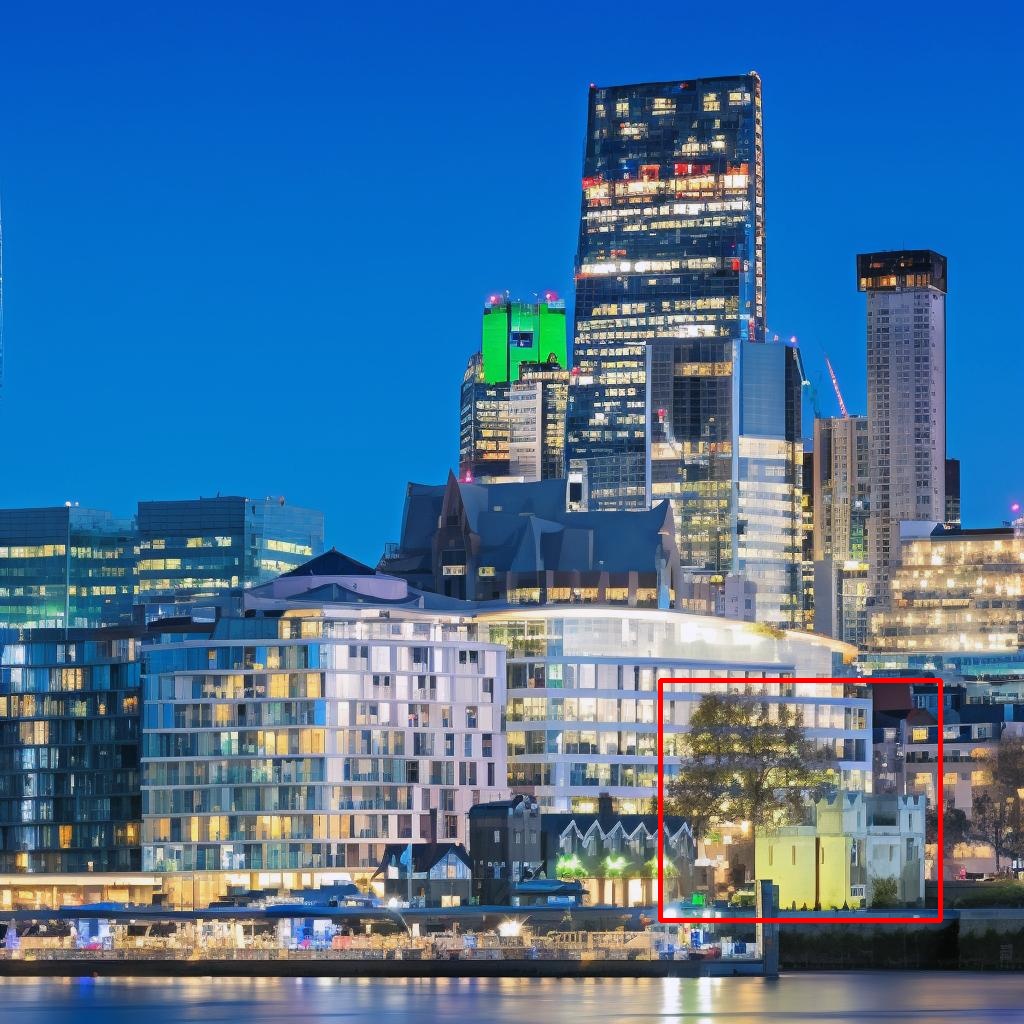}
                            \hspace{\g} &
                            \includegraphics[height=\h \textwidth, width=\w \textwidth]{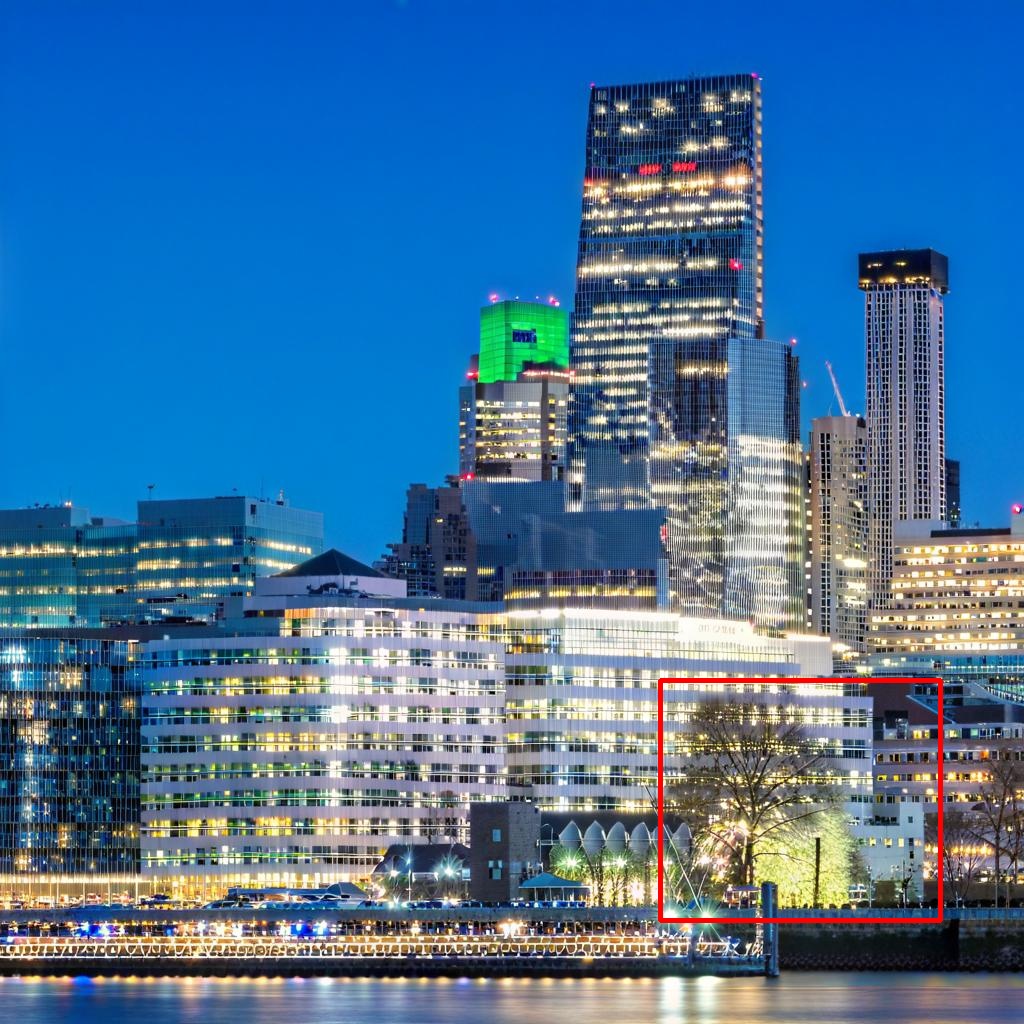}
                            \hspace{\g} &
                            \includegraphics[height=\h \textwidth, width=\w \textwidth]{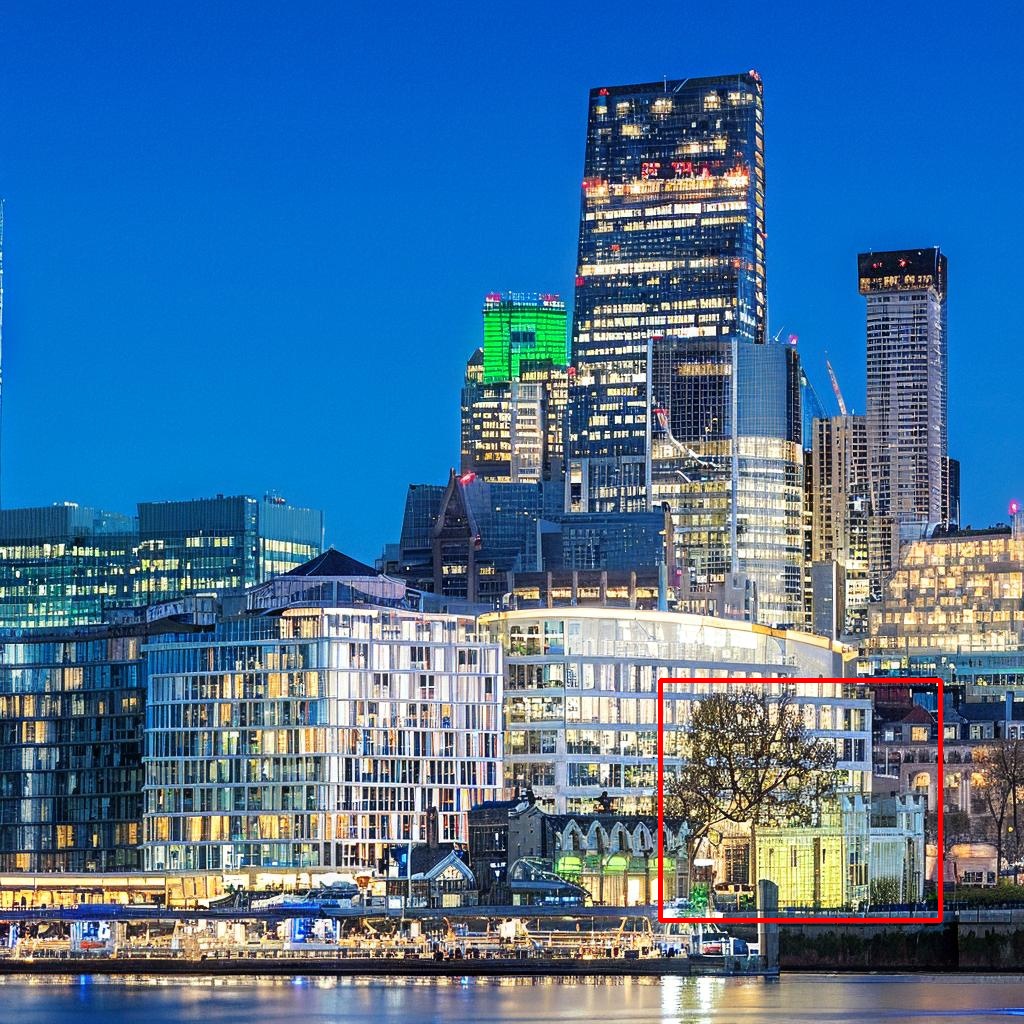}
                                    \vspace{-0.3mm}
						\\
                       						\includegraphics[height=\h \textwidth, width=\w \textwidth]{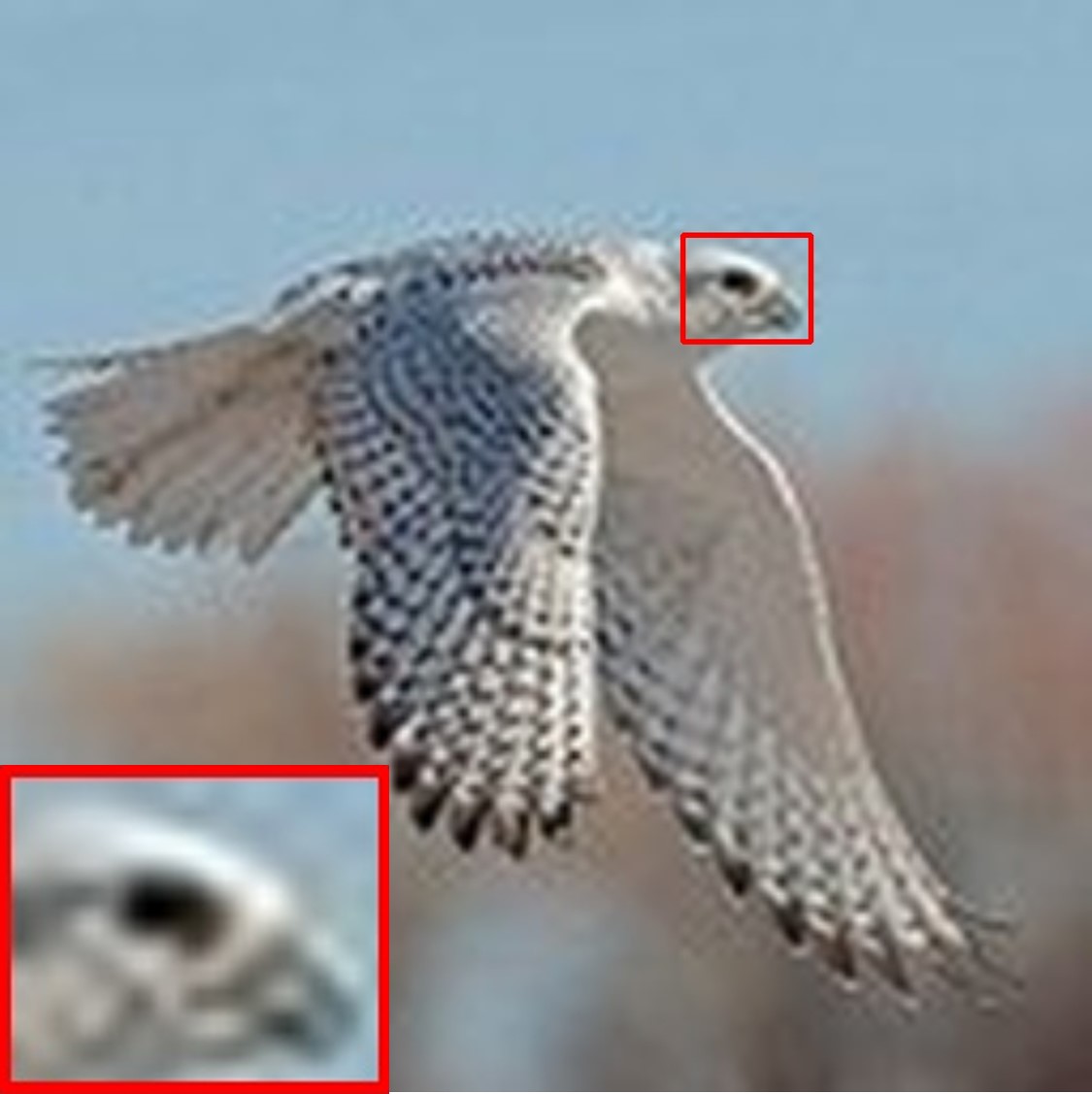} \hspace{\g} &
						\includegraphics[height=\h \textwidth, width=\w \textwidth]{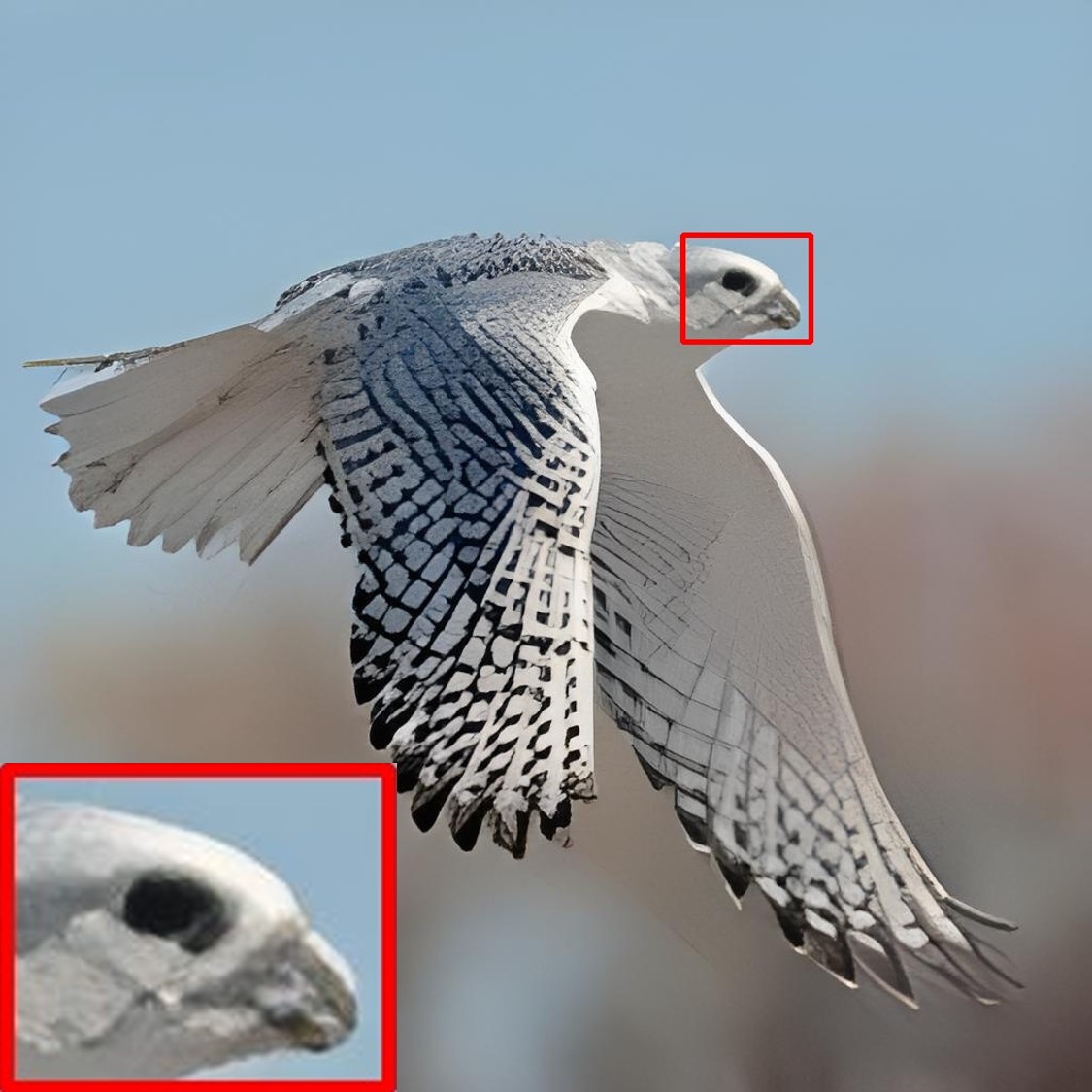} \hspace{\g} &
						\includegraphics[height=\h \textwidth, width=\w \textwidth]{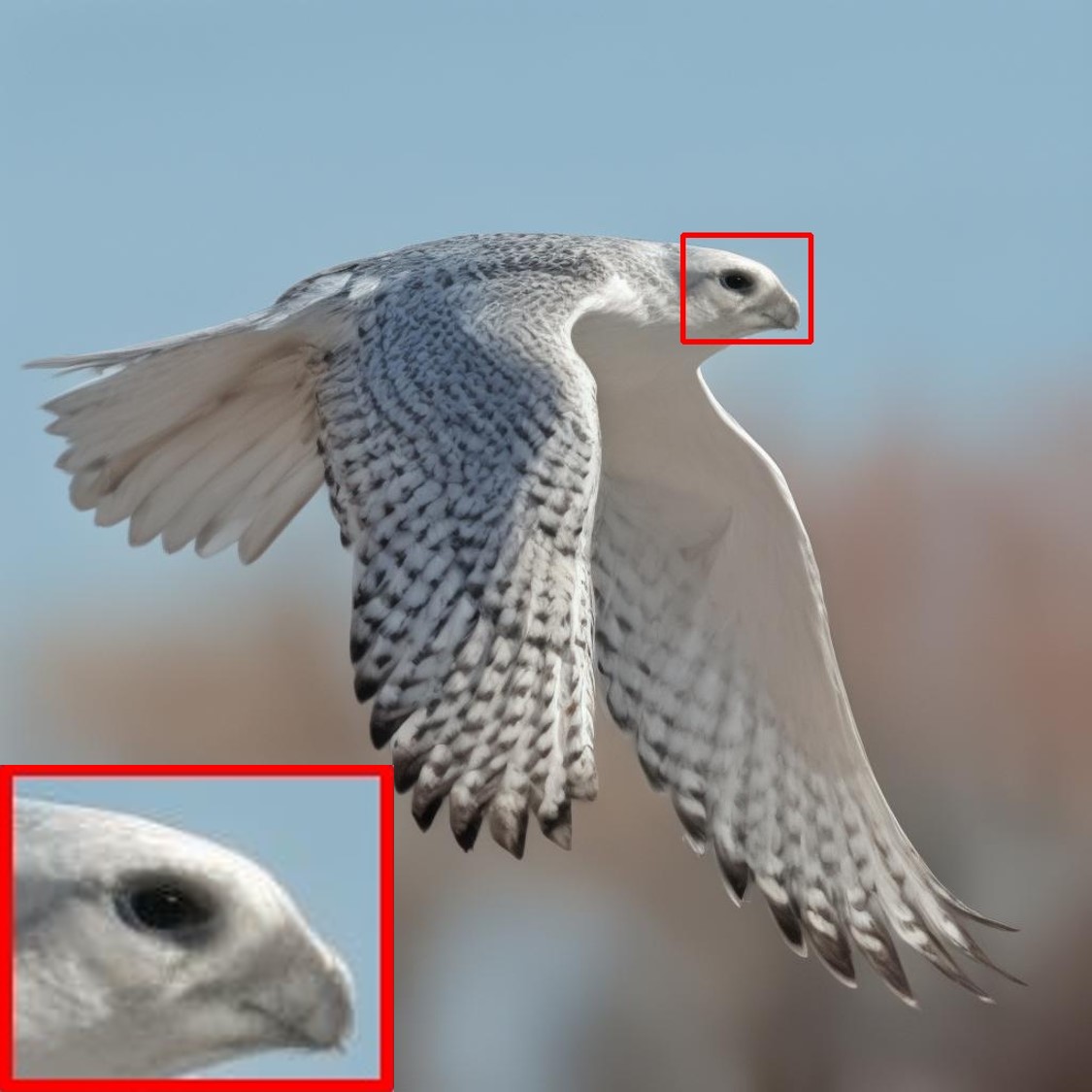} \hspace{\g} &
						\includegraphics[height=\h \textwidth, width=\w \textwidth]{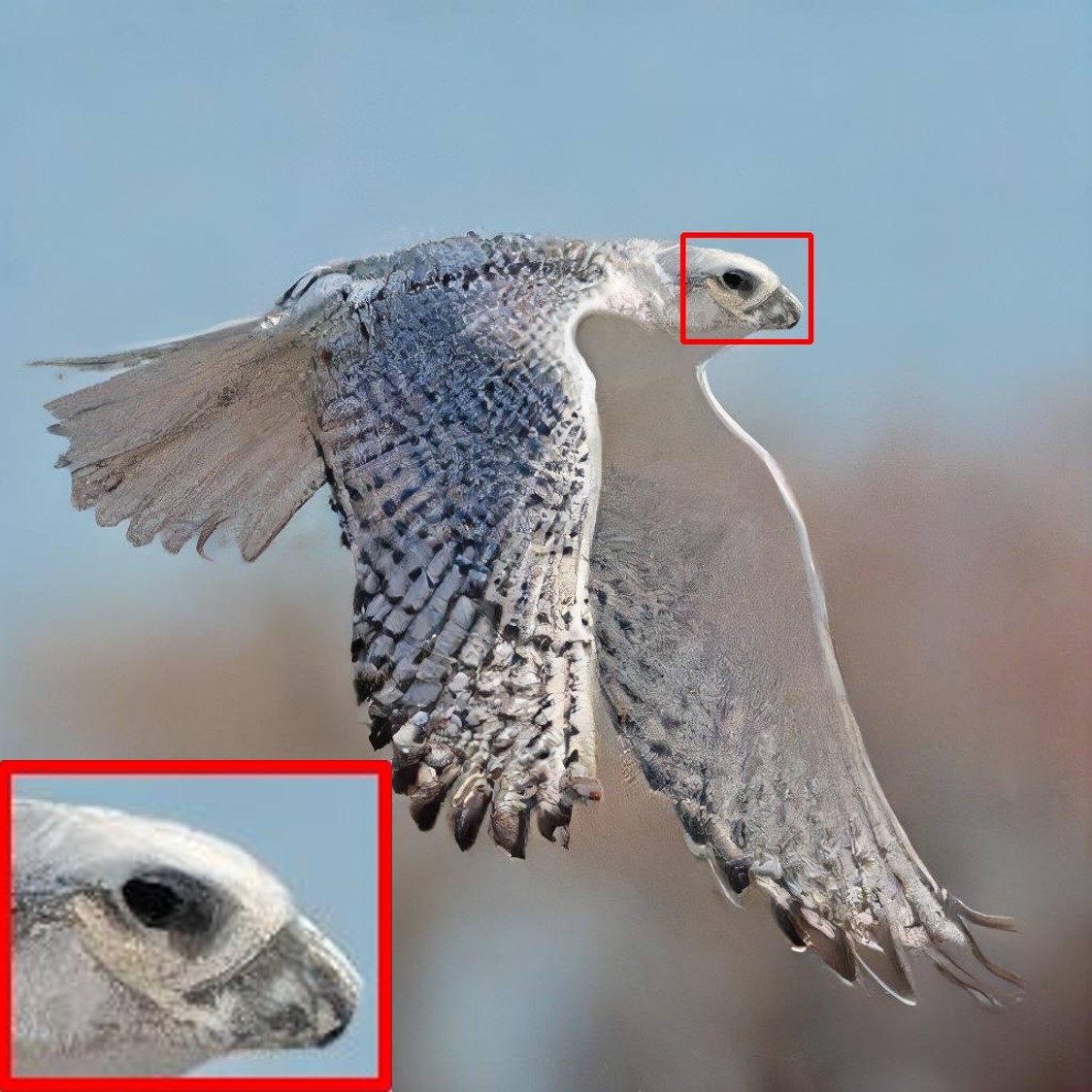} \hspace{\g} &
                            \includegraphics[height=\h \textwidth, width=\w \textwidth]{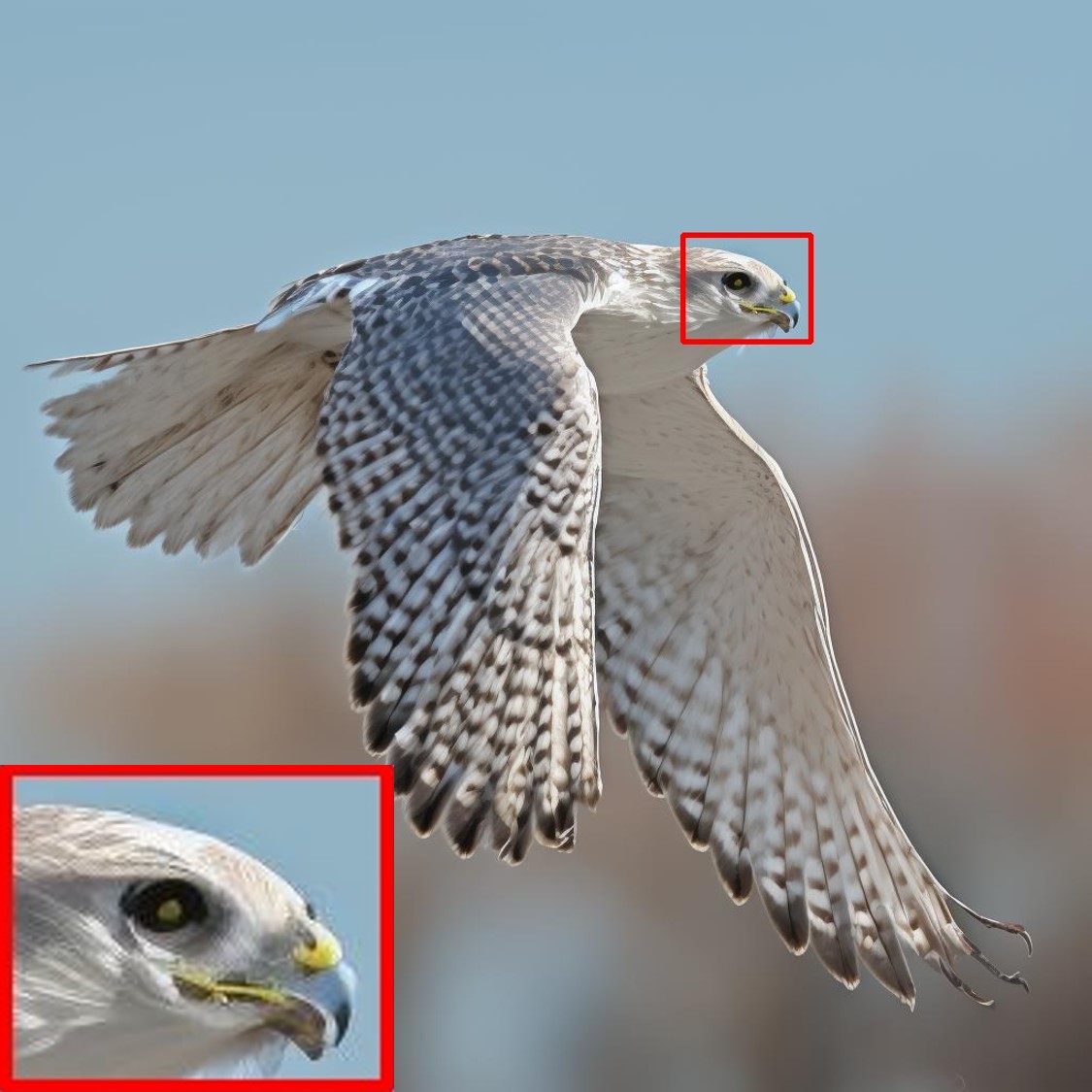}
                            \hspace{\g} &
                            \includegraphics[height=\h \textwidth, width=\w \textwidth]{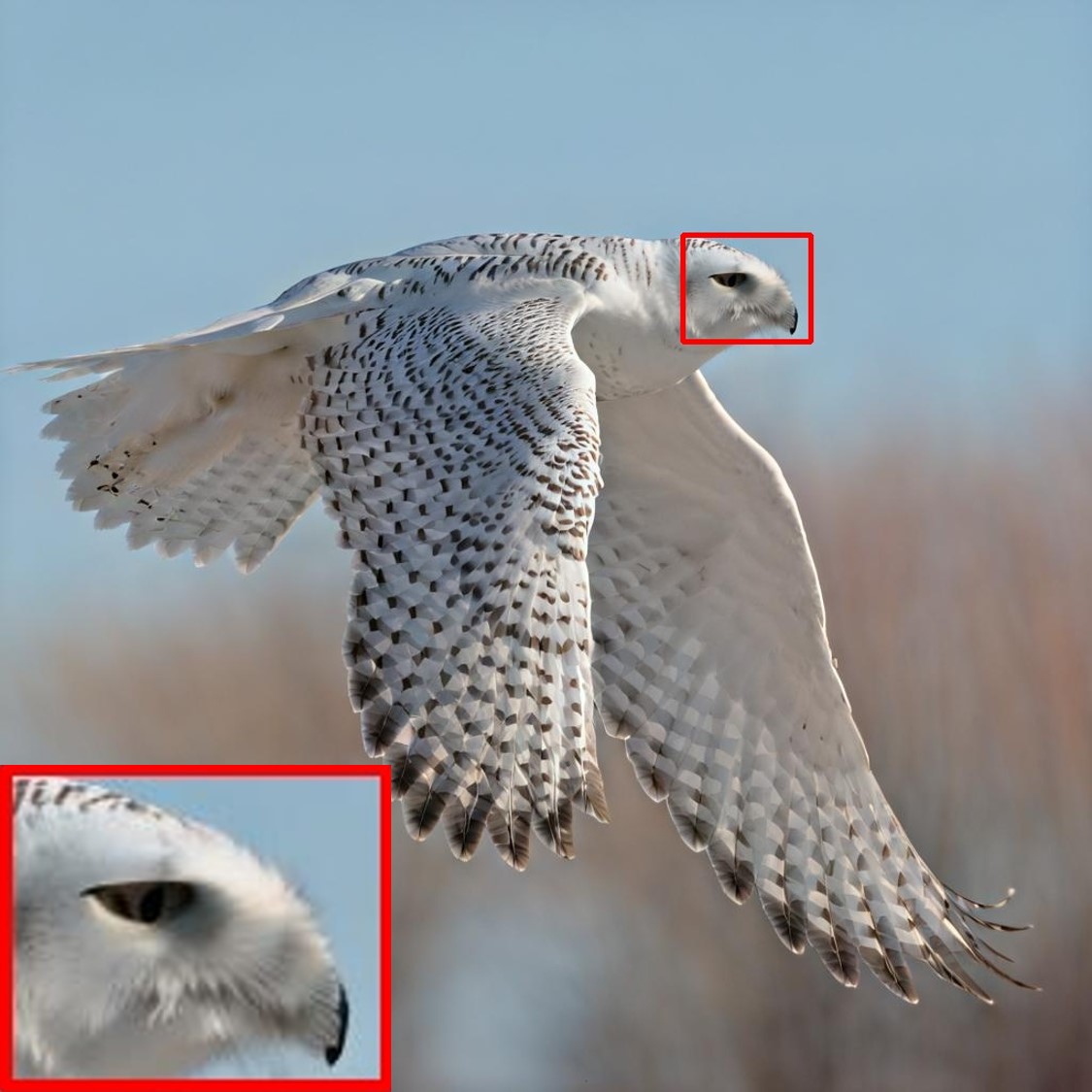}
                            \hspace{\g} &
                            \includegraphics[height=\h \textwidth, width=\w \textwidth]{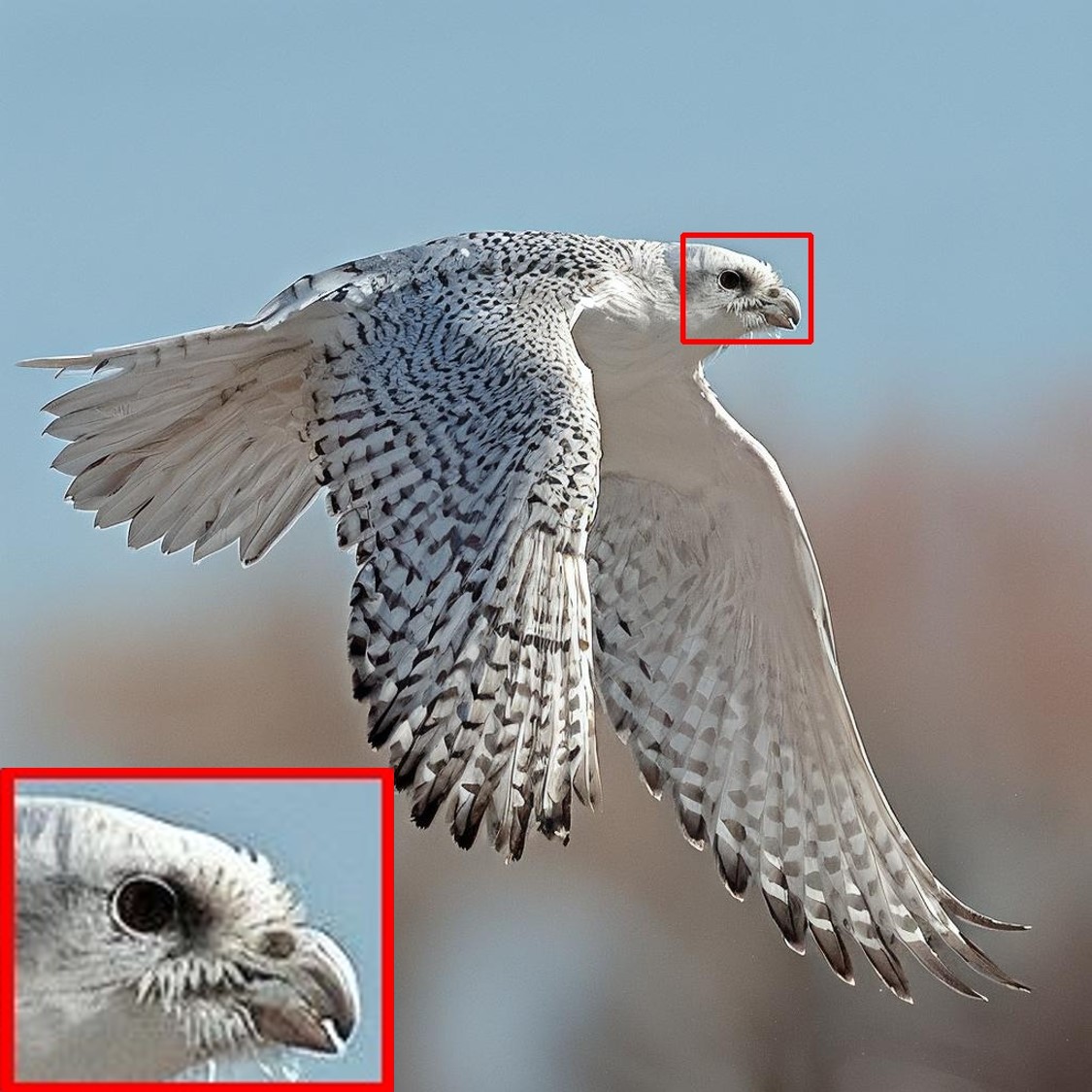}
						\\
						  LQ Input \hspace{\g} &
						Real-ESRGAN~\cite{realesrgan}  \hspace{\g} &
						DiffBIR~\cite{diffbir}  \hspace{\g}& 
      					SinSR~\cite{sinsr} \hspace{\g}& 
                            SeeSR~\cite{seesr} \hspace{\g}&
                            SUPIR~\cite{supir} \hspace{\g}&
                            \textbf{\modelname{}} (Ours)
						\\
					\end{tabular}
				\end{adjustbox}
        }
    	\caption{Qualitative comparisons on both synthetic (the first row) and real-world (the last two rows) benchmarks. Please zoom in for a better view.}
	\label{fig:sota_compare}
\end{figure}

%% file: figs/tex/userstudy.tex
\begin{figure}[th]  
\centering
  \includegraphics[width=0.98\linewidth]{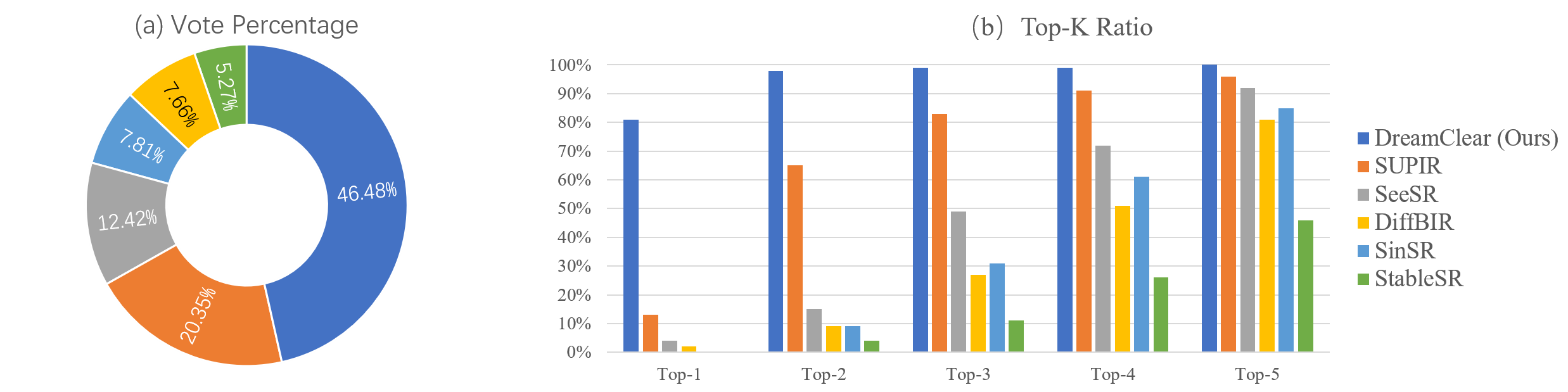}
    \caption{User study. Vote percentage denotes average user preference per model. The Top-K ratio indicates the proportion of images preferred by top K users. Our model is highly preferred, with most images being rated as top quality by the majority. }
    \label{fig:user-study}
\end{figure}

%% file: figs/tex/abla_data.tex
\begin{figure}[!t]
\captionsetup{font=small}
\begin{center}
\subfigure[\hspace{-0.cm}]{\includegraphics[width=0.26\textwidth]{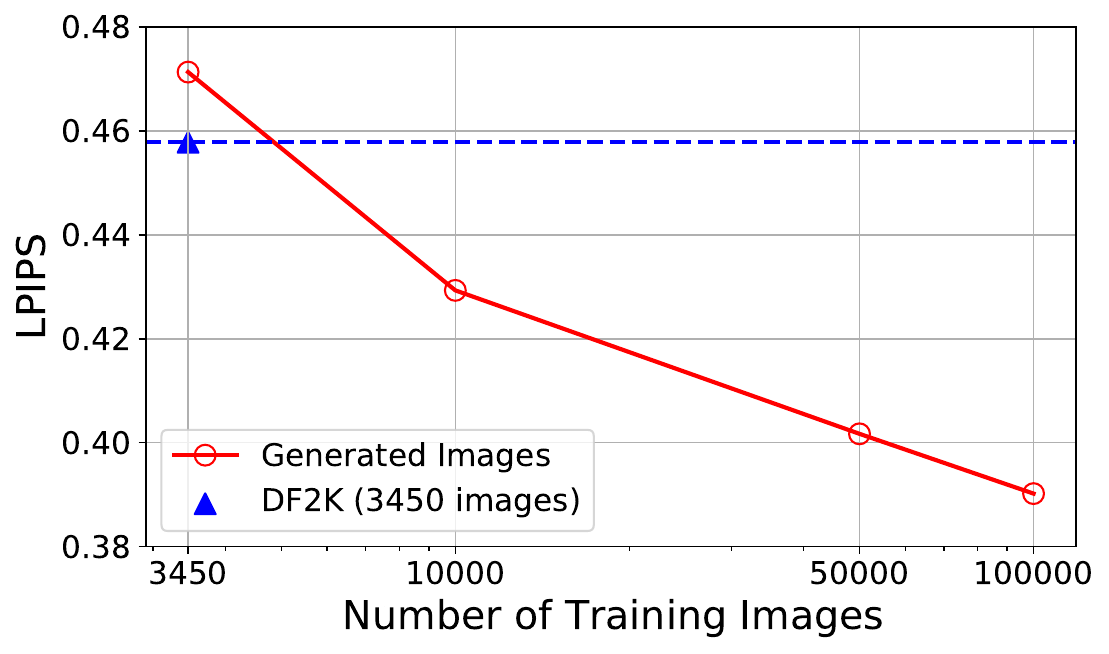}}\hspace{0.05\textwidth}\vspace{-0.1cm}
\subfigure[\hspace{-0.5cm}]{\includegraphics[width=0.26\textwidth]{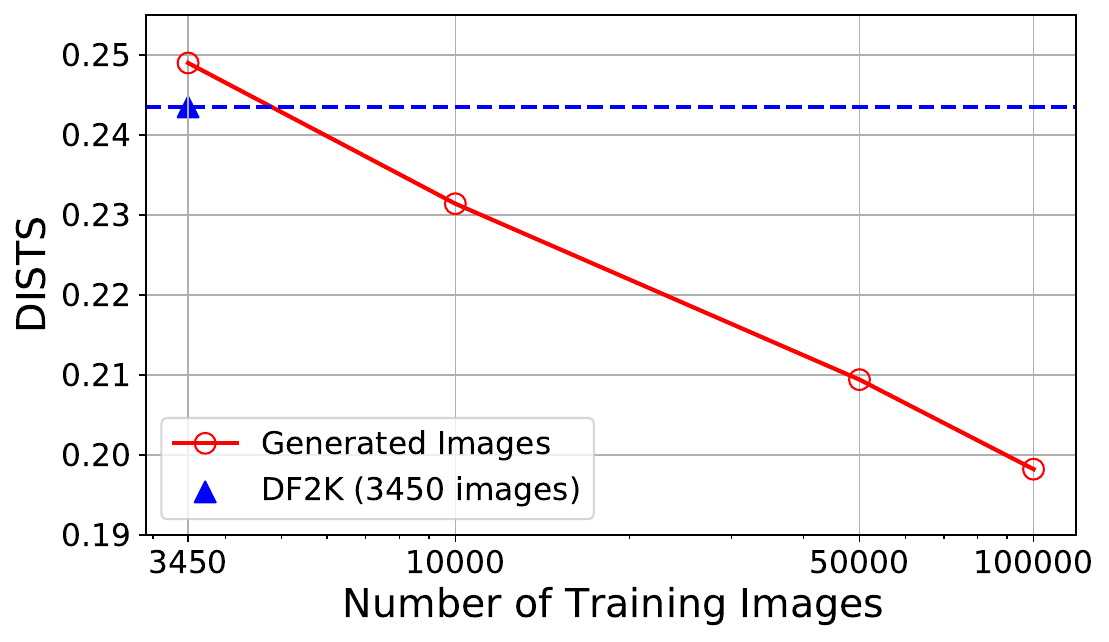}} \hspace{0.05\textwidth}\vspace{-0.1cm}
\subfigure[\hspace{-0.5cm}]{\includegraphics[width=0.26\textwidth]{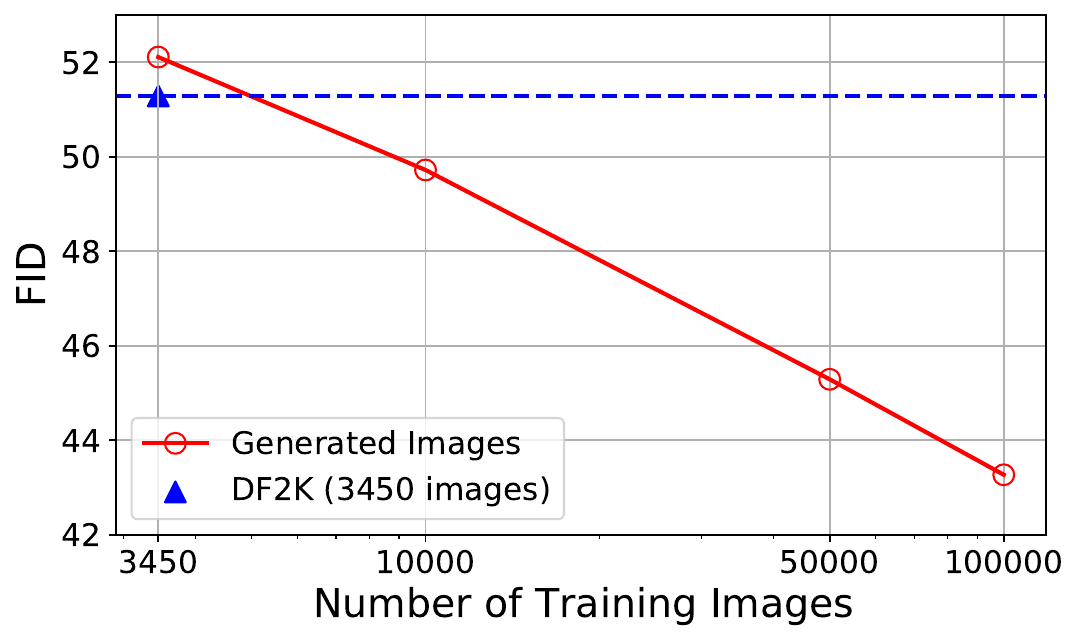}} \vspace{-0.1cm}
\subfigure[\hspace{-0.cm}]{\includegraphics[width=0.26\textwidth]{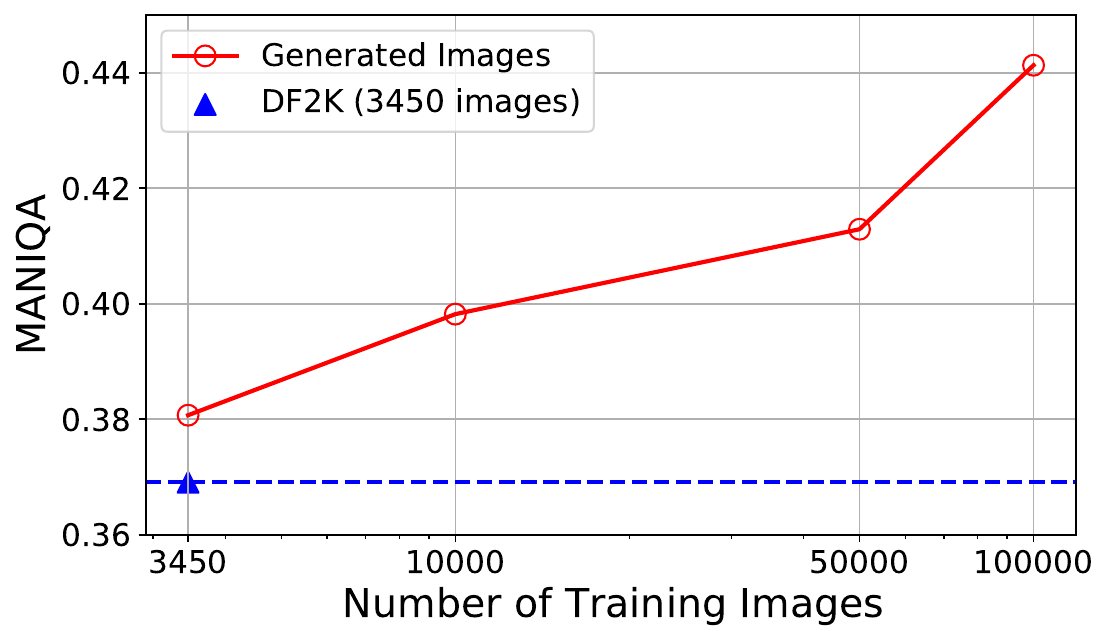}}\hspace{0.05\textwidth}
\subfigure[\hspace{-0.5cm}]{\includegraphics[width=0.26\textwidth]{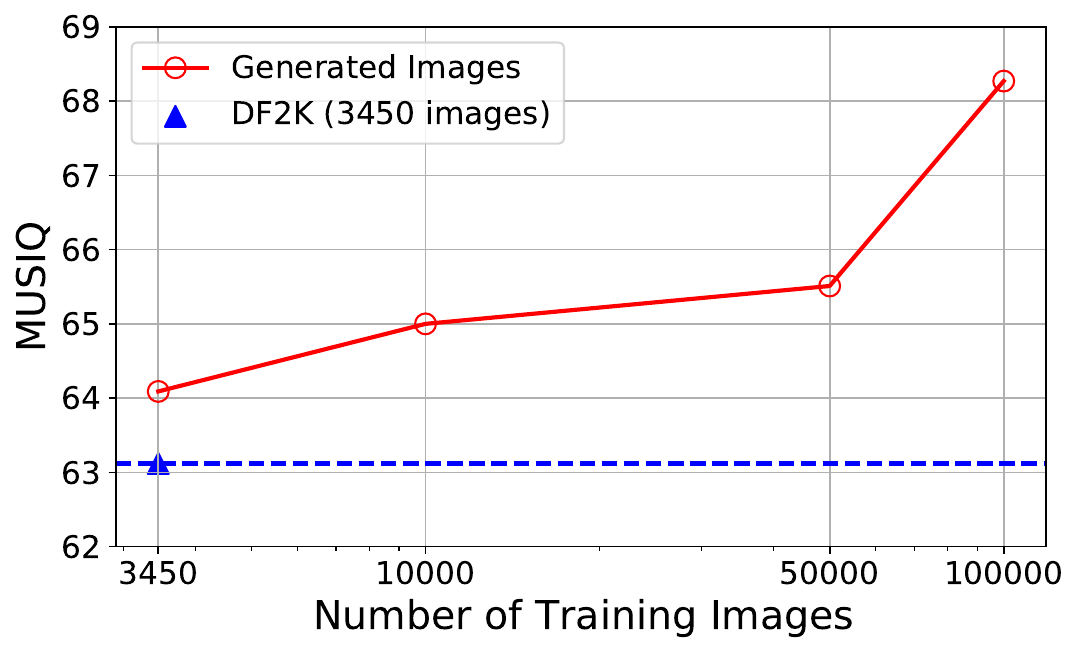}} \hspace{0.05\textwidth}
\subfigure[\hspace{-0.5cm}]{\includegraphics[width=0.26\textwidth]{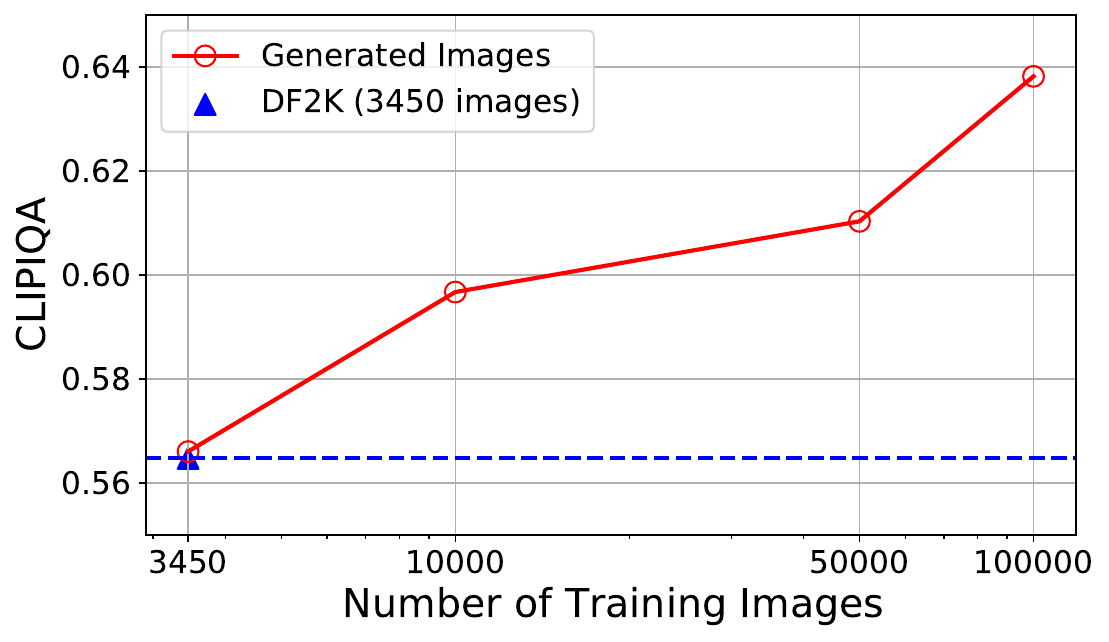}} 
\end{center}\vspace{-0.2cm}
\caption{Impact of synthetic training data. As data size increases, performance improves on \textit{LSDIR-Val}.}
\label{fig:ablation:datascaling}
\end{figure}

%% file: tables/abla_model.tex
\begin{table}[t]
\captionsetup{font=small}%
\scriptsize
\center
\caption{Ablation results on \textit{DIV2K-Val}, COCO val2017 and ADE20K for \modelname{}.} 
\vspace{0.2cm}
\setlength\tabcolsep{5pt}
\begin{center}
\resizebox{1.0\linewidth}{!}{
\begin{tabular}{l|ccc|ccc|ccc}
    \toprule
          & LPIPS $\downarrow$ & DISTS $\downarrow$ & FID $\downarrow$ & MANIQA $\uparrow$ & MUSIQ $\uparrow$ & CLIPIQA $\uparrow$ & $AP^b$ & $AP^m$ & mIoU
         \\ \midrule
         Mixture of AM & \textbf{0.3657} & \textbf{0.1637} & \textbf{20.61} &\textbf{0.4320}&\textbf{68.44}& \textbf{0.6963} & \textbf{19.3} & \textbf{16.7} & \textbf{31.9} \\
         AM & 0.3981&	0.1843&	25.75&	0.4067&	66.18&	0.6646 & 18.0&	15.6&	28.6\\
        Cross-Attention & 0.4177&	0.2016&	29.74&	0.3785&	63.21&	0.6497 & 17.2&	15.1&	26.3\\
        Zero-Linear & 0.4082 &	0.1976&	29.89&	0.4122&	66.11&	0.6673 & 17.6&	15.3&	27.2\\
        \midrule
        Dual-Branch &\textbf{0.3657} & \textbf{0.1637} & \textbf{20.61} &\textbf{0.4320}&\textbf{68.44}& \textbf{0.6963} & \textbf{19.3} & \textbf{16.7} & \textbf{31.9}\\
        w/o Reference Branch & 0.4207	&0.2033&30.91&	0.3985&	64.04&	0.6582 &15.9&	14.0&	24.7\\
        \midrule
        Detailed Text Prompt&0.3657 & 0.1637 & 20.61 &\textbf{0.4320}&\textbf{68.44}& \textbf{0.6963} & \textbf{19.3} & \textbf{16.7} & \textbf{31.9}\\
        Null Prompt & \textbf{0.3521} &	\textbf{0.1607}& \textbf{20.47}&	0.4230&	67.26&	0.6812 & 18.8&	16.2&	29.8\\
         \bottomrule
    \end{tabular}}
\end{center}
    \label{tab:ablation_model}
\end{table}

%% file: tables_appendix/abla_genir.tex
\begin{table}[t]
\captionsetup{font=small}%
\scriptsize
\center
\caption{Ablations for GenIR on \textit{LSDIR-Val} using SwinIR-GAN.} 
\vspace{0.2cm}
\begin{center}
\resizebox{1.0\linewidth}{!}{
\begin{tabular}{l|ccc|ccc}
    \toprule
          Training Data& LPIPS $\downarrow$ & DISTS $\downarrow$ & FID $\downarrow$ & MANIQA $\uparrow$ & MUSIQ $\uparrow$ & CLIPIQA $\uparrow$
         \\ \midrule
         Pre-trained T2I Model (3450images) & 0.4819 &	0.2790&	60.12&	0.3271&	61.94&	0.5423 \\ 
     Ours GenIR (3450images) & \textbf{0.4578}&	\textbf{0.2435}&\textbf{51.29}&	\textbf{0.3691}&\textbf{63.12}&\textbf{0.5647} \\ 
         \bottomrule
    \end{tabular}}
\end{center}
    \label{tab:ablation_genir}
\end{table}

%% file: figs_appendix/tex/visual_1.tex
\begin{figure}[!htbp]
	\scriptsize
	\centering
	\newcommand{\h}{0.105}
	\newcommand{\wa}{0.12}
	\newcommand{\wb}{0.16}
	\newcommand{\g}{-0.7mm}
 	\setlength\tabcolsep{1.5pt}
	\renewcommand{\arraystretch}{1}
	\resizebox{1.00\linewidth}{!} {
			\renewcommand{\h}{0.186}
			\newcommand{\w}{0.186}
				\begin{adjustbox}{valign=t}
					\begin{tabular}{ccc}
						\includegraphics[height=\h \textwidth, width=\w \textwidth]{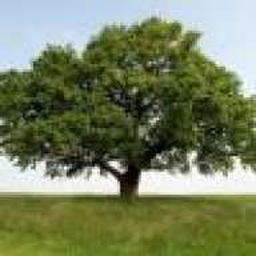} \hspace{\g} &
						\includegraphics[height=\h \textwidth, width=\w \textwidth]{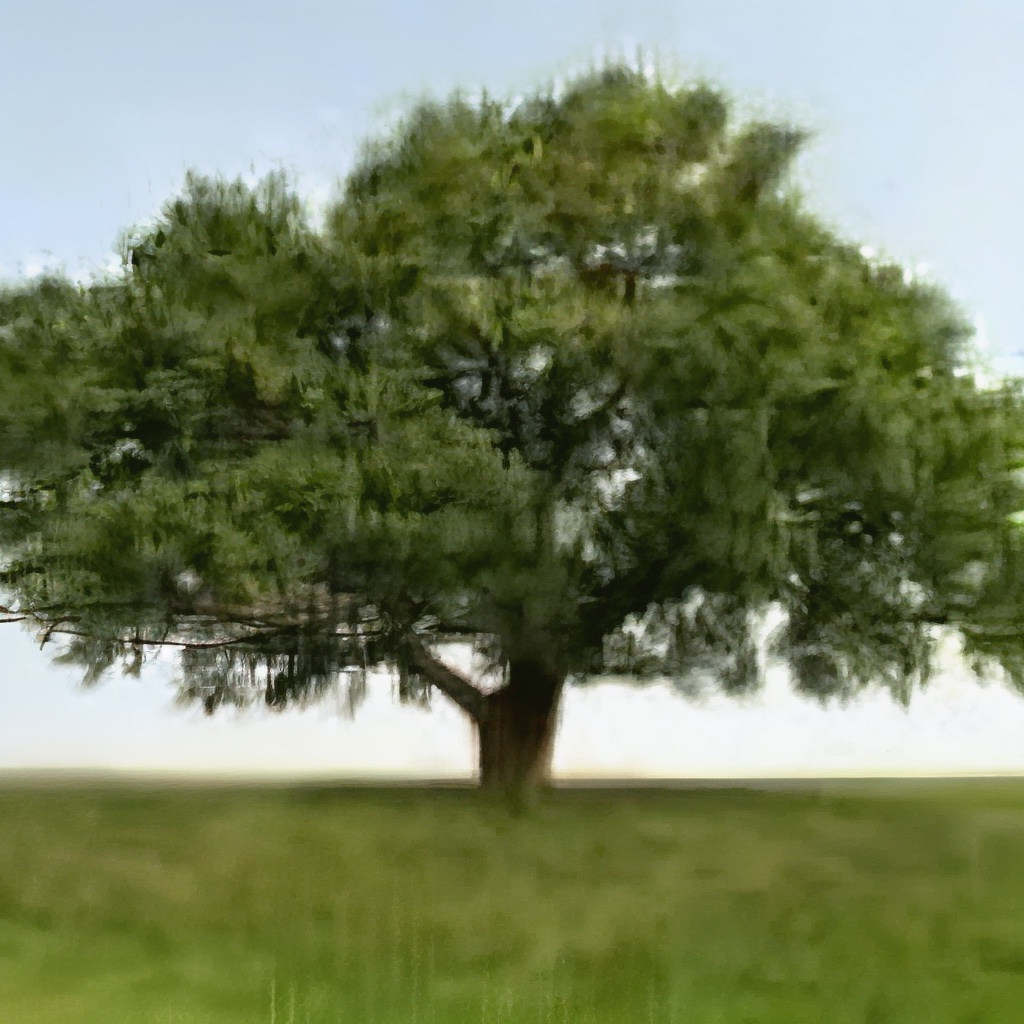} \hspace{\g} &
						\includegraphics[height=\h \textwidth, width=\w \textwidth]{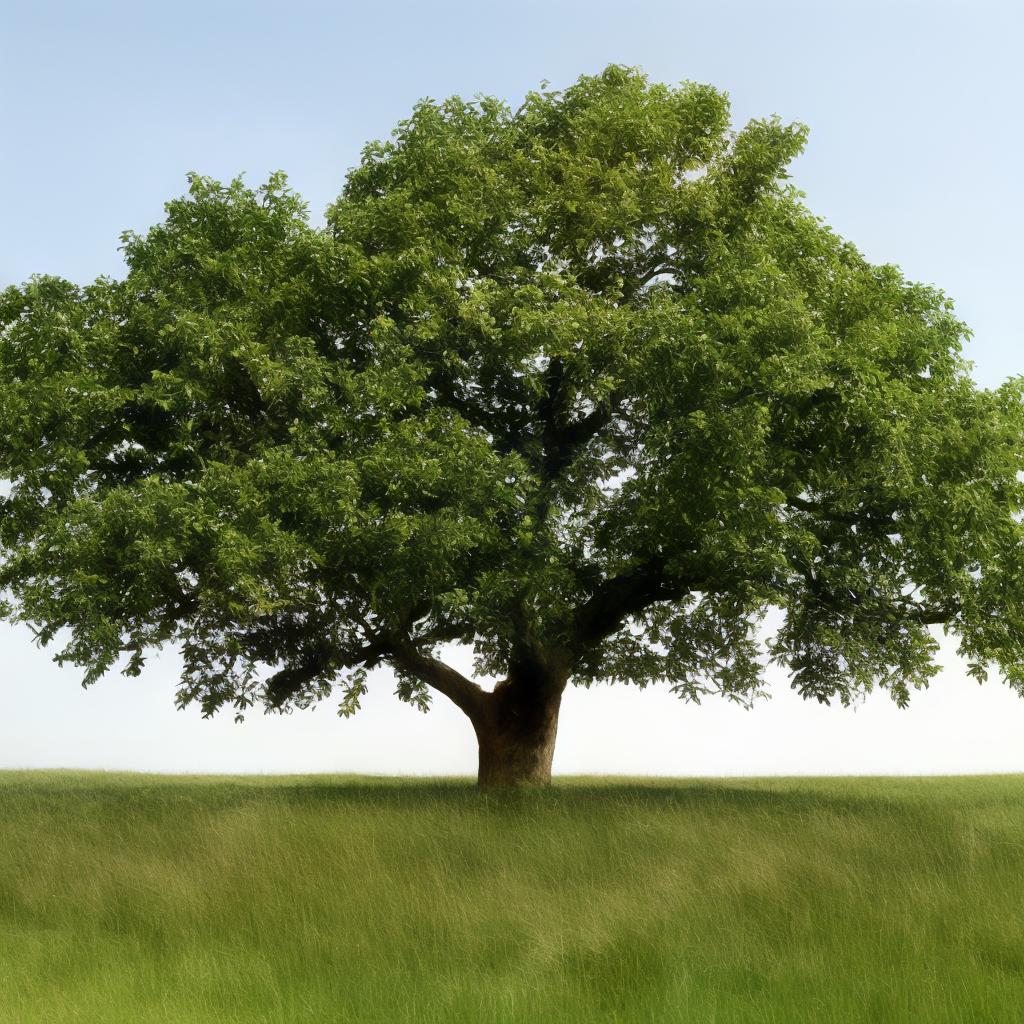} 
						\\
      					LQ Input \hspace{\g} &
						StableSR~\cite{stablesr}  \hspace{\g} &
						DiffBIR~\cite{diffbir}  
						\\
      						\includegraphics[height=\h \textwidth, width=\w \textwidth]{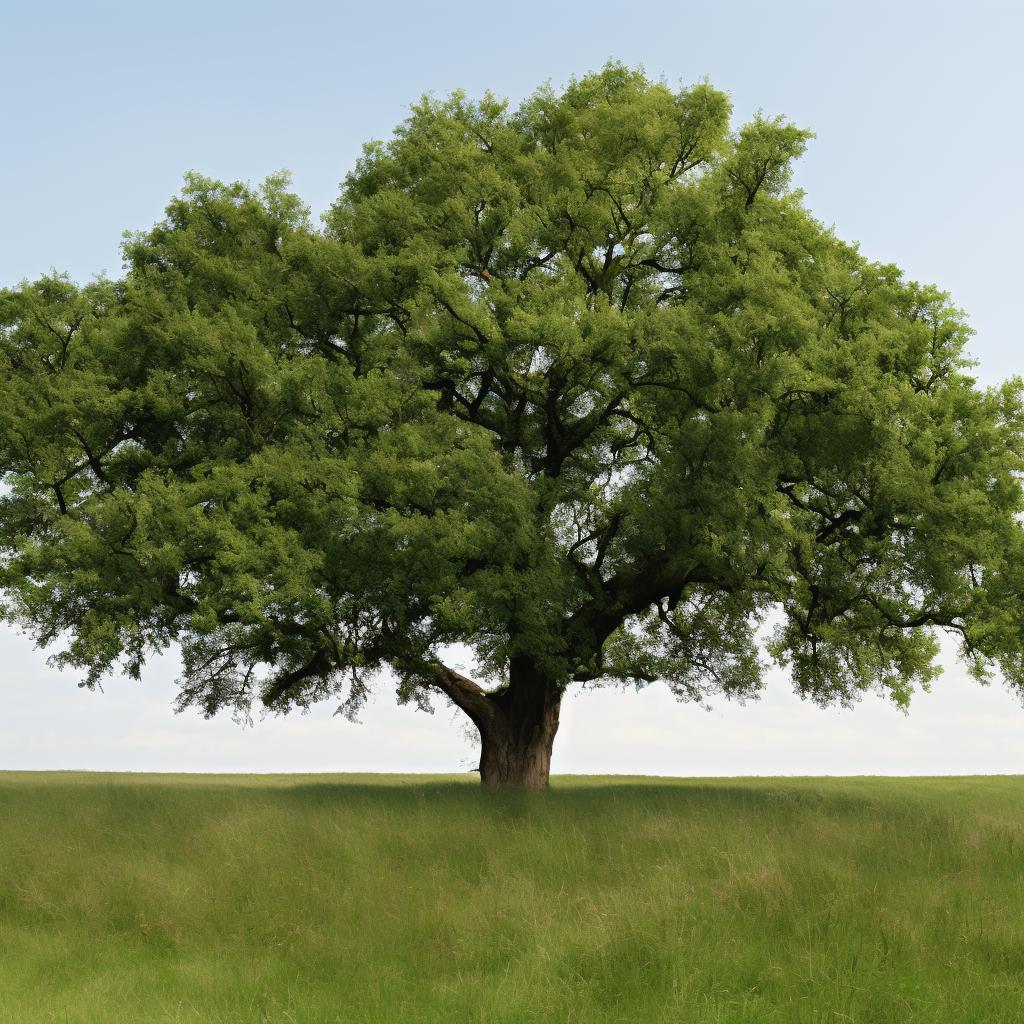} \hspace{\g} &
						\includegraphics[height=\h \textwidth, width=\w \textwidth]{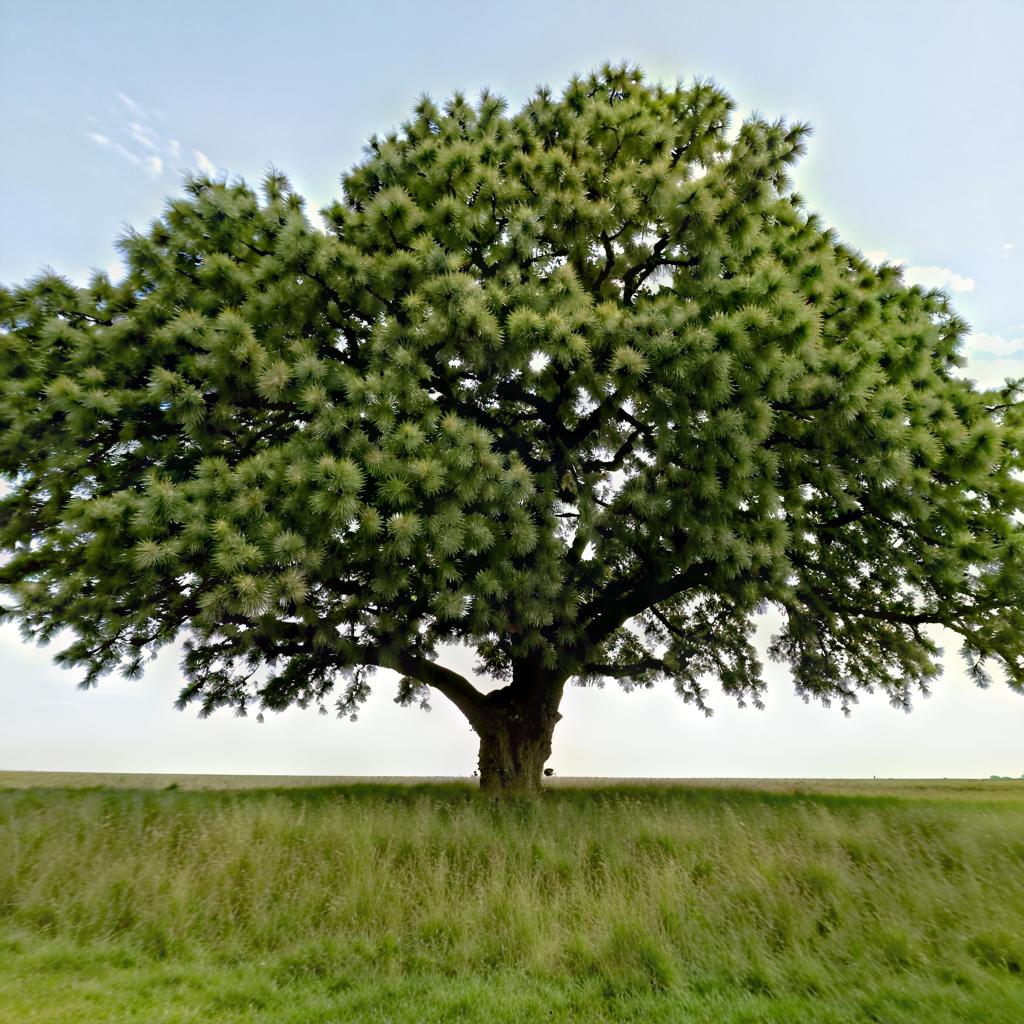} \hspace{\g} &
						\includegraphics[height=\h \textwidth, width=\w \textwidth]{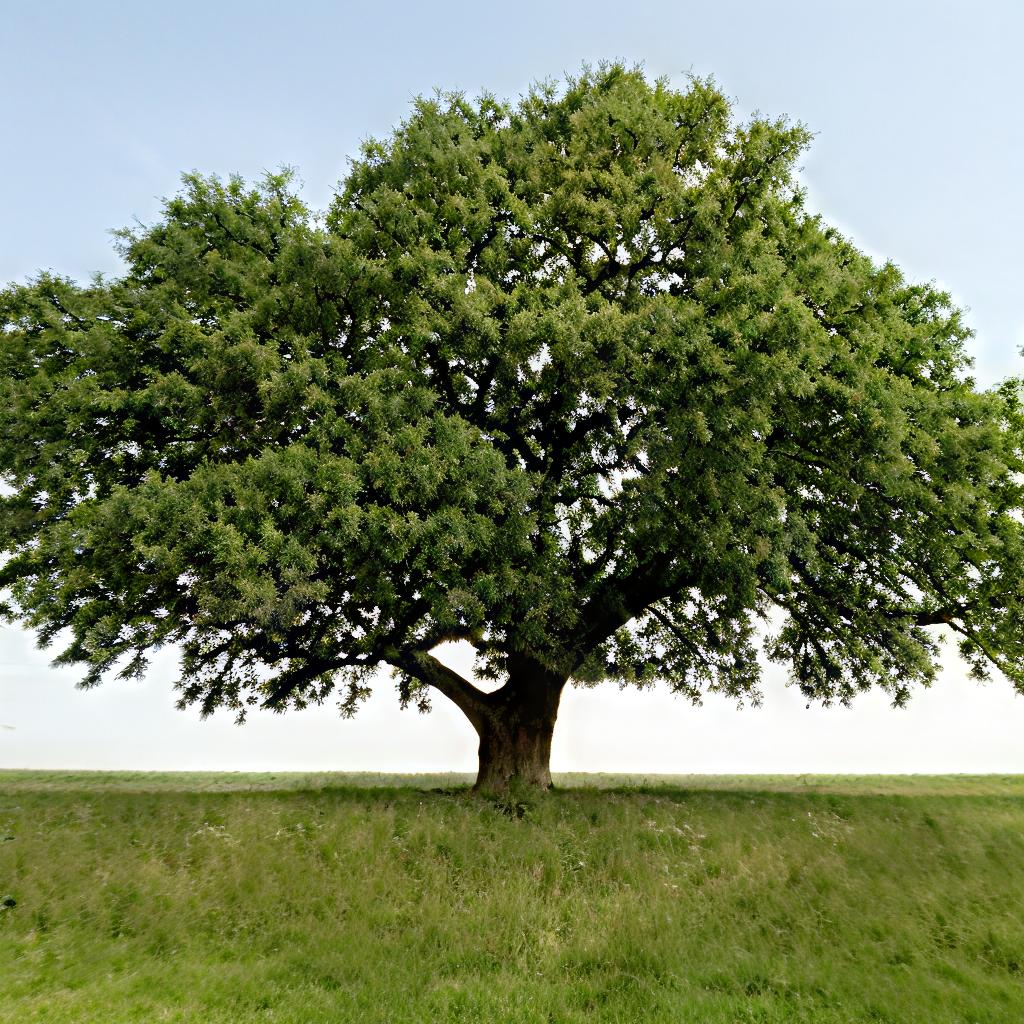} 
						\\
      					SeeSR~\cite{seesr} \hspace{\g}&
                            SUPIR~\cite{supir} \hspace{\g}&
						 \textbf{\modelname{}} (Ours)
                                          \vspace{1mm}

						\\
      						\includegraphics[height=\h \textwidth, width=\w \textwidth]{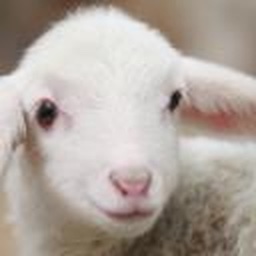} \hspace{\g} &
						\includegraphics[height=\h \textwidth, width=\w \textwidth]{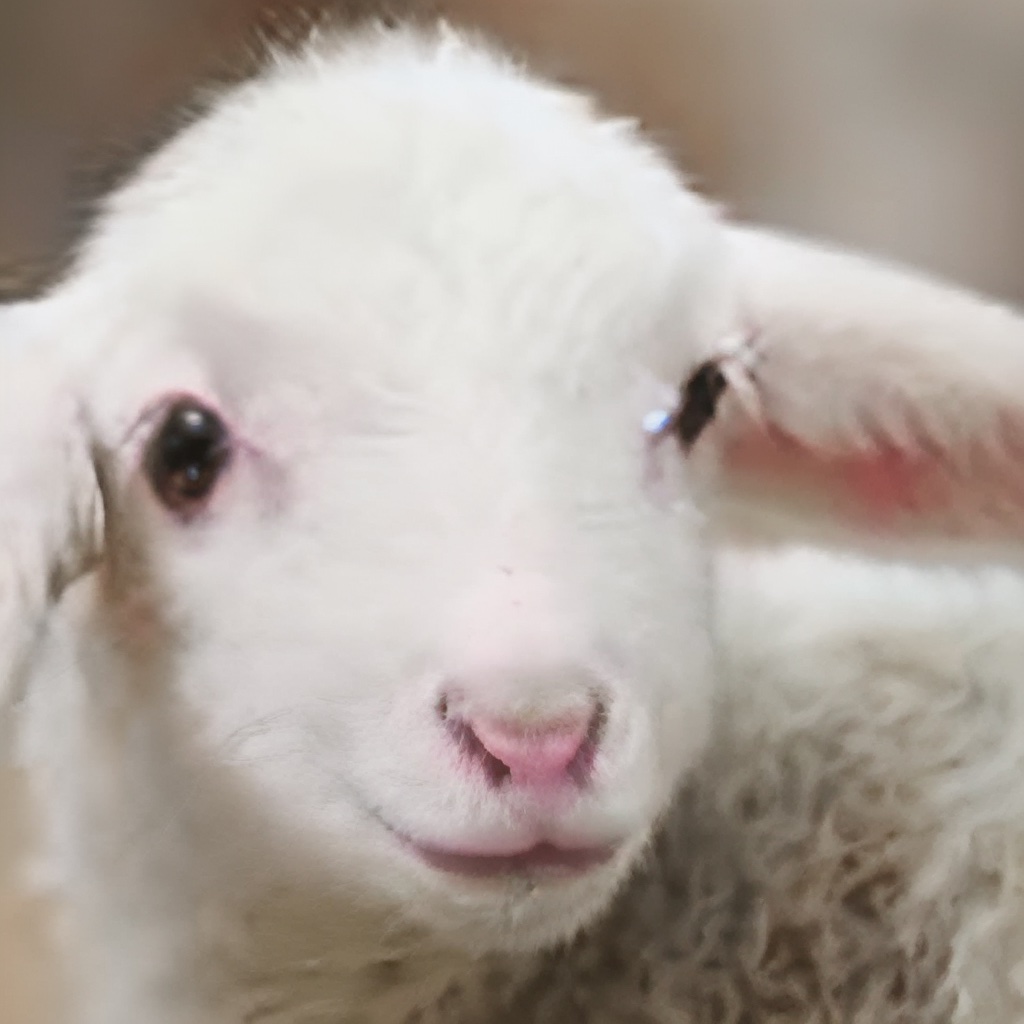} \hspace{\g} &
						\includegraphics[height=\h \textwidth, width=\w \textwidth]{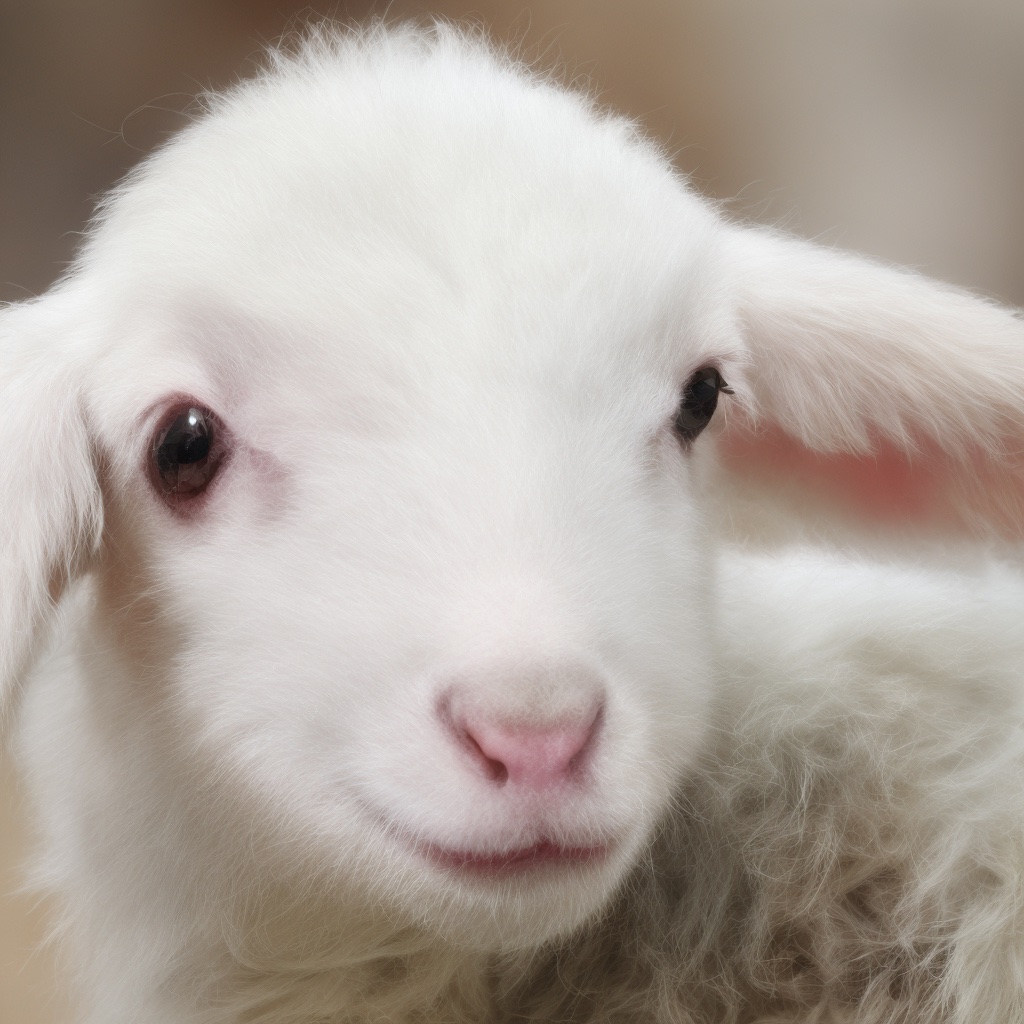} 
						\\
      					LQ Input \hspace{\g} &
						StableSR~\cite{stablesr}  \hspace{\g} &
						DiffBIR~\cite{diffbir}  
						\\
      						\includegraphics[height=\h \textwidth, width=\w \textwidth]{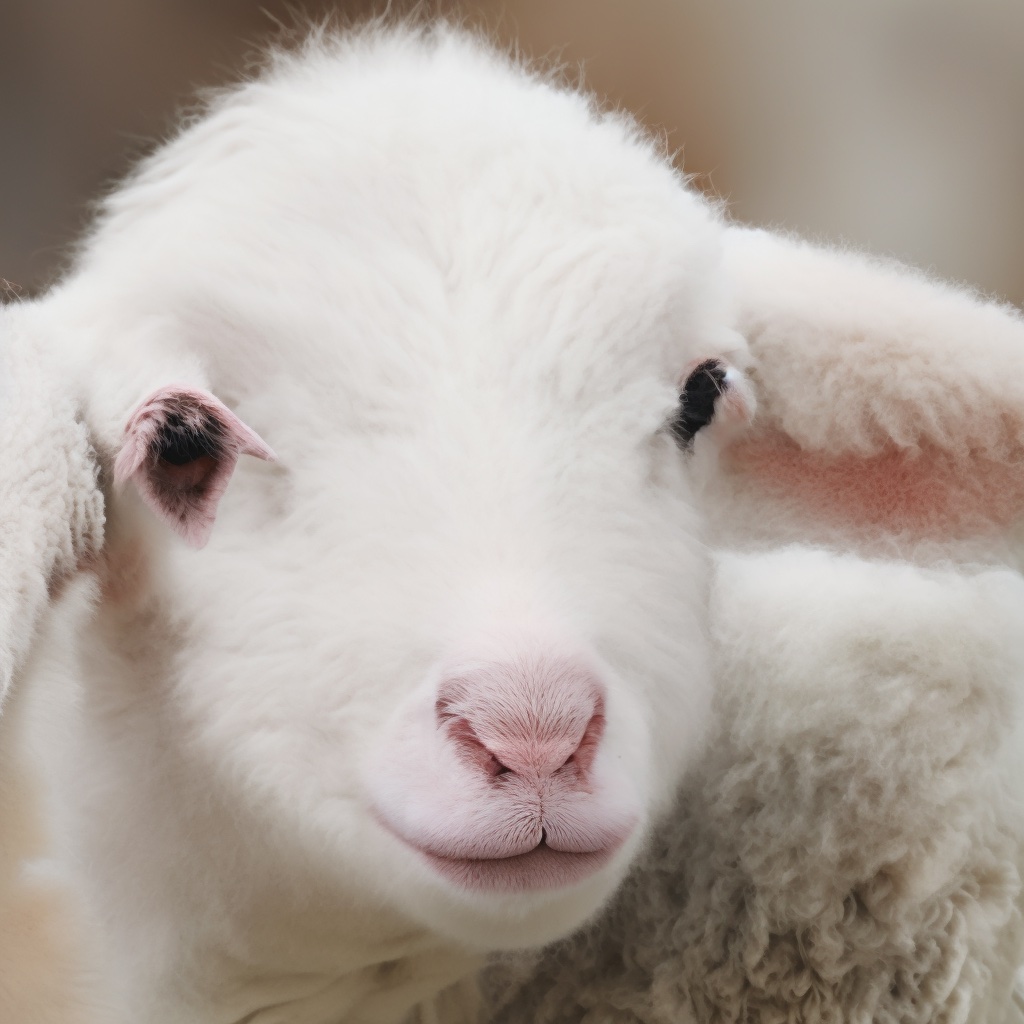} \hspace{\g} &
						\includegraphics[height=\h \textwidth, width=\w \textwidth]{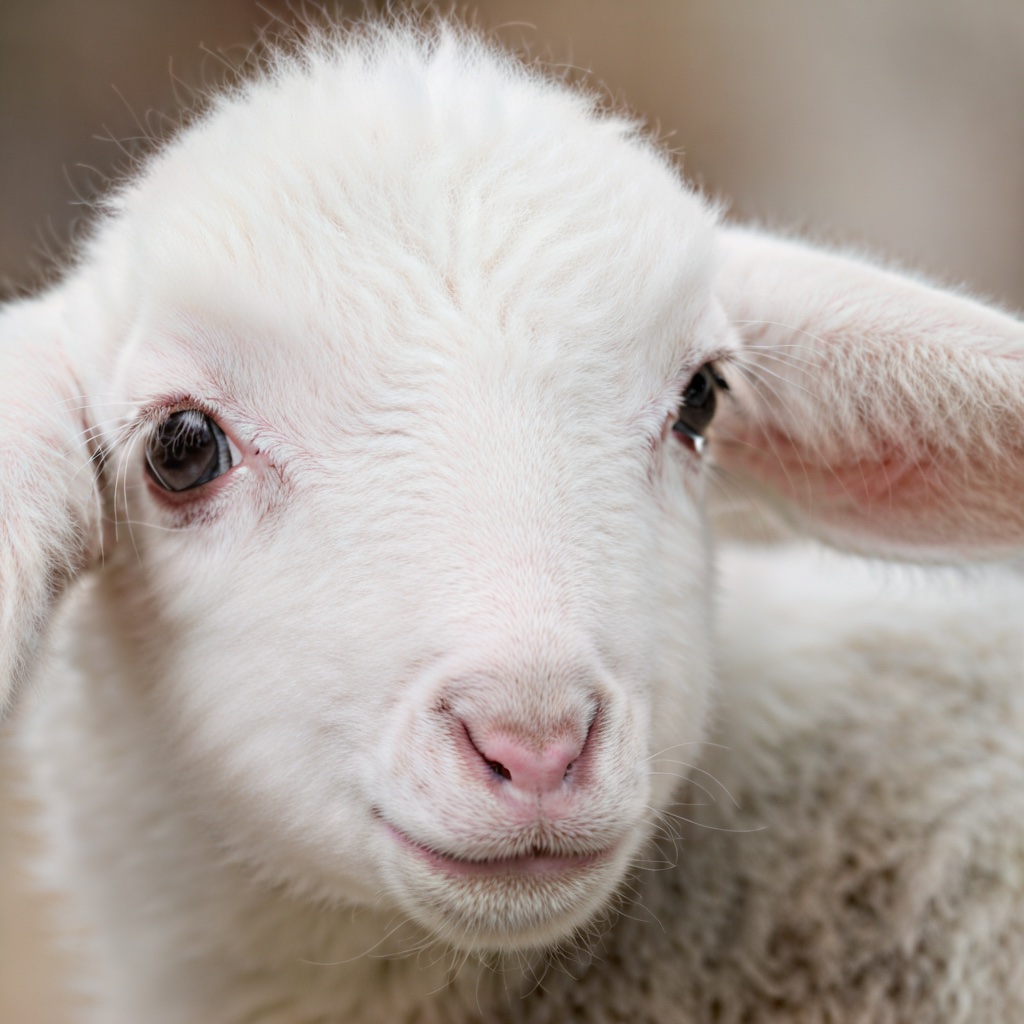} \hspace{\g} &
						\includegraphics[height=\h \textwidth, width=\w \textwidth]{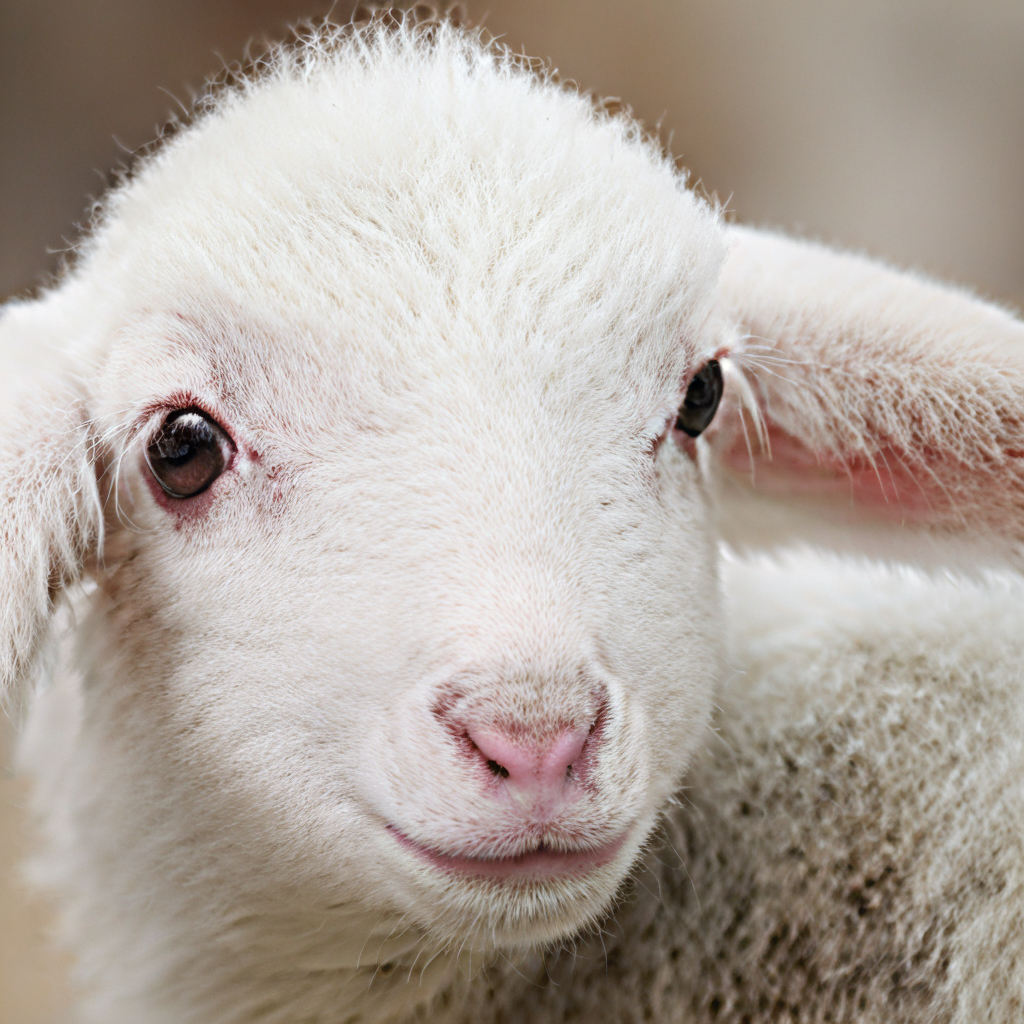} 
						\\
      					SeeSR~\cite{seesr} \hspace{\g}&
                            SUPIR~\cite{supir} \hspace{\g}&
						 \textbf{\modelname{}} (Ours)
						\\

					\end{tabular}
				\end{adjustbox}
        }
    	\caption{Visual comparisons on real-world benchmarks (1/3). Please zoom in for a better view.}
	\label{fig:visual_1}
\end{figure}

%% file: figs_appendix/tex/visual_2.tex
\begin{figure}[!htbp]
	\scriptsize
	\centering
	\newcommand{\h}{0.105}
	\newcommand{\wa}{0.12}
	\newcommand{\wb}{0.16}
	\newcommand{\g}{-0.7mm}
 	\setlength\tabcolsep{1.5pt}
	\renewcommand{\arraystretch}{1}
	\resizebox{1.00\linewidth}{!} {
			\renewcommand{\h}{0.186}
			\newcommand{\w}{0.186}
				\begin{adjustbox}{valign=t}
					\begin{tabular}{ccc}
						\includegraphics[height=\h \textwidth, width=\w \textwidth]{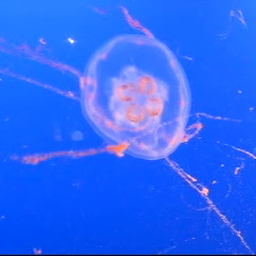} \hspace{\g} &
						\includegraphics[height=\h \textwidth, width=\w \textwidth]{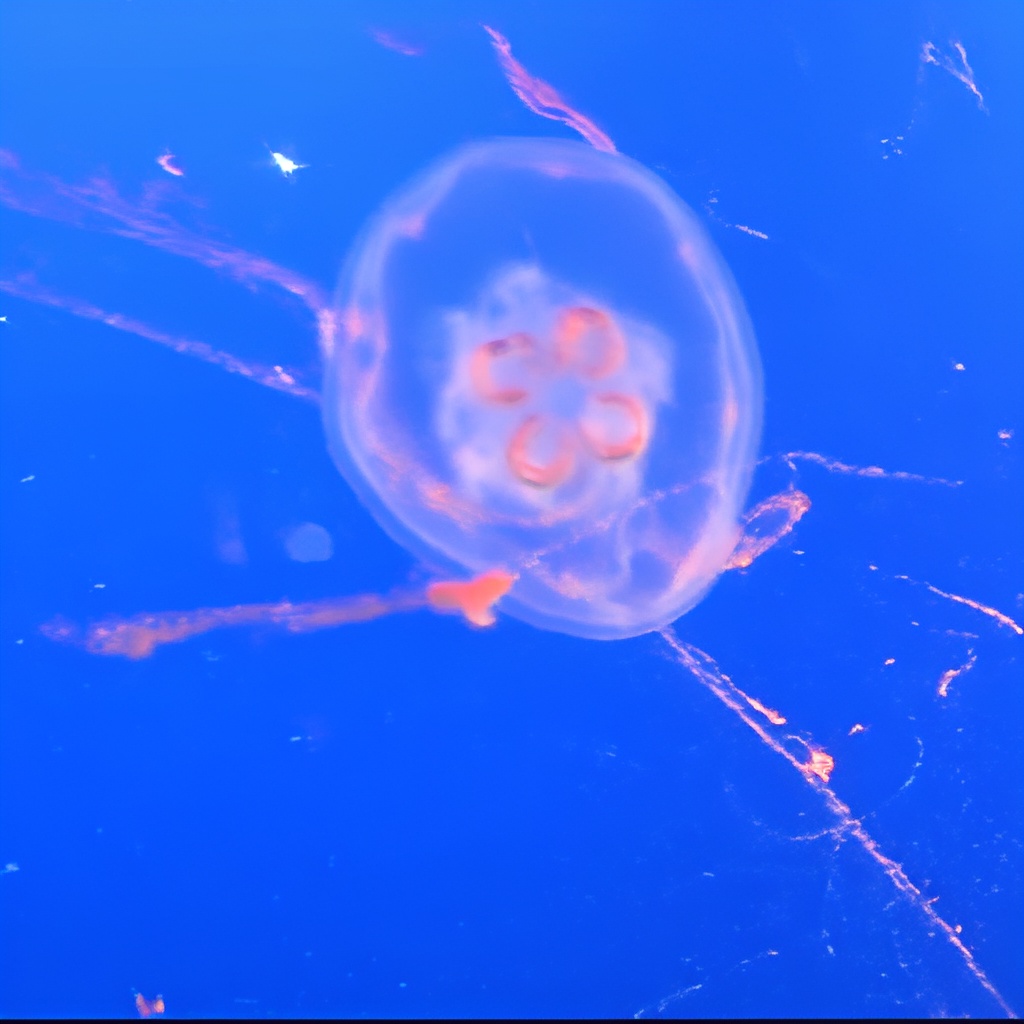} \hspace{\g} &
						\includegraphics[height=\h \textwidth, width=\w \textwidth]{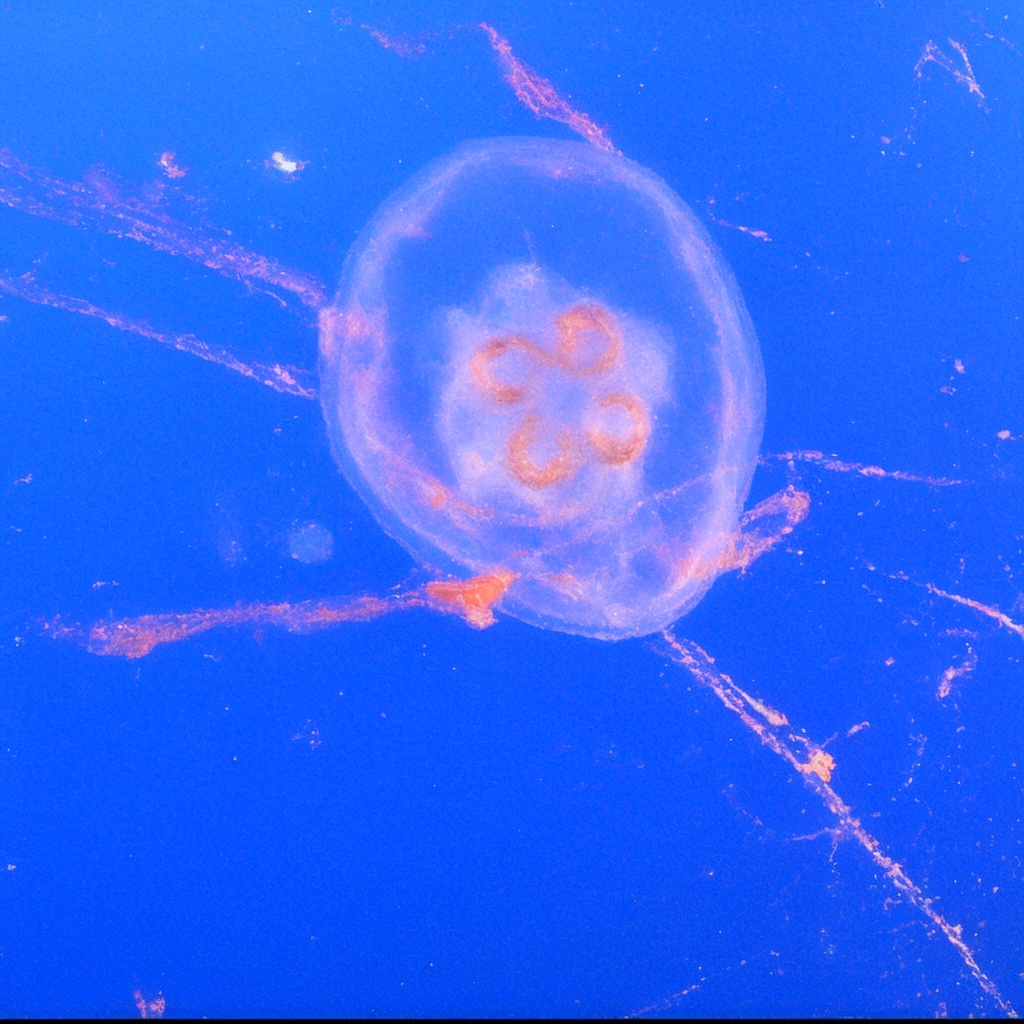} 
						\\
      					LQ Input \hspace{\g} &
						StableSR~\cite{stablesr}  \hspace{\g} &
						DiffBIR~\cite{diffbir}  
						\\
      						\includegraphics[height=\h \textwidth, width=\w \textwidth]{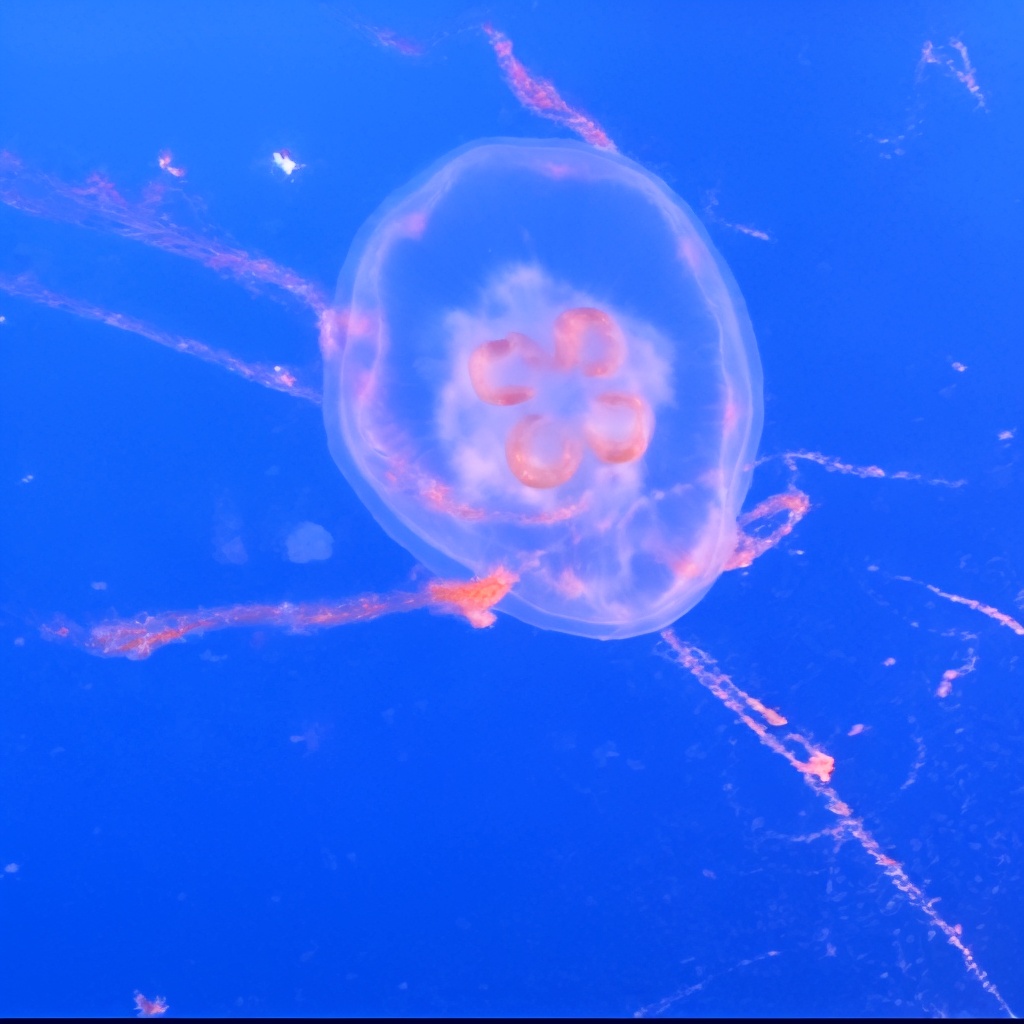} \hspace{\g} &
						\includegraphics[height=\h \textwidth, width=\w \textwidth]{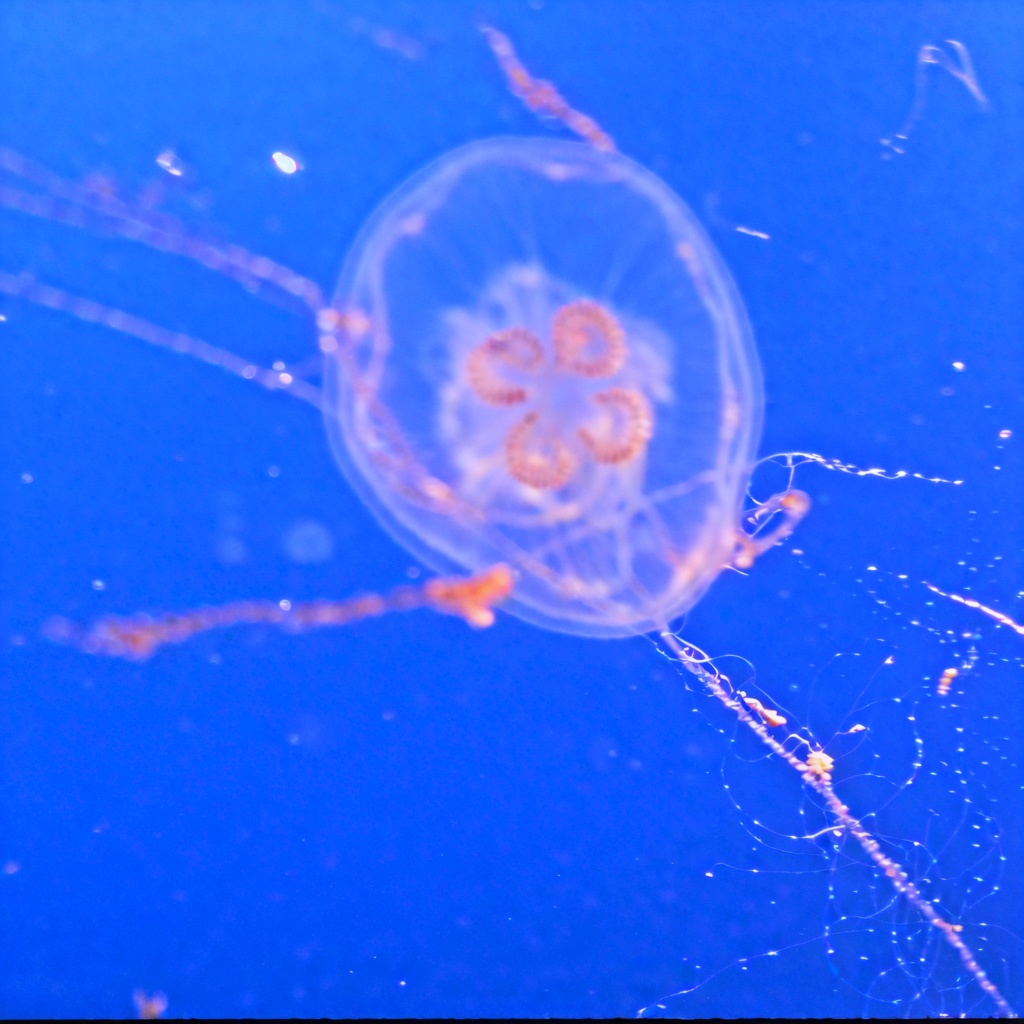} \hspace{\g} &
						\includegraphics[height=\h \textwidth, width=\w \textwidth]{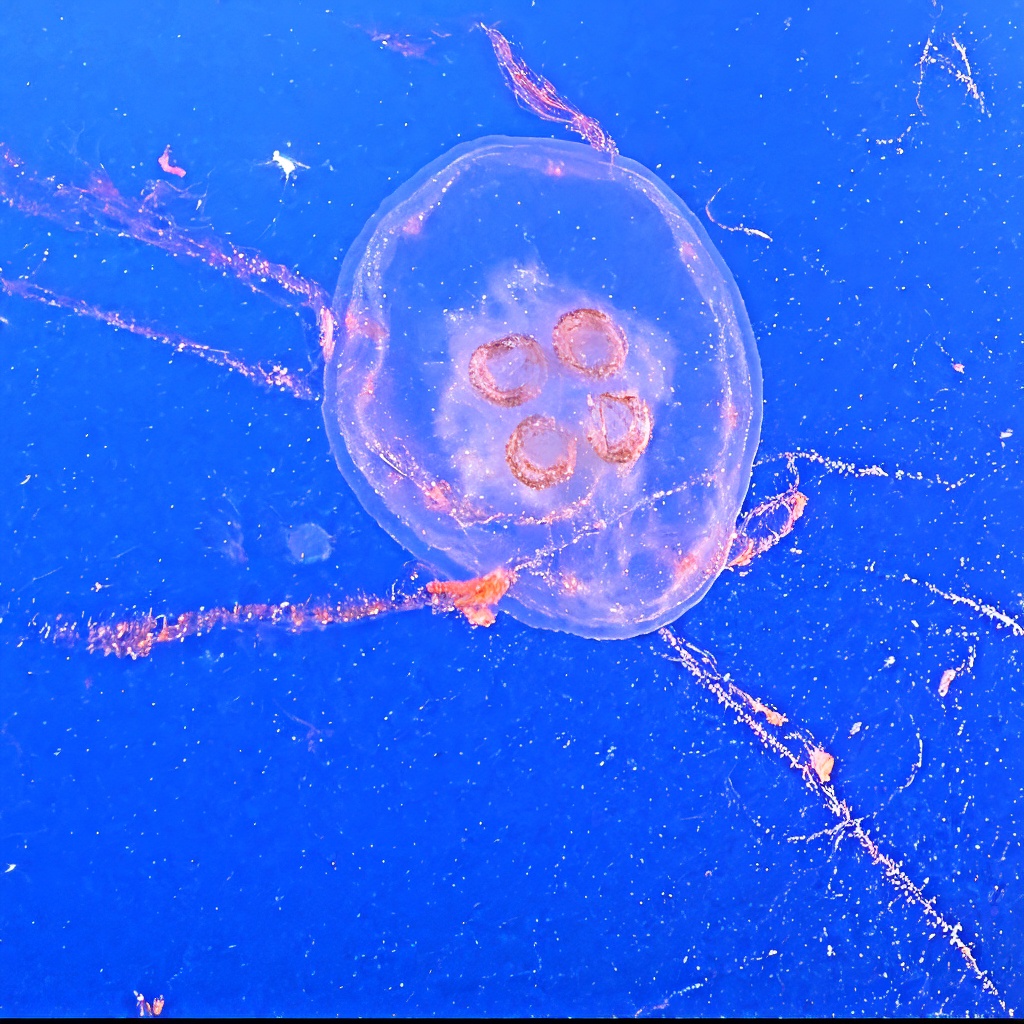} 
						\\
      					SeeSR~\cite{seesr} \hspace{\g}&
                            SUPIR~\cite{supir} \hspace{\g}&
						 \textbf{\modelname{}} (Ours)
                                          \vspace{1mm}

						\\
      						\includegraphics[height=\h \textwidth, width=\w \textwidth]{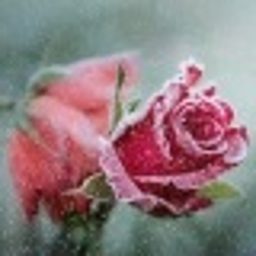} \hspace{\g} &
						\includegraphics[height=\h \textwidth, width=\w \textwidth]{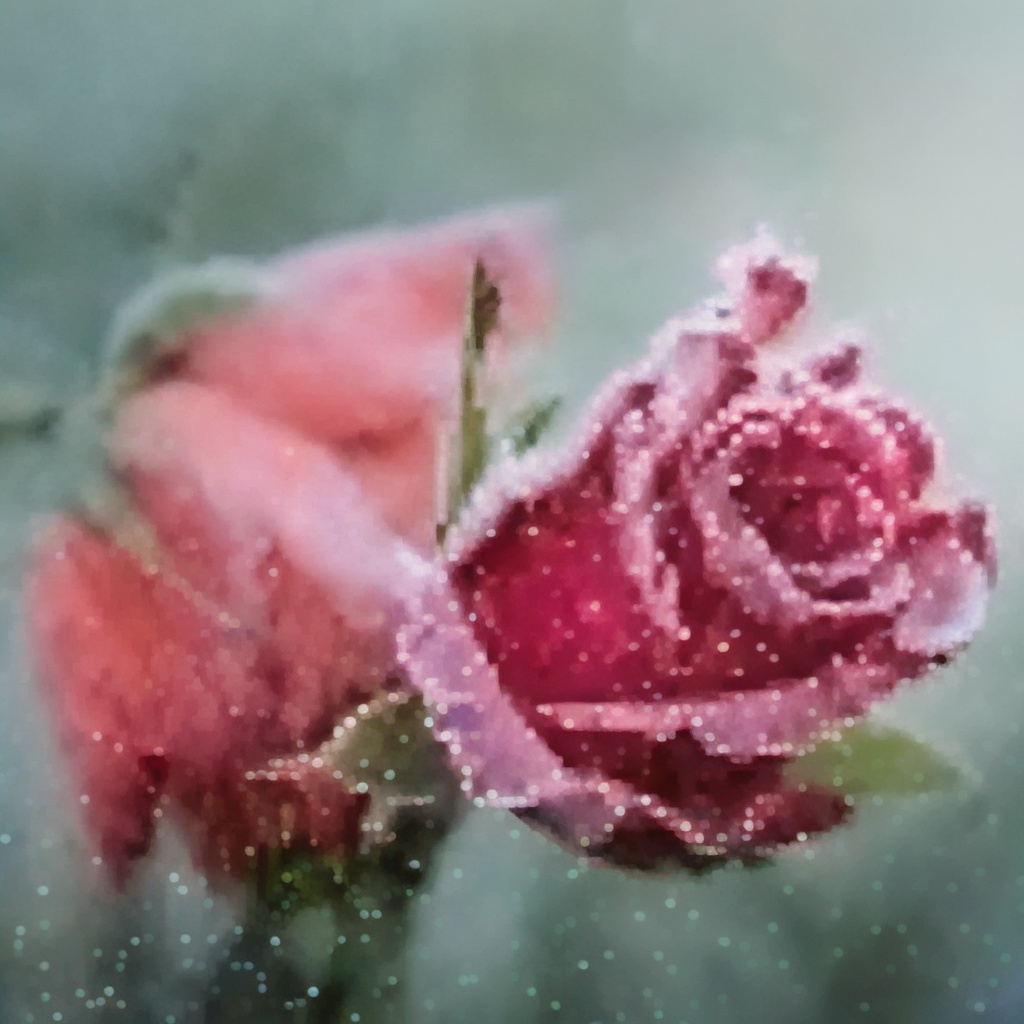} \hspace{\g} &
						\includegraphics[height=\h \textwidth, width=\w \textwidth]{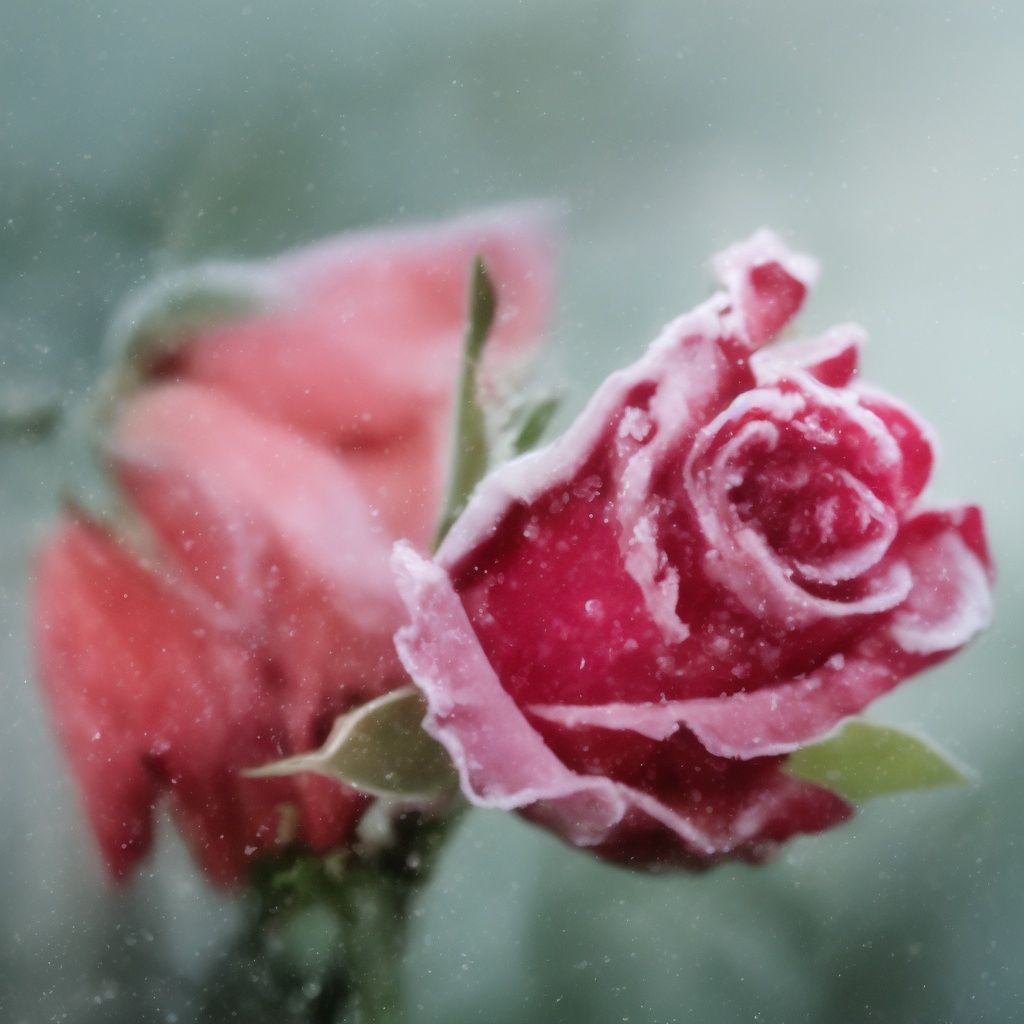} 
						\\
      					LQ Input \hspace{\g} &
						StableSR~\cite{stablesr}  \hspace{\g} &
						DiffBIR~\cite{diffbir}  
						\\
      						\includegraphics[height=\h \textwidth, width=\w \textwidth]{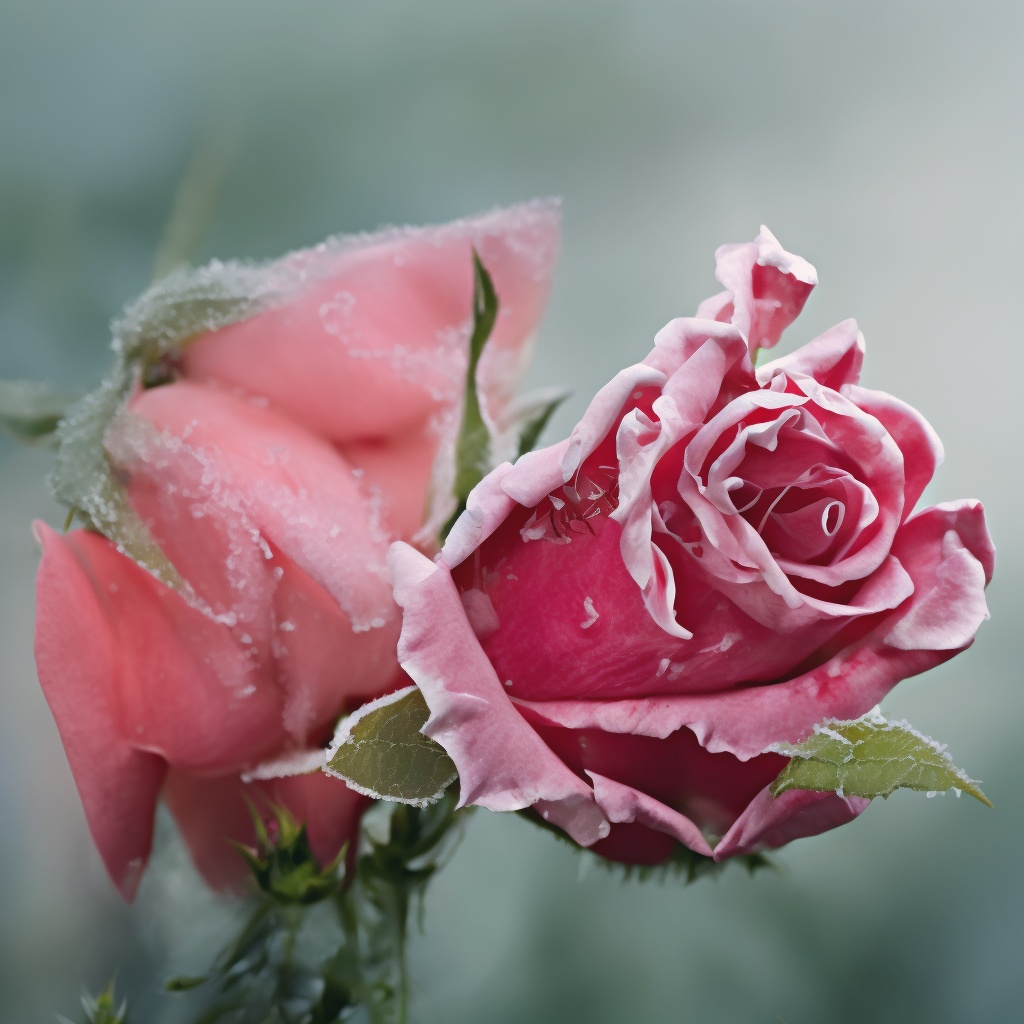} \hspace{\g} &
						\includegraphics[height=\h \textwidth, width=\w \textwidth]{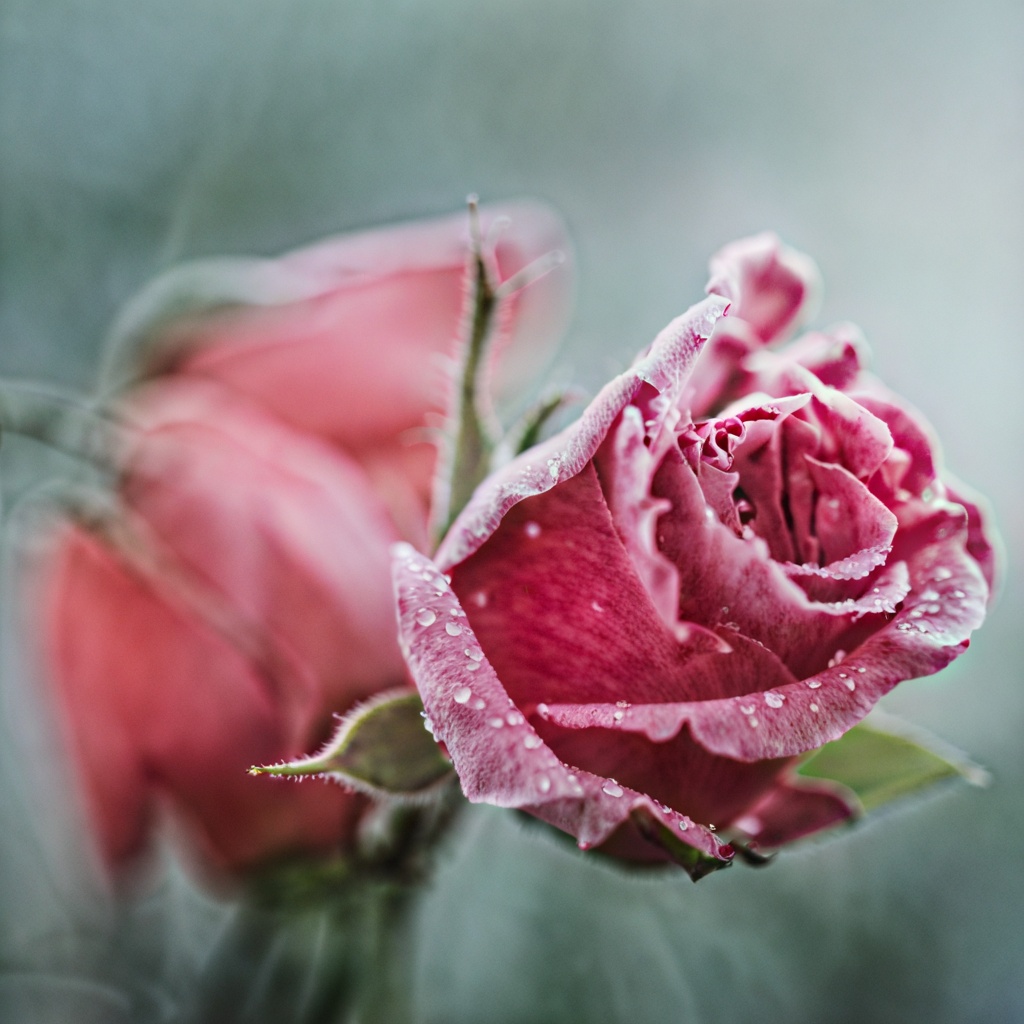} \hspace{\g} &
						\includegraphics[height=\h \textwidth, width=\w \textwidth]{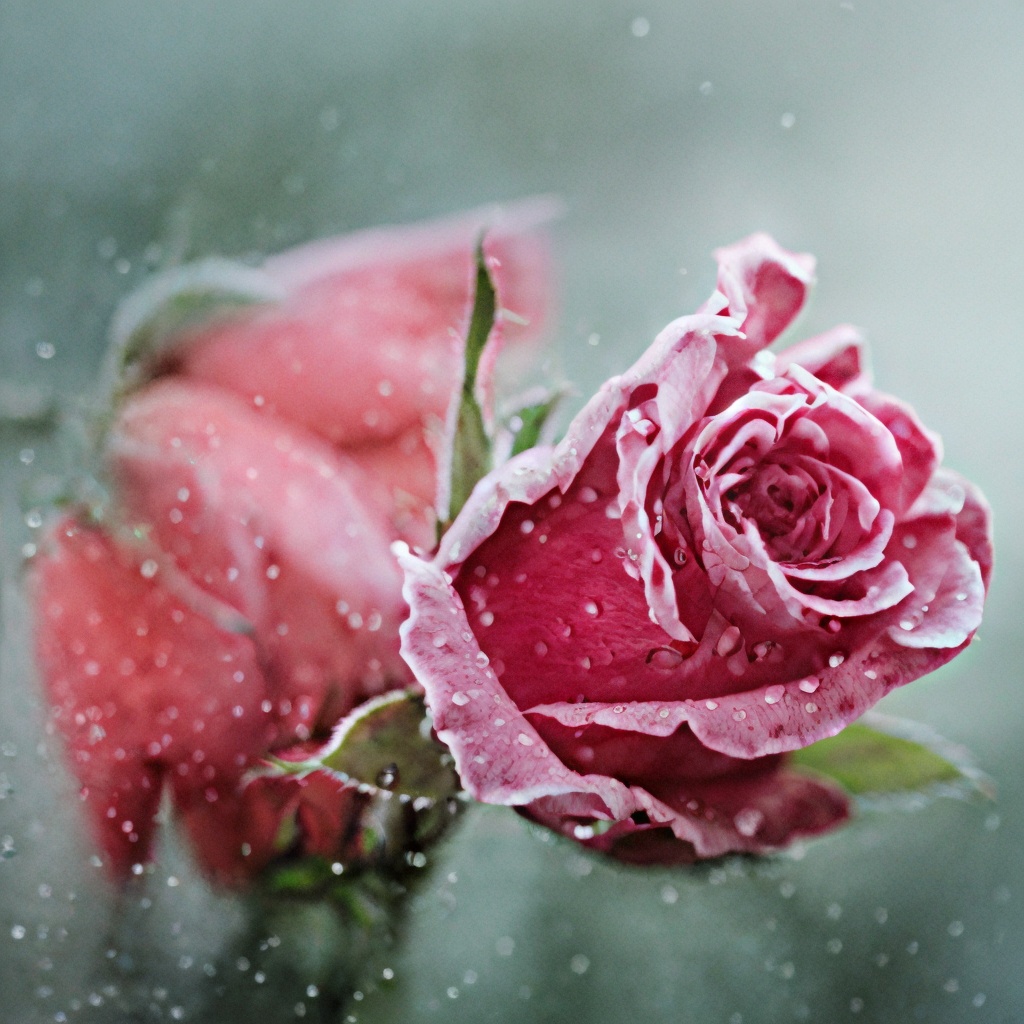} 
						\\
      					SeeSR~\cite{seesr} \hspace{\g}&
                            SUPIR~\cite{supir} \hspace{\g}&
						 \textbf{\modelname{}} (Ours)
						\\

					\end{tabular}
				\end{adjustbox}
        }
    	\caption{Visual comparisons on real-world benchmarks (2/3). Please zoom in for a better view.}
	\label{fig:visual_2}
\end{figure}

%% file: figs_appendix/tex/visual_3.tex
\begin{figure}[!htbp]
	\scriptsize
	\centering
	\newcommand{\h}{0.105}
	\newcommand{\wa}{0.12}
	\newcommand{\wb}{0.16}
	\newcommand{\g}{-0.7mm}
 	\setlength\tabcolsep{1.5pt}
	\renewcommand{\arraystretch}{1}
	\resizebox{1.00\linewidth}{!} {
			\renewcommand{\h}{0.186}
			\newcommand{\w}{0.186}
				\begin{adjustbox}{valign=t}
					\begin{tabular}{ccc}
						\includegraphics[height=\h \textwidth, width=\w \textwidth]{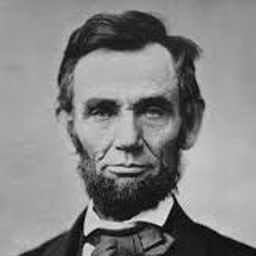} \hspace{\g} &
						\includegraphics[height=\h \textwidth, width=\w \textwidth]{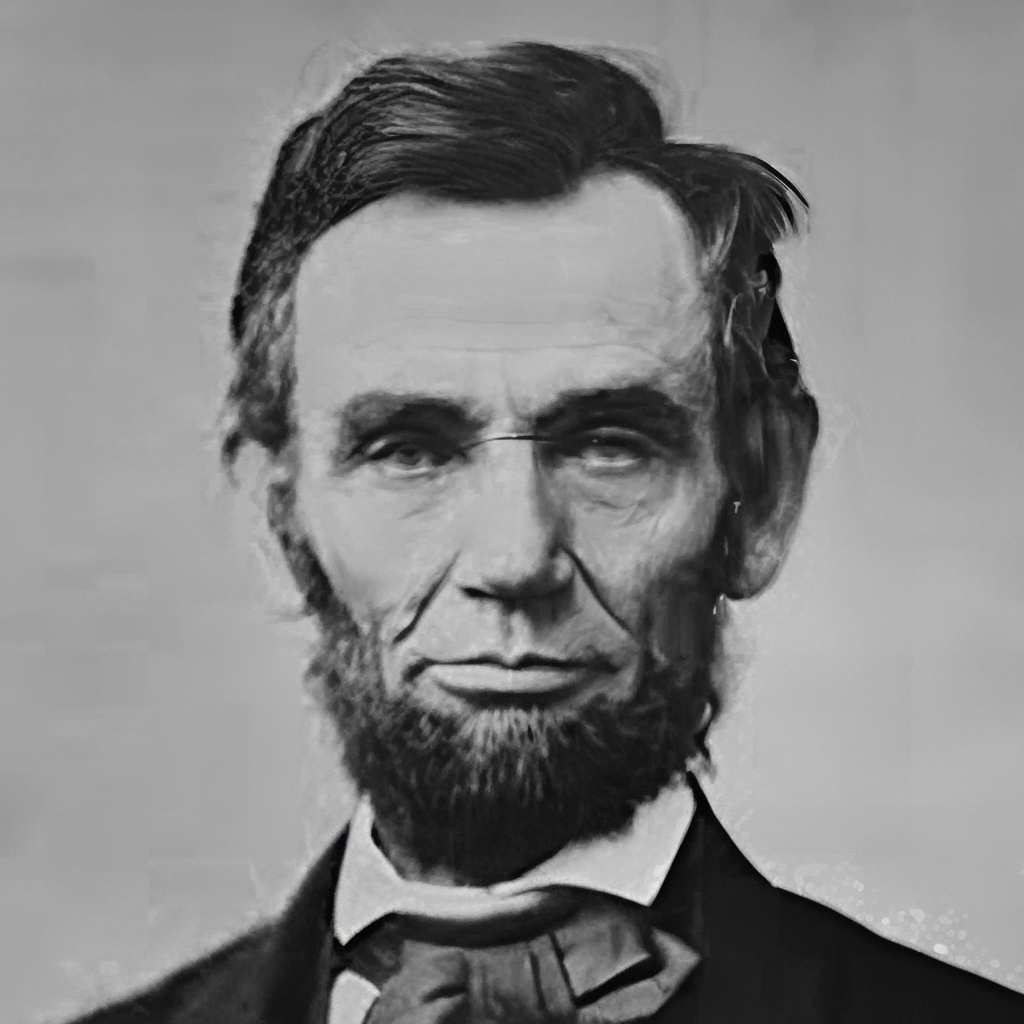} \hspace{\g} &
						\includegraphics[height=\h \textwidth, width=\w \textwidth]{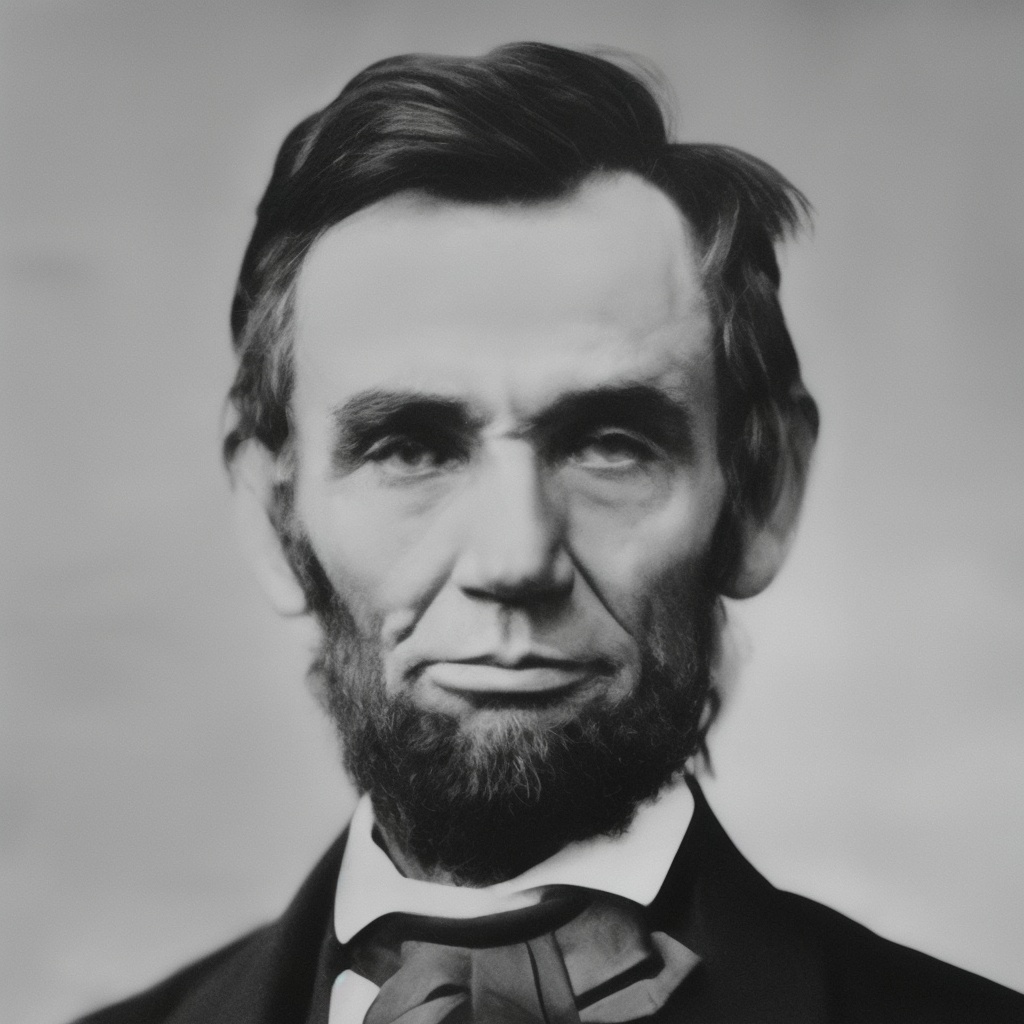} 
						\\
      					LQ Input \hspace{\g} &
						StableSR~\cite{stablesr}  \hspace{\g} &
						DiffBIR~\cite{diffbir}  
						\\
      						\includegraphics[height=\h \textwidth, width=\w \textwidth]{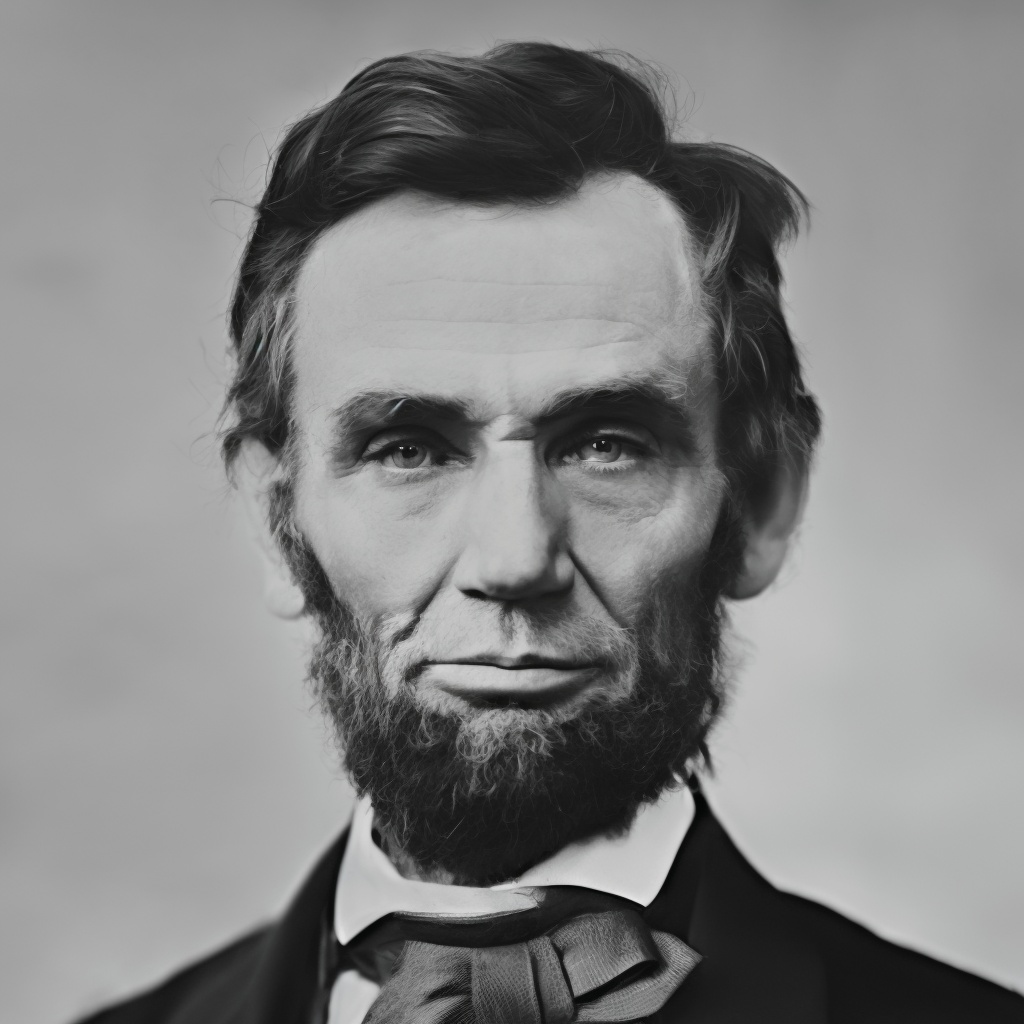} \hspace{\g} &
						\includegraphics[height=\h \textwidth, width=\w \textwidth]{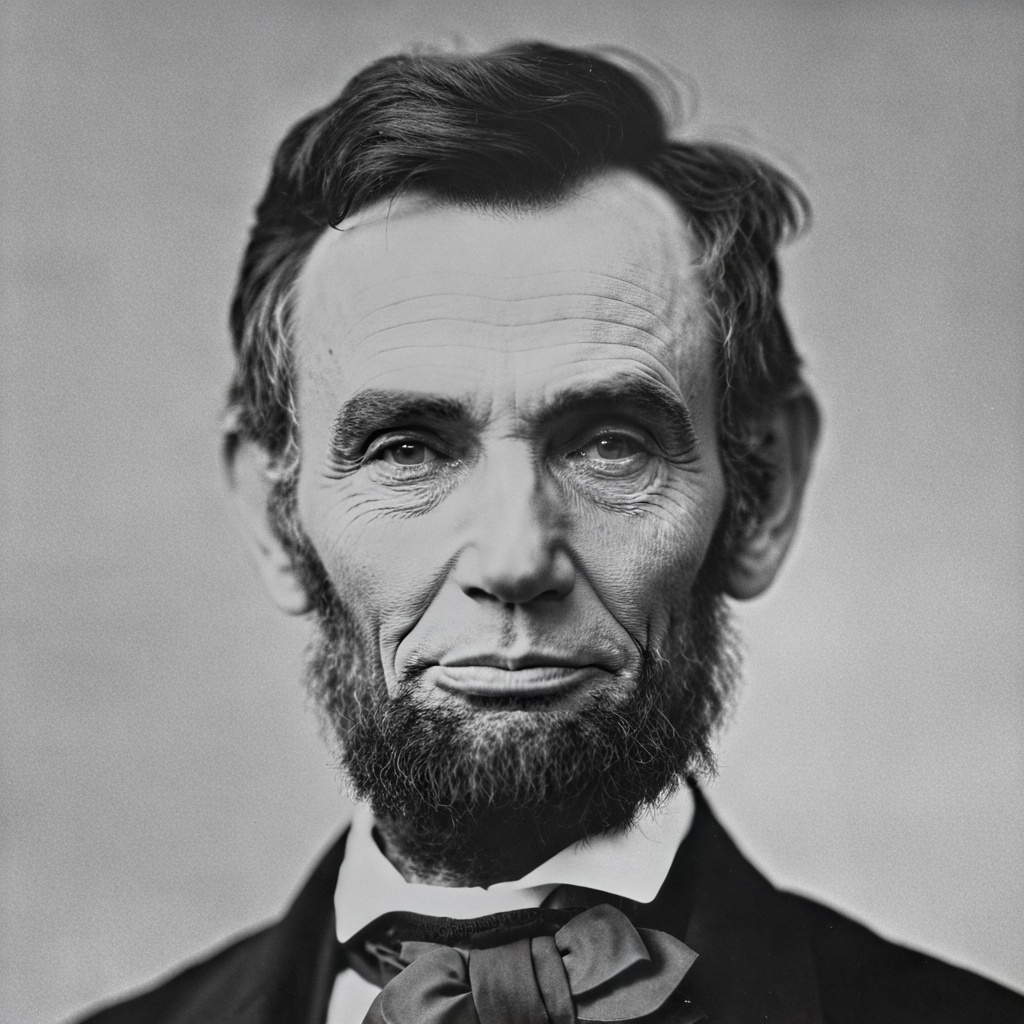} \hspace{\g} &
						\includegraphics[height=\h \textwidth, width=\w \textwidth]{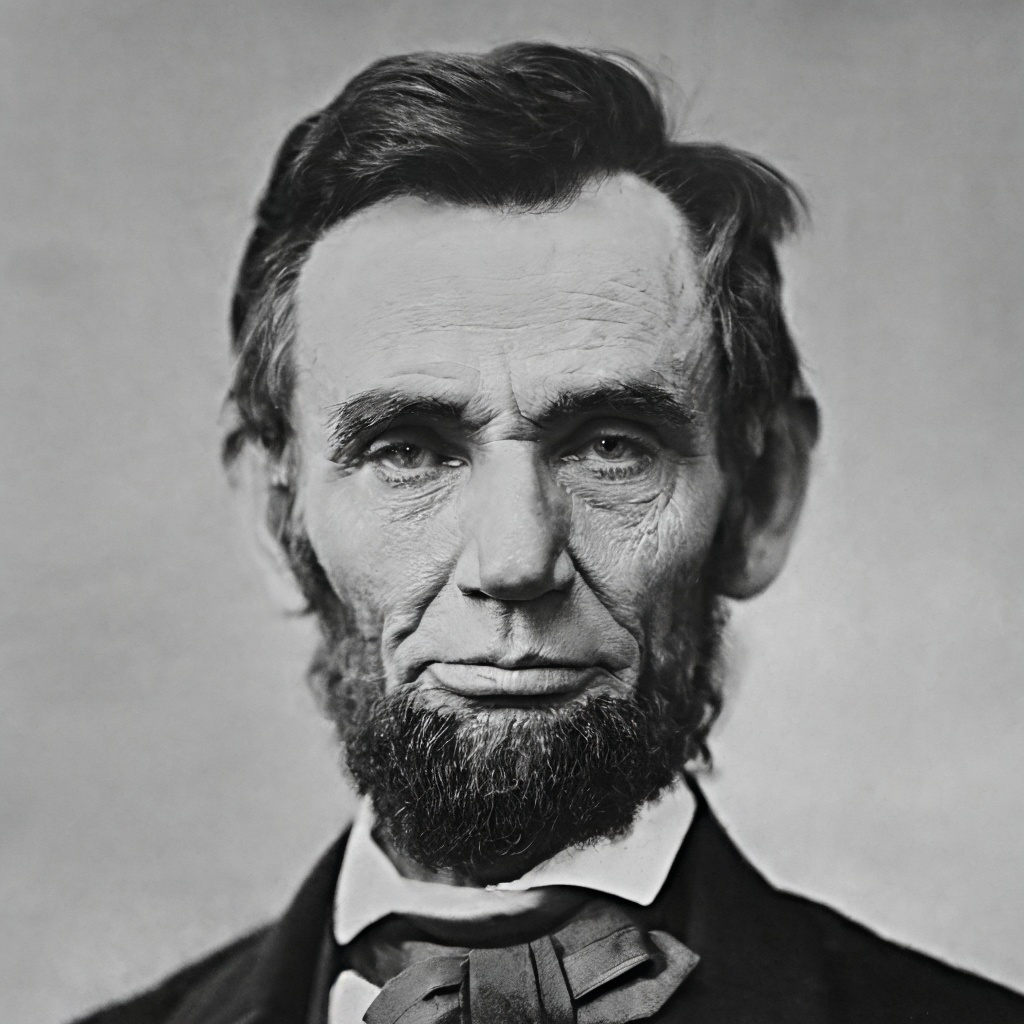} 
						\\
      					SeeSR~\cite{seesr} \hspace{\g}&
                            SUPIR~\cite{supir} \hspace{\g}&
						 \textbf{\modelname{}} (Ours)
                                          \vspace{1mm}

						\\
      						\includegraphics[height=\h \textwidth, width=\w \textwidth]{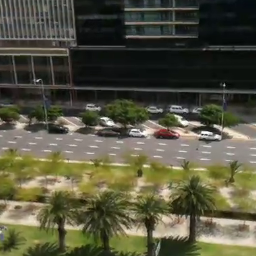} \hspace{\g} &
						\includegraphics[height=\h \textwidth, width=\w \textwidth]{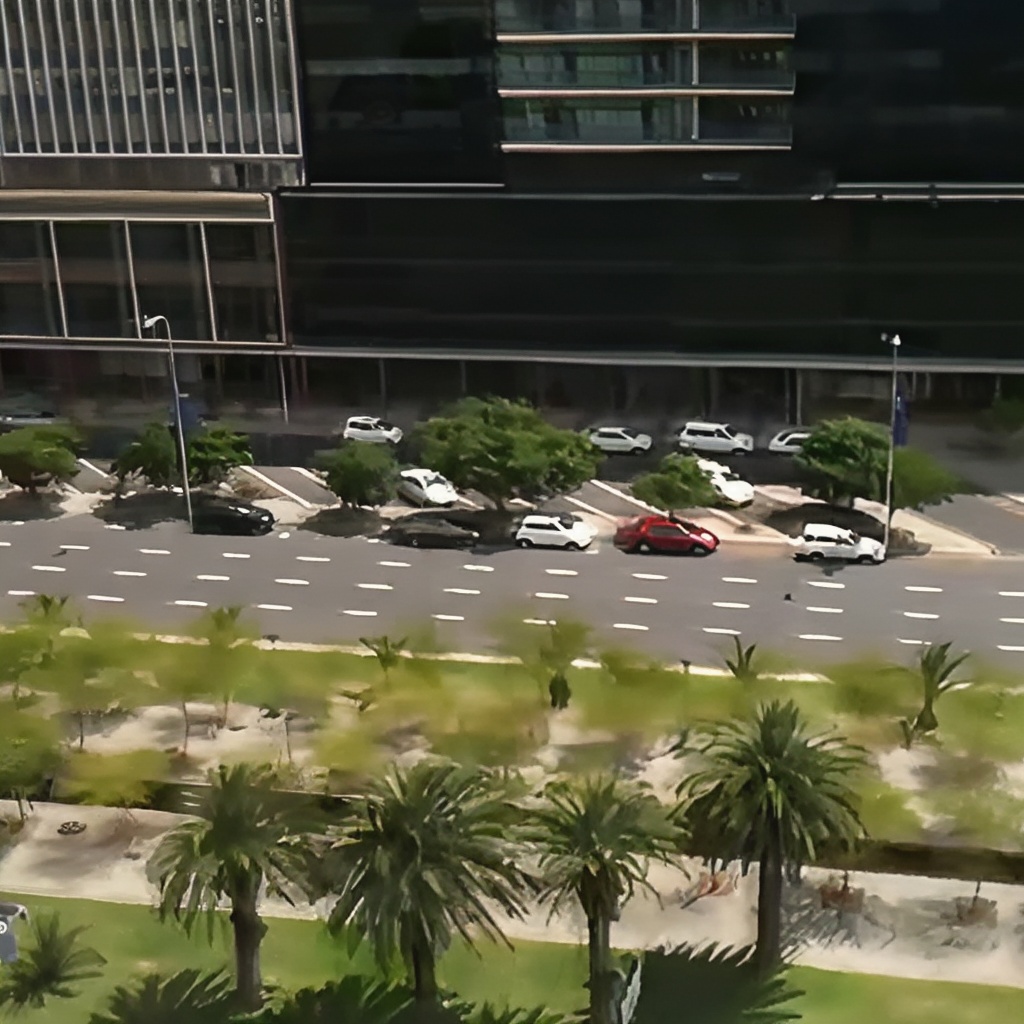} \hspace{\g} &
						\includegraphics[height=\h \textwidth, width=\w \textwidth]{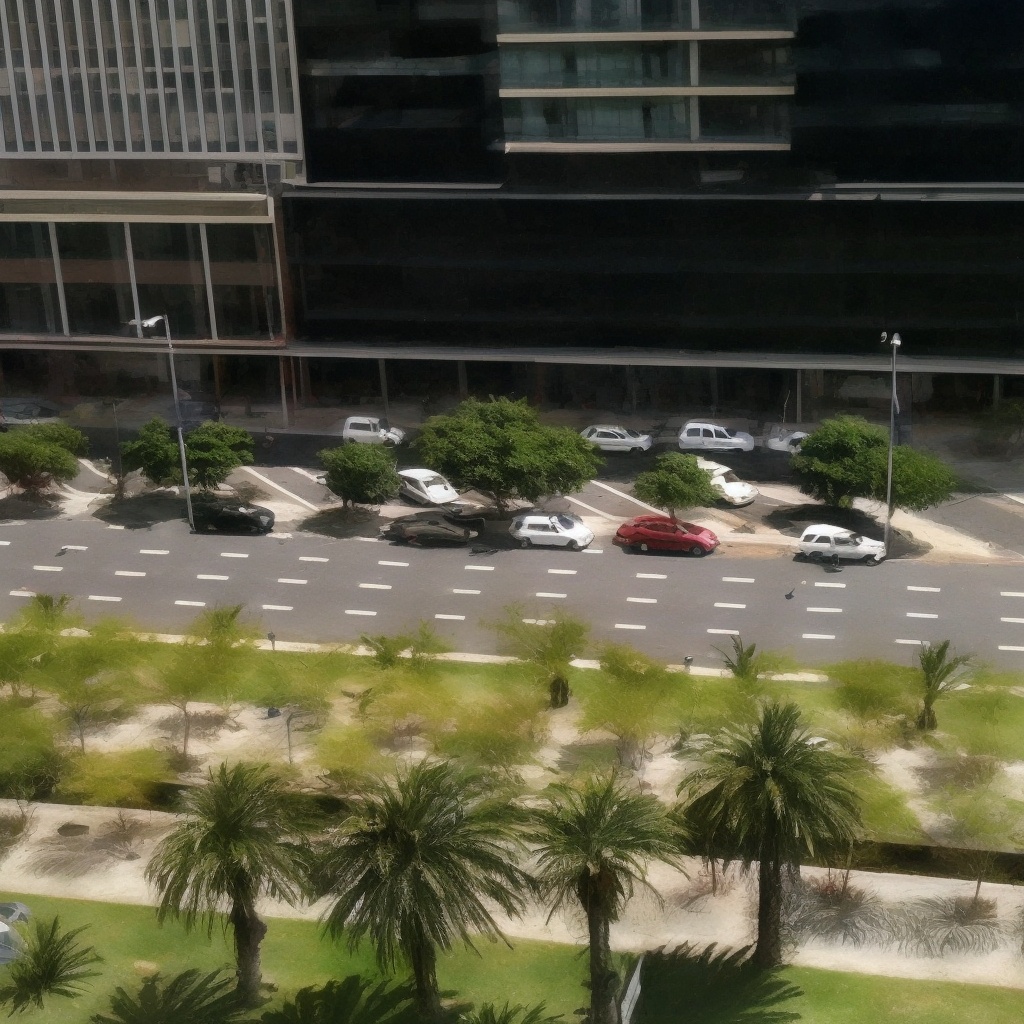} 
						\\
      					LQ Input \hspace{\g} &
						StableSR~\cite{stablesr}  \hspace{\g} &
						DiffBIR~\cite{diffbir}  
						\\
      						\includegraphics[height=\h \textwidth, width=\w \textwidth]{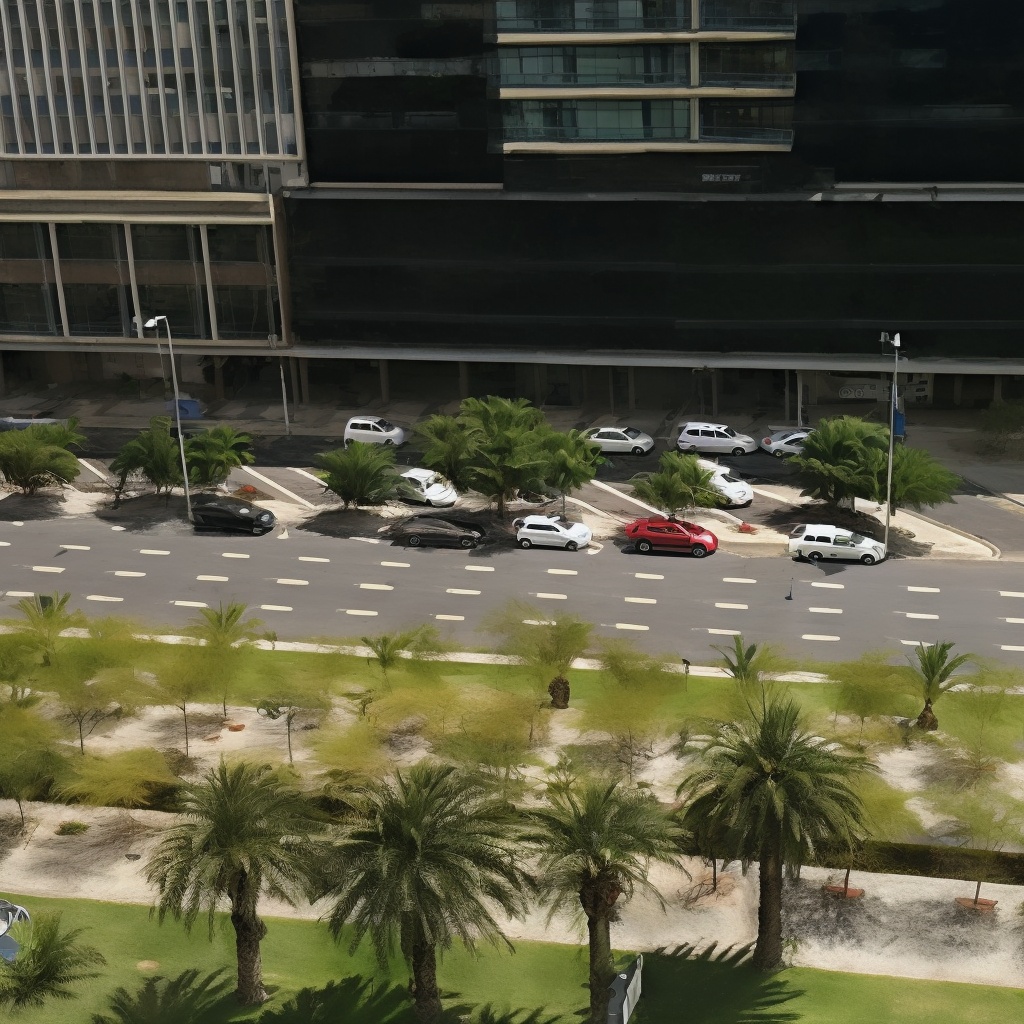} \hspace{\g} &
						\includegraphics[height=\h \textwidth, width=\w \textwidth]{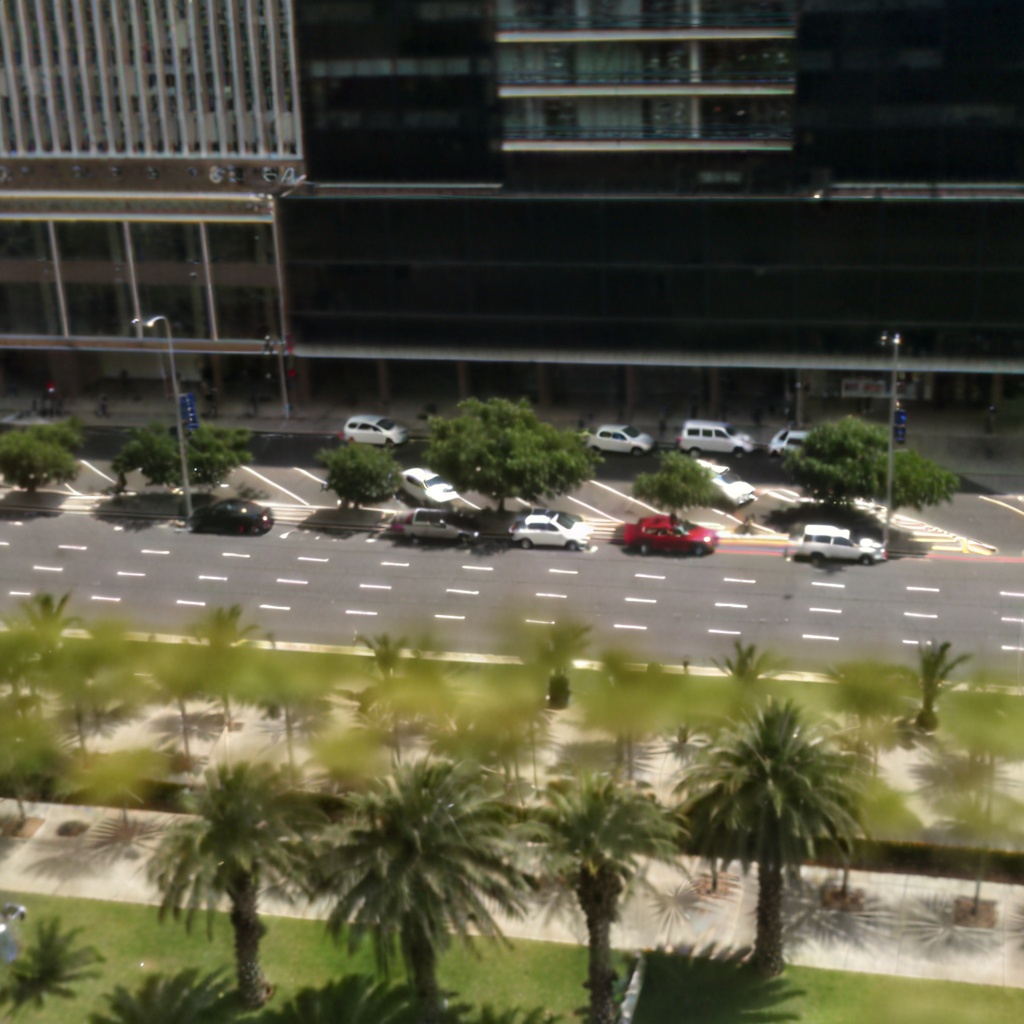} \hspace{\g} &
						\includegraphics[height=\h \textwidth, width=\w \textwidth]{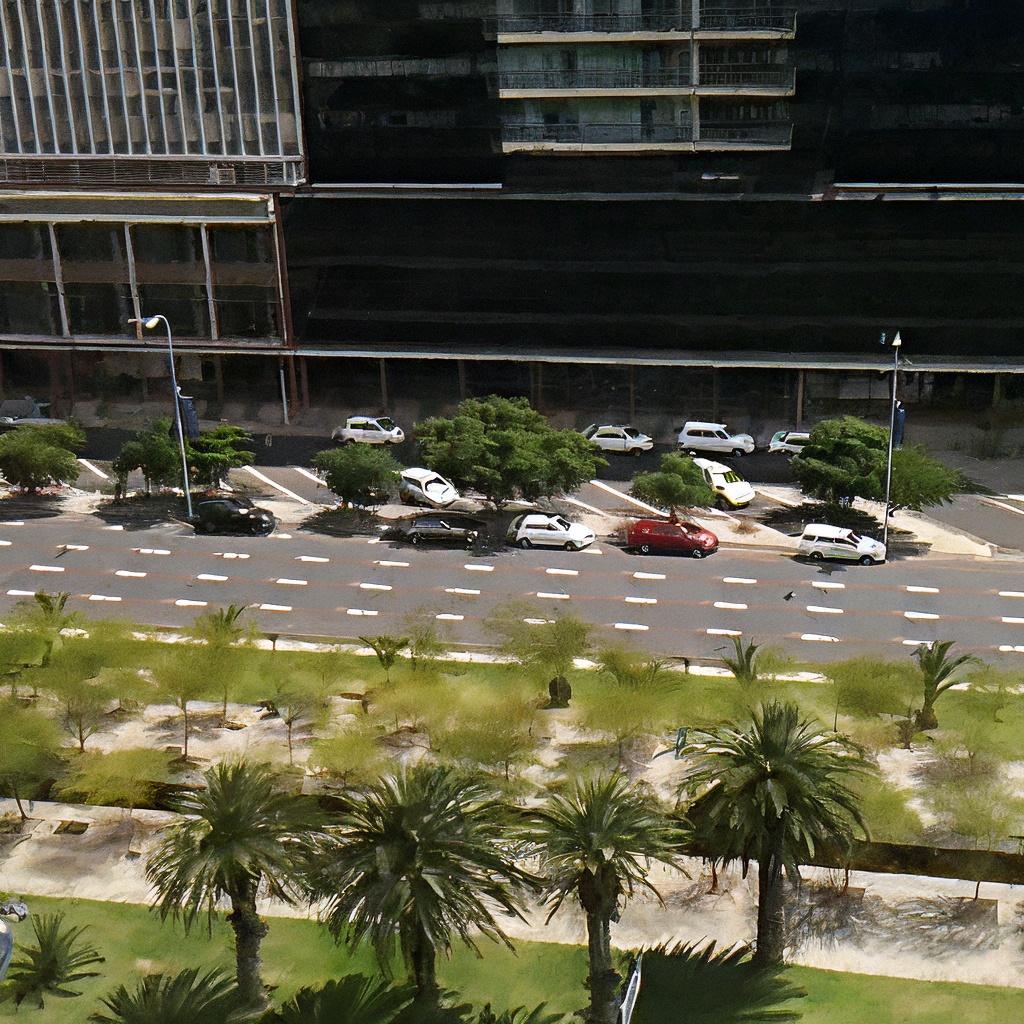} 
						\\
      					SeeSR~\cite{seesr} \hspace{\g}&
                            SUPIR~\cite{supir} \hspace{\g}&
						 \textbf{\modelname{}} (Ours)
						\\

					\end{tabular}
				\end{adjustbox}
        }
    	\caption{Visual comparisons on real-world benchmarks (3/3). Please zoom in for a better view.}
	\label{fig:visual_3}
\end{figure}